\let\latexdocument\document
\let\latexenddocument\enddocument
\documentclass{clv3_arxiv}
\let\document\latexdocument
\let\enddocument\latexenddocument

\makeatletter
\AtBeginDocument{%
  \if@filesw
    \immediate\openout\@mainqry=\jobname.qry
  \fi
}
\AtEndDocument{%
   \ifx\@biography\@empty\else{\par\ifbrief\vskip10pt\fi\biofont\noindent\@biography\par}\fi
   \immediate\closeout\@mainqry
      \ifodd\c@page\clearpage\thispagestyle{empty}\null\clearpage\else\clearpage\fi
}
\makeatother

\NewCommandCopy{\cnumdef}{\numdef}
\NewCommandCopy{\endcnumdef}{\endnumdef}
\let\numdef\relax \let\endnumdef\relax

\usepackage[utf8]{inputenc}

\usepackage[dvipsnames]{xcolor}
\usepackage{mathtools}
\DeclarePairedDelimiter{\norm}{\lVert}{\rVert}
\usepackage{multirow}
\usepackage{multicol}
\usepackage{graphicx}
\usepackage{subfig}
\usepackage{amsmath}
\usepackage{bbm}
\usepackage{arydshln}
\PassOptionsToPackage{hyphens}{url}\usepackage[hidelinks]{hyperref}
\hypersetup{
    colorlinks,
    linkcolor={red!50!black},
    citecolor={blue!50!black},
    urlcolor={blue!80!black}
}
\usepackage{relsize}
\usepackage{makecell}

\usepackage{listings}
\usepackage{rotating}
\usepackage{pdflscape}

\newcommand{\minisection}[1]{\noindent{\bf #1}\hspace{0.6em}}

\title{Towards Faithful Model Explanation in NLP: A Survey}
\author{Qing Lyu}
\affil{University of Pennsylvania\\
\texttt{lyuqing@sas.upenn.edu}}
\author{Marianna Apidianaki}
\affil{University of Pennsylvania\\
\texttt{marapi@seas.upenn.edu}}
\author{Chris Callison-Burch}
\affil{University of Pennsylvania\\
\texttt{ccb@seas.upenn.edu}}

\runningtitle{Towards Faithful Model Explanation in NLP: A Survey}
\runningauthor{Lyu et al.} 

\begin{document}

\maketitle

\begin{abstract}
End-to-end neural Natural Language Processing (NLP) models are notoriously difficult to understand. This has given rise to numerous efforts towards model explainability in recent years. One desideratum of model explanation is \emph{faithfulness}, i.e. an explanation should accurately represent the reasoning process behind the model's prediction. In this survey, we review over 110 model explanation methods in NLP through the lens of faithfulness. We first discuss the definition and evaluation of faithfulness, as well as its significance for explainability. We then introduce recent advances in faithful explanation, grouping existing approaches into five categories: similarity-based methods, analysis of model-internal structures, backpropagation-based methods, counterfactual intervention, and self-explanatory models. For each category, we synthesize its representative studies, strengths, and weaknesses. Finally, we summarize their common virtues and remaining challenges, and reflect on future work directions towards faithful explainability in NLP.

\end{abstract}
\section{Introduction}
\label{section:intro}

Since the birth of deep learning, end-to-end neural Language Models (LMs) have achieved remarkable success in a wide range of NLP tasks \cite[][i.a.]{clark_think_2018, wang_glue_2018, wang_superglue_2019, zellers_hellaswag_2019, cobbe_training_2021, hendrycks_measuring_2021, sakaguchi_winogrande_2021}. However, they largely remain a black-box to humans, i.e., they lack \textbf{explainability}. This hinders their real-life application, especially in high-stake scenarios where user trust is crucial. To address this issue, a plethora of explanation methods have been developed to shed light on how these LMs work. However, there are still many open questions in this field: How do we objectively evaluate the quality of explanation methods? How do we choose the most appropriate one(s) for a given use case? Does interpretability come at the cost of performance?

In this survey, we review over 110 model explanation methods through the lens of \textbf{faithfulness}, a fundamental principle of explanations. Faithfulness refers to the extent to which an explanation accurately reflects a model's reasoning process \cite{jacovi_towards_2020}. In other words, an explanation should not ``lie'' about the underlying mechanism at work. Explanations that lack faithfulness can be dangerous, especially when they still appear \textbf{plausible}, i.e., convincing to humans. This can mislead the users into over-trusting the model even if it has unwanted biases. In a variety of tasks including recidivism prediction, college acceptance prediction, and credit prediction, it is shown that when a model relies on sensitive features like race and gender, an unfaithful explanation can be exploited to hide these biases, leading users into believing that the model is innocuous \cite{pruthi_learning_2020, slack_fooling_2020}.

Despite the critical nature of faithfulness, in practice, it is frequently either overlooked or conflated with other principles of explanation, predominantly plausibility \cite{jacovi_towards_2020}. Moreover, even in cases where faithfulness is deliberately pursued and assessed, there is still no established consensus on how to quantitatively measure it. In cases where faithfulness is evaluated, authors typically commit to a single evaluation method, sometimes underpinned by problematic assumptions (see Secion~\ref{section:faithfulness_evaluation}). As a result, faithfulness evaluation results from various studies are often mutually incompatible, leading to a landscape of inconsistent findings.

Within the scope of this survey, we have chosen to spotlight faithfulness, as it is arguably the most important principle for explainability that remains to be formalized in NLP research. Most other principles, such as plausibility, have already received great attention in existing interpretability studies.\footnote{For example, see the encouraging progress on reducing visual noise (i.e., improving plausibility) in Backpropagation-based Methods in Section~\ref{section:backprop_methods}. } Also, they typically have a relatively stable definition and established evaluation methods, setting them apart from the more complicated notion of faithfulness.

In this survey, we will first clarify the concept of faithfulness itself and synthesize its multifaceted evaluation methodologies. Following this, we intend to provide a thorough critique of existing model explanation methods in the context of faithfulness, elucidating their strengths and shortcomings. Finally, we will summarize their common merits and remaining challenges, and conclude with potential avenues for further improvement. We hope that this survey will contribute to fostering a more transparent and standardized field of interpretability research. \\

\minisection{Target Audience.} For NLP students, researchers, and practitioners hoping to better understand the reasoning mechanism of their models, this survey will serve as an introductory manual of existing explanation methods and will help them choose the most suitable one(s) for their own use cases. For researchers interested in studying NLP interpretability, this survey will offer an accessible and comprehensive overview of state-of-the-art work in the field, laying the basis for further exploration. The challenges and future work directions identified in this survey will also pave the way for developing novel methodologies addressing the shortcomings of existing approaches.\\

\minisection{Contributions.} In the field of NLP interpretability, a few existing surveys/reviews have predominantly focused on an individual class of techniques, such as SHAP-based methods \cite{mosca_shap-based_2022}, saliency methods \cite{ding_evaluating_2021}, neuron-level analysis \cite{sajjad_neuron-level_2022}, and instance attribution methods \cite{pezeshkpour_empirical_2021}. More broadly, \citet{belinkov_analysis_2019} provide a systematic overview of model analysis methods in NLP, with an emphasis on methods that uncover what knowledge is encoded in LMs, rather than why models make certain predictions. In addition, \citet{danilevsky-etal-2020-survey} present a comprehensive review of existing model explanation methods in NLP, categorized along five axes. In our survey, we aim to refine the taxonomy to be more complete and intuitive (see Table~\ref{table:properties}), as well as place a specific emphasis on faithfulness in reviewing current explainability methods.

In summary, our contributions are as follows:
\begin{itemize}
    \item We present a comprehensive review of over 110 model explanation methods, offering a broad perspective of the field.
    \item We introduce a taxonomy that categorizes these methods into five families, which help clarify previously ambiguous terminology.\footnote{For example, ``saliency methods''. See Section~\ref{section:properties} for details.}
    \item We delve into an in-depth discussion on the concept of faithfulness, with a particular focus on critically reviewing its various methods of evaluation.
\end{itemize}

This survey does \textbf{not} purport to:
\begin{itemize}
    \item Conduct an empirical faithfulness evaluation of all discussed methods: Currently, there is still no universally accepted standard for evaluating faithfulness. Instead of committing to a single standard, we critically examine all existing evaluation methodologies, highlighting their motivation, assumptions, and potential constraints. In the subsequent discussion of various explanation methods, we aim to introduce empirical results under each evaluation standard whenever possible.
    \item Provide a simple answer to ``which family of methods is the most faithful'': Given the diversity of methods in each family, it is impractical and unjust to provide a sweeping assertion of which family is categorically more faithful than others. Our objective instead is to elucidate (a) which families are intrinsically more/less driven by faithfulness, and (b) within each family, what faithfulness challenges each method confronts, and which methods are more faithful than others. In essence, we hope to convince the reader that when deciding which explanation methods to use in practice, it is recommended to choose those that are intrinsically motivated and empirically validated in terms of faithfulness, while still remaining aware of potential pitfalls highlighted afterwards. 
\end{itemize}

% \begin{figure}[!t]
%   \centering
%   \includegraphics[width=\textwidth]{imgs/diagram.pdf}
%   \caption{An outline of the survey.}\label{figure:outline}
% \end{figure}

% \onecolumn
\begin{sidewaysfigure*}
    \centering
    % \hspace{3cm}
    \vspace{12cm}
  \includegraphics[width=0.95\linewidth]{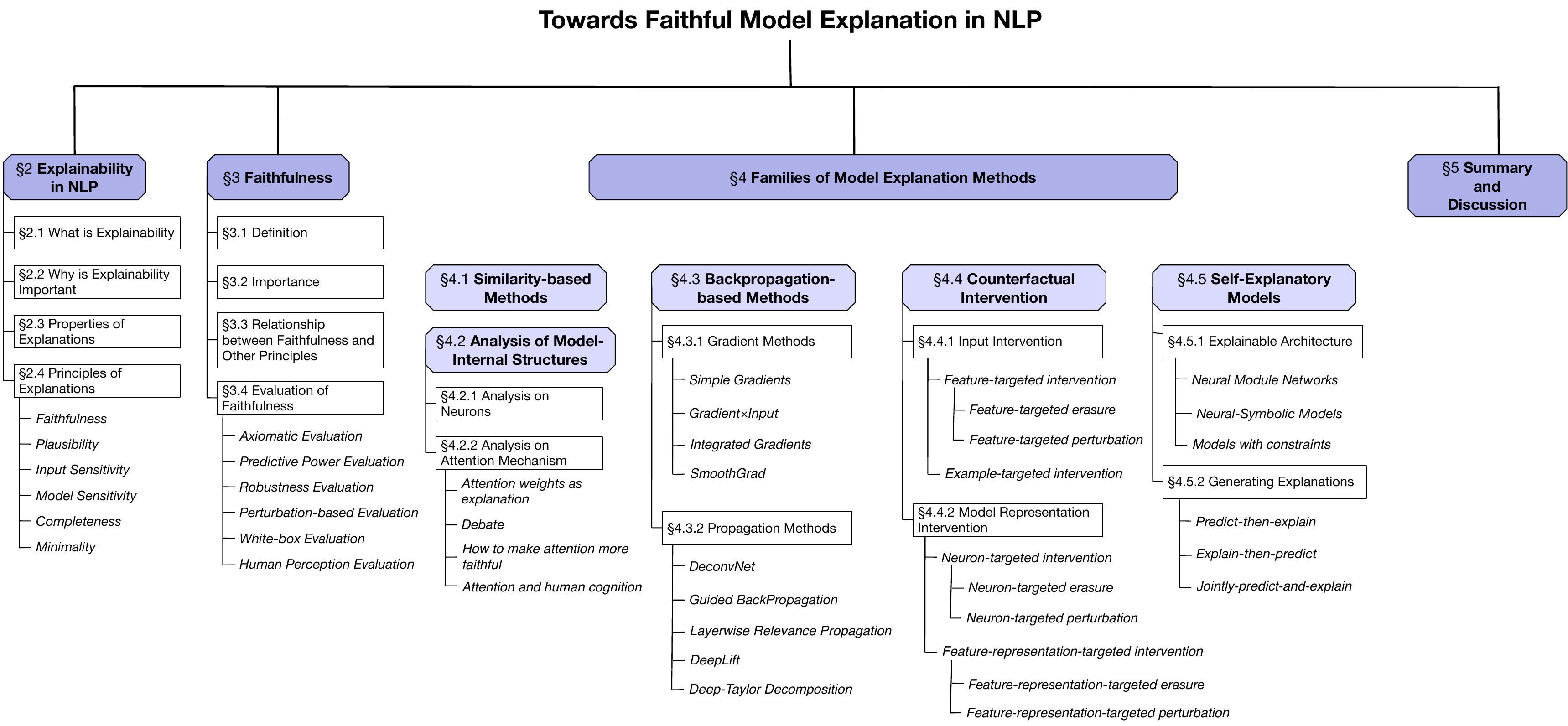}
  \vspace{1cm}
  \caption{An overview of the survey.}\label{figure:outline}
\end{sidewaysfigure*}

\minisection{Survey Organization.} The survey is structured as follows: 
\begin{itemize}
    \item Section~\ref{section:explanability} introduces the general notion of explainability in NLP, including its definition, importance, characterizing properties, and principles.
    \item Section~\ref{section:faithfulness} zooms in on faithfulness as a fundamental principle of model explanations, discussing its definition, importance, relationship with other principles, and possible means of evaluation.
    \item Section~\ref{section:attempts} synthesizes existing model explanation methods in pursuit of faithfulness, by grouping them into five categories: similarity-based methods, analysis of model-internal structures, backpropagation-based methods, counterfactual intervention, and self-explanatory models.
    \item Section~\ref{section:summary_discussion} discusses their common virtues and current challenges, and identifies future work directions towards improving faithfulness in explainable NLP.
    \item Section~\ref{section:conclusion} concludes the survey.
\end{itemize}

Figure~\ref{figure:outline} shows a structured overview of the survey. To make it easier to navigate through each section, we provide a more detailed \textbf{outline} with method names and relevant pointers in the Supplementary Materials.

\section{Explainability in NLP}
\label{section:explanability}
We start by introducing the notion of explainability in NLP, discussing its definition and importance. To prepare for our analysis of model explanation methods in subsequent sections, we will also present a set of properties that can serve to categorize different methods, as well as several common design principles.

\subsection{What Is Explainability}
\label{section:explainability_definition}

In the context of Machine Learning, \textit{explainability} (also referred to as \textit{interpretability}\footnote{Despite their subtle distinctions in some previous literature, we use the terms interchangeably in this survey.}) stands for

\begin{quote}
the extent to which the \emph{internal mechanics} of a model can be presented in understandable terms to a \emph{human} \cite{lipton_mythos_2017, murdoch_definitions_2019, barredo_arrieta_explainable_2020}. 
\end{quote}

\noindent Despite this intuitive notion, explainability has no established technical definition in the community, which results in numerous papers ``wielding the term in a quasi-scientific way'', essentially referring to different concepts \cite{lipton_mythos_2017, doshi-velez_towards_2017, miller_explanation_2018, murdoch_definitions_2019}. We argue that the confusion mainly lies in the interpretation of two key concepts in the above definition: (a) what are the \textbf{``internal mechanics''} and (b) who is the \textbf{``human''}.\\

\minisection{Internal mechanics.} This can refer to either (i) \textbf{\emph{what} knowledge a model encodes}, or (ii) \textbf{\emph{why} a model makes certain predictions}.\footnote{There can be interesting ``how'' questions as well, for example, ``how to make interpretability insights useful'', which we believe are beyond the scope of the internal mechanics of the model, but rather address how to best apply the insights obtained from interpretability methods. Another example can be ``how a model makes certain predictions'', but this is essentially equivalent to ``why a model makes certain predictions'' in our terms.} Existing work in NLP explainability can be categorized according to which of these questions it is addressing:\\

\noindent (i) The ``what'' type of work aims to accurately characterize the extent to which a model $M$ encodes some target knowledge $K$, which can be linguistic, commonsense, world knowledge, etc. For example, given a Machine Translation (MT) model, does it implicitly capture linguistic knowledge such as the syntactic tree of the source sentence \cite{belinkov_linguistic_2020}? Previous research answers such questions with methods including but not limited to:
\begin{itemize}
    \item \textbf{Probing classifiers} (or auxiliary/diagnostic classifiers) \cite{veldhoen_diagnostic_2016, adi_fine-grained_2017, conneau_what_2018};
    \item \textbf{Information-theoretic measurement} \cite{voita_information-theoretic_2020, lovering_information-theoretic_2020};
    \item \textbf{Behavioral tests}: including challenge sets \cite{levesque_winograd_2012, gardner_evaluating_2020}, adversarial attacks \cite{ebrahimi_hotflip_2018, wallace_universal_2019}, and prompting \cite{petroni_language_2019, choenni_stepmothers_2021};
    \item \textbf{Visualization} \cite{coenen_visualizing_2019, ethayarajh_how_2019, karidi_putting_2021}.
\end{itemize}

\noindent It should be noted that ``what is known'' does not imply ``what is used''. For example, even if an MT model \textit{encodes} knowledge about syntax trees, it does not necessarily \textit{use} it in translation. In fact, it has been shown that LMs do incidentally encode linguistic features even when they are irrelevant to the end task labels \cite{ravichander_probing_2021}. This leads us to the ``why'' question.\\

\noindent (ii) The ``why'' type of work addresses the \textit{causal} question of what factors (input features, model structures, decision rules, etc.) have led the model $M$ to certain predictions $Y$. This line of research aims to establish causality between these potential factors and the prediction through various approaches. These approaches will be the main focus of this survey, since there already exists a comprehensive review of studies on the ``what'' question \cite{belinkov_analysis_2019}.\\

\minisection{Human.} This refers to the target audience of the explanation, which can include but is not limited to the model developers, fellow researchers, industry practitioners, or end users. Depending on the audience, the form and goal of the explanation can be entirely different. We will revisit this in Section~\ref{section:properties} - Target Audience.\\

Instantiating the initial definition with the two clarified concepts, this survey will cover work on NLP explainability in the following sense: \textbf{the extent to which \emph{why a model makes certain predictions} can be presented in understandable terms to \emph{a certain type of target audience}}.

\subsection{Why Is Explainability Important}
\label{section:explainability_importance}

Compared to classic Machine Learning models, end-to-end neural NLP models are intrinsically harder to understand in terms of their reasoning mechanism \cite{bommasani_opportunities_2021}. For example, it is easy to interpret a Decision Tree, since every node denotes a decision rule. By contrast, it is much more opaque what a node/layer in a Neural Network (NN) represents and how it contributes to the prediction. Nevertheless, NNs have become the predominant paradigm in NLP research and are increasingly being adopted in real-life applications, which has driven the need to better understand their behavior. Concretely, we identify three key reasons why explainability is important:

First, explainability can allow us to \textbf{discover artifacts in datasets}. Solving the dataset does not mean learning the task, since there can often be unexpected shortcuts (e.g., statistical cues) in data creation. Models are surprisingly good at exploiting them \cite[][i.a.]{kaushik_how_2018, mccoy_right_2019, geva_are_2019}. For example, only using the hypothesis allows the model to perform decently on Natural Language Inference (NLI)\footnote{Given a premise $P$ and a hypothesis $H$, the NLI task is to determine if $P$ semantically entails $H$ (whenever $P$ is true, is $H$ also true?). For example, \textit{it rains heavily today} entails \textit{it rains today}.} datasets \cite{poliak_hypothesis_2018}. Explaining the contribution of various features to the prediction will help us discover such artifacts and create more reliable datasets.

Additionally, explainability can assist in \textbf{diagnosing a model's strengths and weaknesses, and debugging it}. Explainability allows us to find where a model succeeds or fails, and fix the weaknesses before they can be exploited by adversaries. For example, if a model secretly relies on unwanted biases on gender and race, we can diagnose and eliminate them through explanations \cite{ravfogel_null_2020}. Also, if a model is susceptible to subtle perturbations in the data, it is better to discover and guard this in development prior to deployment \cite{wallace_universal_2019}.

Finally, explainability may help \textbf{calibrate user trust in high-stake applications}. In domains like health, law, finance, and human resources, an end user may not trust a model if it only provides a prediction but no explanation. For example, in computer-aided diagnosis, if an algorithm provides a prediction along with supporting evidence, such as relevant symptoms, it could be easier for human decision-makers to determine when to trust the model prediction and when to be skeptical. 
% In such high-stake scenarios, it is crucial to ensure that models are making safe decisions for the right reasons.
Empirical studies have found that explanation quality highly influences the level of user trust in the model decision \cite{kunkel_let_2019, ye_unreliability_2022}. In some cases, however, it is also possible for users to blindly trust the model decision simply because of the presence (instead of the content) of the explanation \cite{bansal_does_2021}. Even worse, malicious actors can manipulate user trust by carefully designing misleading explanations \cite{lakkaraju_how_2020}. All these underscore the need for rigorous evaluation of explanation methods and emphasize the importance of fostering a broader understanding and literacy in AI interpretability among end users.

One important caveat lies in the interplay between explainability and performance. In some cases, it has been found that there exists an empirical trade-off between these two factors \cite[][i.a.]{camburu_e-snli_2018, narang_wt5_2020, subramanian_obtaining_2020, hase_leakage-adjusted_2020}, where a more interpretable model can result in lower accuracy on the end task. Nonetheless, there are also studies that have shown otherwise, suggesting explanations can boost models' performance across a variety of tasks, particularly under low-resource settings \cite[][i.a.]{wei_chain_2022, wang_self-consistency_2022}. We will delve into this topic in greater depth in Section~\ref{section:self-explantory_models}.

\subsection{Properties of Explanations}
\label{section:properties}

We propose to characterize model explanation methods in terms of the following set of properties:

\begin{table*}[!t]
\small
\caption{Comparison of different model explanation methods in terms of their properties. Different colors denote different values of a property. See Section~\ref{section:properties} for details.}
\centering
\scalebox{0.85}{
\renewcommand{\arraystretch}{2}
\begin{tabular}{l|ccccc}
    \hline\hline \thead{Method} & \thead{Time} & \thead{Model\\accessibility} & \thead{Scope} & \thead{Unit of\\explanation} & \thead{Form of\\explanation}\\
    \hline 
    
    \makecell{Similarity-based \\methods} & \textcolor{cyan}{post-hoc} & \textcolor{OliveGreen}{white-box} & \textcolor{Dandelion}{local}  & \makecell[c]{\textcolor{Periwinkle}{examples},\\\textcolor{DarkOrchid}{concepts}} & \makecell[c]{\textcolor{Sepia}{importance scores}} \\
    
    \makecell{Analysis of \\model-internal\\ structures} & \textcolor{cyan}{post-hoc} & \textcolor{OliveGreen}{white-box} & \makecell[c]{\textcolor{Dandelion}{local},\\\textcolor{Orange}{global}}  & \makecell[c]{\textcolor{Orchid}{features},\\\textcolor{Violet}{interactions}} & \makecell[c]{\textcolor{Bittersweet}{visualization},\\\textcolor{Sepia}{importance scores}} \\
    
    \makecell{Backpropagation \\-based methods} & \textcolor{cyan}{post-hoc} & \textcolor{OliveGreen}{white-box} & \textcolor{Dandelion}{local}  & \makecell[c]{\textcolor{Orchid}{features},\\\textcolor{Violet}{interactions}} & \makecell[c]{\textcolor{Bittersweet}{visualization},\\\textcolor{Sepia}{importance scores}} \\
    
    \makecell{Counterfactual\\ intervention} & \textcolor{cyan}{post-hoc}& \makecell[c]{\textcolor{YellowGreen}{black-box},\\\textcolor{OliveGreen}{white-box}} & \makecell[c]{\textcolor{Dandelion}{local},\\\textcolor{Orange}{global}}  & \makecell[c]{\textcolor{Orchid}{features},\\\textcolor{Periwinkle}{examples},\\\textcolor{DarkOrchid}{concepts}} & \makecell[c]{\textcolor{Sepia}{importance scores}} \\
    
    \makecell{Self-explanatory \\models} & \textcolor{blue}{built-in} & \makecell[c]{\textcolor{OliveGreen}{white-box}} & \makecell[c]{\textcolor{Dandelion}{local},\\\textcolor{Orange}{global}}  & \makecell[c]{\textcolor{Orchid}{features},\\\textcolor{Periwinkle}{examples},\\\textcolor{DarkOrchid}{concepts}} & \makecell[c]{\textcolor{Sepia}{importance scores},\\\textcolor{BurntOrange}{natural language},\\\textcolor{Tan}{causal graphs}}\\
   
    \hline\hline
\end{tabular}
}
\label{table:properties}
\end{table*}

\begin{enumerate}
    \item[(a)] \textbf{Time}: when the explanation happens. An explanation can be \textbf{post-hoc}, i.e., it is produced after the prediction. Any opaque model is given, and then an external method explains its predictions. Or, an explanation can be \textbf{built-in}, i.e., it is produced at the same time as the prediction. This type of model is so-called ``self-explanatory''.
    \item[(b)] \textbf{Model accessibility}: what parts of the model the explanation method has access to. A \textbf{black-box} explanation can only see the model's input and output, while a \textbf{white-box} explanation can additionally access the model-internal structures and representations.
    \item[(c)] \textbf{Scope}: where the explanation applies in the dataset. A \textbf{local} explanation only explains a model's behavior on a single data point (or a local vicinity of the data point), whereas a \textbf{global} explanation provides insights into the general reasoning mechanisms for the entire data distribution.
    \item[(d)] \textbf{Unit of explanation}: what the explanation is in terms of. A prediction can be explained in terms of \textbf{input features} \cite{ribeiro_why_2016}, \textbf{examples} \cite{wallace_interpreting_2018}, \textbf{neurons} \cite{sajjad_neuron-level_2022}, \textbf{concepts}\footnote{Prior work has different definitions of ``concepts'', including but not limited to phrases \cite{rajagopal_selfexplain_2021} and high-level features \cite{jacovi_contrastive_2021, abraham_cebab_2022}.} \cite{rajagopal_selfexplain_2021, dalvi_discovering_2022}, \textbf{feature interactions} \cite{hao_self-attention_2021}, or a \textbf{combination} of them \cite{jacovi_contrastive_2021}.
    \item[(e)] \textbf{Form of explanation}: how the explanation is presented. Typical forms include \textbf{visualization} \cite{li_visualizing_2016}, \textbf{importance scores} \cite{arras_explaining_2016}, \textbf{natural language} \cite{kumar_nile_2020}, or \textbf{causal graphs} \cite{dalvi_explaining_2021}. 
    Note that unit and form are different: to illustrate, for gradient methods in Table~\ref{table:properties}, the unit of explanation is input features, and the form is importance scores. 
    \item[(f)] \textbf{Target audience}: who is the explanation provided for. As mentioned in Section~\ref{section:explainability_definition}, different audience groups can have distinct goals with respect to model explanations. \textbf{Model developers} may want to debug the model; \textbf{fellow researchers} may want to find how the model can be extended/improved; \textbf{industry practitioners} may want to assess if the model complies with practical regulations; and \textbf{end users} may want to verify that they can safely rely on the model's decisions.

\end{enumerate}

 As a preview, Table~\ref{table:properties} compares the model explanation approaches to be discussed in terms of these properties.\footnote{Note that the target audience is not included in the table, since it largely depends on the task and the specific instance of the method.} Note that many existing taxonomies do not explicitly state which properties are taken into account, thus often producing confusing terms. For instance, certain taxonomies juxtapose example-based, local, and global as three classes of explanations, but they are in fact not about the same property. As another example, the term \textit{saliency methods} has been used to refer to ``backpropagation-based methods'' in our taxonomy. However, \textit{saliency} only describes the form and the unit of explanation -- importance scores of input features. Then, technically speaking, all methods in Table~\ref{table:properties} have instances that can be called a \textit{saliency method}. In our taxonomy, we characterize each method using all these properties, so as to provide a clearer comparison along each dimension.

\subsection{Principles of Explanations}
\label{section:principles}

To motivate principled design and evaluation of model explanations, existing research identifies various principles that a good explanation should satisfy. We hereby provide a non-exhaustive synthesis of these principles.\\

\noindent \textbf{Faithfulness} (also referred to as \emph{fidelity} or \emph{reliability}): \textit{An explanation should accurately reflect the reasoning process behind the model's prediction} \cite{harrington_harvey_1985, ribeiro_why_2016, jacovi_towards_2020}. This is often considered the most fundamental requirement for any explanation, and sometimes used interchangeably with the term ``interpretability'' \cite[][i.a.]{jain_attention_2019, bastings_elephant_2020, jacovi_towards_2020}. After all, what is an explanation if it \textit{lies} about what the model does under the hood? An unfaithful explanation can look plausible to humans, but has little to do with how the model makes the prediction. For example, by looking at the attention weights\footnote{We will elaborate on the attention mechanism in Section~\ref{section:analysis_structure}.} of a sentiment classification model, it may be intuitive to interpret tokens with higher weights as ``more important'' to the prediction, whereas empirically it is questionable if such causal relation exists \cite{jain_attention_2019}.\\

\noindent \textbf{Plausibility} (also referred to as \emph{persuasiveness} or \emph{understandability}): \textit{An explanation should be understandable and convincing to the target audience} \cite{herman_promise_2019, jacovi_towards_2020}. This implies that plausibility depends on who the target audience is. For example, a relevance propagation graph across NN layers may be a perfectly understandable explanation for model developers, but not at all meaningful to non-expert end users.\\

\noindent \textbf{Input Sensitivity}: This term has been perceived in slightly different senses, but the general idea is that \textit{an explanation should be sensitive (respectively insensitive) to changes in the input that influence (respectively do not influence) the prediction.} Specifically, it includes the following senses:

(i) \citet{sundararajan_axiomatic_2017}:
With respect to explanations whose unit is features and form is importance scores, if two inputs differ only at one feature and lead to different model predictions, then the explanation should assign non-zero importance to the feature.

(ii) \citet{adebayo_sanity_2018}: if the input data labels are randomized, the explanation should also change. This is because label randomization does change model predictions, so the explanation should be sensitive.

(iii) \citet{samek_reliability_2019}: if a constant shift is added to every input example in the training data, the explanation should not change. This is because the model is invariant to the constant shift, and thus the explanation should not be sensitive.\\

\noindent \textbf{Model Sensitivity}: Similar to Input Sensitivity, the term has been used in different senses. In general it means that \textit{an explanation should be sensitive (resp. insensitive) to changes in the model that influence (resp. do not influence) the prediction.} Specifically, it includes the following senses:

(i) \citet{sundararajan_axiomatic_2017}: an explanation should be insensitive to model implementation details that do not affect the prediction. More specifically, two models are \textit{functionally equivalent} if their outputs are equal for \emph{all} possible inputs, although their implementations might be different. The explanation should always be identical for functionally equivalent models. This principle is called Implementation Invariance.

(ii) \citet{adebayo_sanity_2018}: if the model weights are randomized, the explanation should also change. Similar to Input Sensitivity Definition (ii), weight randomization does change model predictions, so the explanation should be sensitive.\\

\noindent \textbf{Completeness}: \emph{An explanation should comprehensively cover all relevant factors to the prediction} \cite{sundararajan_axiomatic_2017}. More formally, for explanations in the form of importance scores, the importance of all features should sum up to some kind of ``total importance'' of the model output.\footnote{See Section~\ref{section:backprop_methods} -- Propagation Methods for more details on ``total importance''.} \\

\noindent \textbf{Minimality} (also referred to as \emph{compactness}): \textit{An explanation should only include the smallest number of necessary factors} \cite{halpern_causes_2005, miller_explanation_2018}. Intuitively, this is analogous to the Occam's razor principle, which prefers the simplest theory among all competing ones.\\

Note that not all of these principles have an established technical definition or evaluation standard, and there might not be consensus on whether they are necessary.\footnote{See Section~\ref{section:faithfulness_evaluation} -- "Axiomatic evaluation" for a critical discussion of a subset of these principles} In this survey, we mainly focus on faithfulness as it is generally considered one of the most central requirements for explanations \cite[][i.a.]{jain_attention_2019, bastings_elephant_2020, jacovi_towards_2020}, but we will still refer to the other principles when discussing relevant work in subsequent sections.

\section{Faithfulness}
\label{section:faithfulness}

Now, we will zoom in on faithfulness, a fundamental principle of model explanations, analyzing what it means, why it is important, its relationship with other principles, and how it can be measured.

\subsection{Definition}
\label{section:faithfulness_definition}
As mentioned before, a faithful explanation should \textbf{accurately reflect the reasoning process behind the model's prediction} \cite{harrington_harvey_1985, ribeiro_why_2016, jacovi_towards_2020}. This is only a loose description though; in fact, there is not yet a consistent and formal definition of faithfulness in the community. Instead, people often define faithfulness on an ad-hoc basis, in terms of different evaluation metrics, to be detailed in Section~\ref{section:faithfulness_evaluation}.

\subsection{Importance}
\label{section:faithfulness_importance}

We believe that faithfulness is one of the most fundamental principles for explainability, as also established in previous work \cite[][i.a.]{jain_attention_2019, bastings_elephant_2020, jacovi_towards_2020}. By definition (cf. Section~\ref{section:explainability_definition}), the goal of a model explanation is to reveal the mechanism behind the predictions in human-understandable terms. Faithfulness requires that the mechanism presented by the explanation is \textbf{true} to the model's underlying reasoning process. In other words, if an ``explanation'' is unfaithful, it does not even satisfy the definition of an explanation. 

In NLP specifically, there are two additional pieces of empirical evidence supporting the crucial role of faithfulness.

First, \textbf{faithfulness establishes causality.} In interpretability work, it is often implicitly assumed that ``what is known by the model'' is also ``what is used by the model in making predictions''. However, this assumption is not sound. For example, \citet{ravichander_probing_2021} show that language models encode linguistic features like tense and number, although they are irrelevant to the label of an artificially crafted end task. This means that a model can encode more features than what eventually gets used. Therefore, findings from the ``what is known by the model'' type of work are correlational but not causal. To establish causality, we need faithful explanations of how the model makes predictions.

Second, \textbf{an unfaithful explanation can be dangerous.} Consider an explanation that is not faithful but extremely plausible. Now, even if the model makes a wrong prediction, users may still trust it simply because the explanation looks convincing \cite{bansal_does_2021}. For example, \citet{pruthi_learning_2020} show that attention weights can be a deceiving explanation to end users. They train the model to attend minimally to gender-related tokens (\textit{he} and \textit{she}), therefore hiding the fact that it is relying on gender bias in prediction. Users may still find it safe to deploy the model in real-world scenarios since it seems to be free from bias.

\subsection{Relationship between Faithfulness and Other Principles}
\label{section:relation_principles}

We now elucidate the relationship between faithfulness and several other principles introduced in Section~\ref{section:principles}, since they are often implicitly conflated in the literature.\\

\minisection{Faithfulness vs. Plausibility.} There is an intrinsic tension between these two notions. Think about an extreme case: If an ``explanation'' is just a copy of all model weights, it would be perfectly faithful but not at all plausible to any target audience. Now consider the other extreme: We could ask the target audience which features are most important when they themselves make a prediction, then simply copy their response as our ``explanation'' of how the model works. This ``explanation'' would be perfectly plausible since it fully matches the human reasoning process, but not at all faithful since it has nothing to do with how the model works. Therefore, plausibility does not guarantee faithfulness, and vice versa.

Moreover, when we observe that an explanation is implausible in human terms, there can be two possibilities: (a) the model itself is not reasoning in the same way as humans do, or (b) the explanation is unfaithful. For instance, if an explanation says that a sentiment classification model is mainly relying on function words like \textit{the}, \textit{a} instead of sentiment words like \textit{awesome}, \textit{great}, it could be that the model is truly relying on those uninformative words to make predictions (potentially because of spurious correlations in the dataset), \textbf{or} that the model is correctly relying on content words, but the explanation does not faithfully reflect the fact. 

In practice, faithfulness has often been conflated with plausibility. For example, explanations that better align with human perception are often thought to be more faithful, while in fact, they are only more plausible. See Section~\ref{section:human_perception_evaluation} for more discussion. \\

\minisection{Faithfulness vs. Input Sensitivity, Model Sensitivity, and Completeness.} These principles are sometimes seen as necessary conditions for faithfulness, though not always explicitly stated \cite{sundararajan_axiomatic_2017, samek_reliability_2019, yeh_fidelity_2019}. Practically, they are often used to prove that an explanation is \textit{not} faithful via counterexamples. For instance, given an explanation method, researchers run it on a model and check through all the principles. If any of them (for instance, input sensitivity) is violated, then the explanation method is said to be unfaithful. This is also called sanity checks in the literature \cite{adebayo_sanity_2018}. We will revisit them in more detail in Section~\ref{section:axiomatic_evaluation}.

% \minisection{Faithfulness vs. Minimality.} There are no known relations between them, either from the literature or from the authors' point of view.

\subsection{Evaluation of Faithfulness}
\label{section:faithfulness_evaluation}

As explained in previous sections, faithfulness does not have an established formal definition, but is usually defined ad-hoc during evaluation. However, the evaluation metrics are often not directly comparable with each other and yield inconsistent results, making it difficult to objectively assess progress in this field.

In their seminal opinion piece, \citet{jacovi_towards_2020} outline five design principles of faithfulness evaluation metrics, three of which are the most important (and most ignored) in our view. The first principle is ``be explicit in what you evaluate'', especially, do not conflate plausibility and faithfulness. Second, ``faithfulness evaluation should not involve human judgment on explanation quality''. This is because humans do not know whether an explanation is faithful; if they did, the explanation would be unnecessary. Finally, faithfulness evaluation should not involve human-provided gold labels (for the examples to be explained). A faithful explanation method should be able to explain \textit{any} prediction of the model, regardless of whether it is correct or not.\footnote{The other two principles are ``do not trust 'inherent interpretability' claims'' and ``faithfulness evaluation of Intelligent User Interface systems should not rely on user performance''. The former is not relevant to the faithfulness evaluation methods, but rather to self-explanatory models, which we will revisit in Section~\ref{section:self-explantory_models}. The latter is similar to the second principle and thus omitted.}

With the above principles in mind, we review existing faithfulness evaluation methods, which we broadly categorize into: \textbf{axiomatic evaluation}, \textbf{predictive power evaluation}, \textbf{robustness evaluation}, \textbf{perturbation-based evaluation}, \textbf{white-box evaluation}, and \textbf{human perception evaluation}.

\subsubsection{Axiomatic Evaluation} \leavevmode\newline
\label{section:axiomatic_evaluation}

Axiomatic evaluation treats certain principles (also called \textit{axioms}) (e.g., those from Section~\ref{section:principles}) as \textit{necessary conditions} for faithfulness, and tests if an explanation method satisfies them. If it fails any test, then it is unfaithful. Existing tests focus on the following axioms: Model Sensitivity, Input Sensitivity, Polarity Consistency, Robustness Equivalence, and Feature Importance Agreement, among others.\\

\noindent \textbf{Model Sensitivity.} One form of Model Sensitivity, as previously mentioned, is Implementation Invariance \cite{sundararajan_axiomatic_2017}. This means that two functionally equivalent models (which have the same outputs for all inputs) should have the same explanation. An assumption of this test is that two models are functionally equivalent \textit{only if} they have the same reasoning process \cite{jacovi_towards_2020}. If this assumption holds, when a method provides different explanations for functionally equivalent models, it is unfaithful. However, we argue that this assumption may not be grounded. 
% First, it treats the model as a black-box, only considering its input and output. But the reason why we need explanations is that models should not be black-boxes. Second, 
There do exist functionally equivalent models that rely on entirely different reasoning mechanisms, a trivial example of which is various sorting algorithms. The assumption that all of them have the same explanation is counter-intuitive. Therefore, we do not believe that Implementation Invariance is a necessary condition for faithfulness.

Another form of Model Sensitivity is that an explanation should be sensitive to meaningful changes in the model that influence the prediction \cite{adebayo_sanity_2018}. For instance, when model weights are randomized, the explanation should also change. We consider this as a more sensible check than Implementation Invariance because of the reason above.\\

\noindent \textbf{Input Sensitivity.} This means that an explanation should be sensitive (resp. insensitive) to changes in the input that influence (resp. do not influence) the model prediction. We refer the reader back to Section~\ref{section:principles}, where we elaborate on how previous studies instantiate the notion differently \cite{sundararajan_axiomatic_2017, kindermans_learning_2017, adebayo_sanity_2018}. In our view, all of them are reasonable tests of necessary conditions for faithfulness.\\

\noindent \textbf{Polarity Consistency.} This axiom measures the consistency between importance-score-based explanations and the impact polarity on model predictions \cite{liu_rethinking_2022}. For example, if an explanation method assigns a \textit{positive} weight to a feature as its contribution to some predicted label, then after removing this feature, the model confidence in the label should be \textit{suppressed}. More generally, the sign of importance scores should agree with the polarity of the confidence change. We consider this as a valid sanity check, with one caveat: removing features may create nonsensical inputs, which decrease model prediction confidence solely because the inputs are out of distribution (OOD), but not because the removed features are important. We will come back to this issue when discussing perturbation-based evaluation.\\

\noindent \textbf{Robustness Equivalence \& Feature Importance Agreement.} Both axioms are proposed by \citet{wiegreffe_measuring_2021} specifically to measure the association between model-predicted labels and free-text rationales. Robustness Equivalence means that the explanation and the predicted label should be equally robust (or non-robust) under noise. Feature Importance Agreement means that input tokens that are important (resp. unimportant) for label prediction should also be important (resp. unimportant) for explanation generation. We consider both as reasonable sanity checks. However, one concern with the latter axiom is that in the implementation of the test, token importance is computed with gradient-based methods, many of which are themselves known to be unreliable in terms of faithfulness.\footnote{See Section~\ref{section:backprop_methods} for details.} In other words, the Feature Importance Agreement test needs to rely on another explanation method to provide feature importance, and assumes that the method is faithful.\\

Note that all the above axioms only test for necessary instead of sufficient conditions of faithfulness. Passing all tests does not guarantee that an explanation is faithful. Therefore, axiomatic evaluation can only be used to disprove faithfulness via counterexamples. Still, practically speaking, the more tests an explanation method passes, the more confidence we can have in its faithfulness.

\subsubsection{Predictive Power Evaluation}\leavevmode\newline

This type of evaluation uses the explanation to predict model decisions (instead of gold labels) on unseen examples, and considers a higher accuracy as an indicator of higher faithfulness. This accuracy is also called \textit{simulatability} in the literature. Intuitively, the more faithful an explanation is, the more information it should contain about the model's decision mechanism, and thus the easier it would be for an external observer, or \textit{simulator}, to predict the model's behavior based on the explanation. In practice, people often treat the \textit{difference} in simulatability with and without the explanation as a measure of its faithfulness.
The underlying assumption is that if an explanation leads to a different prediction than that made by the model it explains, then it is unfaithful \cite{jacovi_towards_2020}. 

To implement this evaluation, we need a way to derive the prediction from a given explanation. In existing work, predictions are derived either directly from the explanation itself or with a simulator, which can be an external model or humans. 

To derive the prediction directly from the explanation, the explanation should be an executable model itself capable of making predictions (e.g., decision trees or rule lists \cite{sushil_patient_2018}). Alternatively, it should be possible to define simple rules on top of the explanation to predict the label. For example, \citet{ye_connecting_2021} compute a scalar (say, sum) on top of the importance scores assigned to some feature, and compare the scalar to a pre-defined threshold to predict the label. Another example is given by \citet{sia_logical_2022}, who derive the prediction from the explanation using logical rules, specifically in the context of NLI tasks.

Using an external model as the simulator is similar to the idea of model distillation, where a student model (often smaller or simpler) learns to simulate the behavior of the original teacher model (the model to be explained) \cite{li_evaluating_2020, hase_leakage-adjusted_2020, pruthi_evaluating_2022}. Here, the goal is not to obtain a good student model, but to evaluate the quality of given explanations in terms of how well they can help the student simulate the teacher.

Instead of external models, humans can also be considered as the simulator, since they are the eventual target audience. In this evaluation, human subjects are asked to simulate the model's decision on new examples without access to the model itself, but only to the input and the explanation \cite{doshi-velez_towards_2017, ribeiro_anchors_2018, chen_learning_2018, hase_evaluating_2020, zheng_irrationality_2021}.

In our opinion, the first way (directly deriving the prediction from the explanation) is the most rigorous, in the sense that there is no external simulator as a potential confounding factor. However, it can be hard to generate large-scale and high-quality counterfactual examples to test simulatability. Existing work either create them manually, which limits the scale of evaluation \cite{ye_connecting_2021}, or automatically generates them with logical rules, which instead constrains the task scope \cite{sia_logical_2022}.

When there is an external simulator, \textit{label leakage} becomes an issue. This means that an explanation can trivially leak the label to the simulator with superficial regularities. For example, in the task of NLI, if a free-text explanation contains ``not'', then the label is very likely \texttt{contradiction}. Therefore, simulatability may favor explanations that contain trivial cues to the label but are otherwise uninformative: As an extreme example, the explanation can be the label itself.

When an external model serves as the simulator, such leakage can be even harder to avoid, since the explanation and the model simulator can communicate in numerous human-imperceptible ways (e.g., a punctuation mark can be the cue for some class). In addition, it is also questionable how expressive the simulator model should be. If too expressive, the proxy model can easily learn the label itself, and thus there will be little difference in simulatability with and without the explanation. This is analogous to the expressivity concern of probing classifiers expressed by \citet{hewitt_designing_2019}.

When the simulator is human, it is easier to control for leakage by filtering out visible cues. However, there are several other concerns. First, this evaluation mingles plausibility with faithfulness. If humans fail to simulate model predictions, then it could be that \textit{either} the explanation is not plausible (to them) \textit{or} that the explanation is unfaithful. Moreover, when simulating the model's prediction, it is difficult to ensure that humans can eliminate their own judgments of what the gold label should be.

Note that label leakage does not mean that an explanation is necessarily bad. Oftentimes, explanations provided by humans also unavoidably leak the decision. For instance, how would we explain that a premise contradicts a hypothesis without using any negation-related expression at all? This means that we should not blindly reject all label-leaking explanations. Recent studies provide several alternatives. \citet{hase_leakage-adjusted_2020} propose a variant of simulatability called Leakage-Adjusted Simulatability (LAS), which performs a macro-average of leaking and non-leaking explanations. However, this is effectively reduced to vanilla simulatability when most or all explanations are leaking. \citet{pruthi_evaluating_2022} argue that to avoid leakage, explanations should only be provided to the simulator at \textit{training} time but not at \textit{test} time. In their evaluation protocol, during training, the simulator learns to predict the model's decision, and explanations are provided as a multitask learning objective or attention regularization. Then, during inference, the simulator is expected to generalize to unseen examples without explanations. As a third alternative, \citet{chen_rev_2022} propose to measure the additional information provided by an explanation \textit{beyond} what is already in the input or the label, with a metric called REV (Rationale Evaluation with conditional V-information). Suppose we are trying to explain why a model answers ``enjoy nature'' given the question ``why do people go hiking?''. Under this metric, trivial explanations like ``people go hiking to enjoy nature'' will be penalized, although they result in high simulatability; whereas those like ``hiking means the activity of going for long walks, especially across the country, or in nature ...'' will be favored, as they provide extra information. 

To summarize, we believe that the predictive power evaluation is a sensible test for faithfulness. However, when using an external simulator, we should take into account the label leakage issue. In addition, experimental design with human simulators should be cautious of the above-mentioned pitfalls.

\subsubsection{Robustness Evaluation} \leavevmode\newline

Robustness evaluation measures if the explanation is stable against subtle changes in the input examples, such as pixel perturbations that result in images that are indistinguishable from each other. In its earliest version, robustness requires that similar inputs (as perceived by humans) should have similar explanations \cite{alvarez-melis_robustness_2018}. However, this does not rule out the possibility that the model itself can be non-robust to subtle input perturbations. Later work remedies this flaw by imposing constraints on model predictions. Now, robustness means that for similar inputs that have similar model outputs, the explanations should be similar \cite{ghorbani_interpretation_2019, yeh_fidelity_2019, ding_evaluating_2021, zheng_irrationality_2021, wang_fine-grained_2022, yin_sensitivity_2022}. The underlying assumption is that on similar inputs, the model makes similar predictions \textit{only if} the reasoning process is similar \cite{jacovi_towards_2020}. 

We identify two problems with this evaluation. First, though the notion of ``indistinguishable inputs'' makes sense in vision, it is hard to apply in NLP since the input space is discrete. If we substitute tokens with semantically similar ones, we do not know whether their model representations are also indistinguishable. If we instead directly add a noise term to the model representation, then it may create OOD tokens that do not map to anything in the vocabulary. Second, the assumption mentioned in the above paragraph is questionable. Even though the inputs and outputs are similar, the model's reasoning mechanism can still differ. As a simple example, consider two similar cat images, which differ only in the length of the cat's fur. We observe that the model predicts \texttt{cat} for both images with similar confidence. Now, it is still possible that the model is relying on different features (e.g., body shape in the first image, and ear shape in the second) in the two predictions. There is theoretically nothing preventing the model from doing so. \citet{ju_logic_2022} provide empirical evidence for this in realistic scenarios, demonstrating that rationale-based models can predict the same label for semantically similar inputs, yet use different reasoning. Thus, if we observe that an explanation is not robust, we cannot conclude if it is because the explanation is unfaithful \textit{or} because the model's reasoning mechanism is indeed not robust.

As a result, we do not recommend robustness as a good evaluation metric for faithfulness, but this does not mean that it is without value. Robustness may be a necessary metric for user understandability, since stable explanations are easier for the human audience to comprehend than those that change drastically and unpredictably given similar inputs \cite{zhou_exsum_2022}. 

\subsubsection{Perturbation-based Evaluation} \leavevmode\newline

This kind of evaluation perturbs parts of the input and observes the change in the output. It differs from robustness evaluation in that robustness considers extremely similar inputs and expects that the explanation is similar; but in perturbation-based evaluation, we consider inputs that are not necessarily similar, and our expectation of how the explanation should change depends on which parts of the input are perturbed.

Concretely, consider an explanation in the form of feature importance scores. We now remove a fixed portion of features from the input, based on the explanation. If the most important features (as indicated by the explanation) are first removed, the model prediction is expected to change drastically. Conversely, removing the least important features should result in a smaller change. This type of evaluation has been widely adopted in both vision \cite{bach_pixel-wise_2015, samek_evaluating_2015, shrikumar_learning_2017, chen_learning_2018, samek_reliability_2019} and language \cite{arras_explaining_2016, chen_learning_2018, serrano_is_2019, jain_attention_2019, atanasova_diagnostic_2020}. In particular, a popular set of metrics are sufficiency and comprehensiveness \cite{deyoung_eraser_2020}. Sufficiency measures how much the model prediction can be recovered when only including the important features, while comprehensiveness measures to what extent the model prediction is suppressed by removing the important features.

One underlying assumption of this evaluation is that different parts of the input are \textit{independent} in their contribution to the output \cite{jacovi_towards_2020}. However, features can be correlated. When one feature is removed, we cannot guarantee that other features stay untouched. 

Another assumption is that the observed performance change does \textit{not} come from nonsensical inputs. When a feature is perturbed, the resulting input can become out-of-distribution (OOD). Compared to Computer Vision (CV), this has more serious consequences in NLP, since removing a word can make the sentence ungrammatical or meaningless, but removing a pixel almost does not change the semantics of an image. \citet{hooker_benchmark_2019} addresses the issue in CV by proposing the RemOve And Retrain (ROAR) benchmark. According to a given explanation method, the set of most important features is removed from \textit{both} the training and the testing data. The model is then trained and tested again on the new data, and a larger performance drop indicates higher faithfulness. In their experiments, image classification models are found to still achieve decent accuracy even after most input features (90\%) are removed. This indicates that the performance drop observed in previous evaluation approaches without retraining might indeed come from the distribution shift instead of the lack of important features. However, although ROAR ensures that the training and testing data come from the same distribution, it brings about a new problem -- the retrained model is no longer the same as the original model to be explained. Alternatively, \citet{yin_sensitivity_2022} address the OOD issue in NLP by performing small but adversarial perturbations on the feature embeddings, instead of removing or replacing the feature altogether. Intuitively, if the perturbation scale is constant, the model should be more sensitive to perturbations on important tokens than those on unimportant ones. One pitfall we notice is that the \textit{density} of neighboring token vectors in the embedding space might be non-uniform for different tokens. Concretely, suppose token $x_1$ lies in a region with 100 neighboring tokens within distance $d$, while token $x_2$ only has 10 neighbors within the same distance. Then, when we perform a perturbation of the same scale (e.g., in terms of norm) on both tokens, it is more likely to result in an OOD input for $x_2$.

The final and most fundamental assumption is that perturbation-based evaluation implicitly regards another explanation method -- a perturbation-based one, which we will discuss in Section~\ref{section:counterfactual_intervention} -- as the \textit{ground truth}. For example, as \citet{ju_logic_2022} point out, AOPC (Area Over the Perturbation Curve), a perturbation-based evaluation metric, can be reduced to a basic explanation method, leave-one-out \cite{li_visualizing_2016} under constraints on the number of features.

In short, while perturbation-based evaluation has been widely used, we should be cautious about the above-mentioned assumptions and their consequences, especially since there is still no perfect fix in NLP yet.

\subsubsection{White-box Evaluation} \leavevmode\newline

This type of faithfulness evaluation relies on known ground-truth explanations, against which a candidate explanation can be compared. The ground-truth explanations come from either \textbf{transparent tasks} or \textbf{transparent models}. 

In transparent tasks, the data is created in a way such that the set of informative features is controlled. One way to do this is via synthetic tasks, like reconstructing a simple function \cite{chen_learning_2018, hooker_benchmark_2019}, counting and comparing the number of digits \cite{de_cao_how_2020, hase_when_2022}, or text classification on hybrid documents where only one sentence is relevant to the label \cite{poerner_evaluating_2018}. A less artificial way is to modify existing datasets of natural tasks, such that models trained on them have to use the desired feature(s) in order to achieve high performance. This method is first proposed in the vision domain \cite{yang_benchmarking_2019, adebayo_debugging_2020}, and then adapted to NLP. Specifically, \citet{zhou_feature_2021} and \citet{bastings_will_2021} concurrently explore the idea of using spurious features to evaluate faithfulness. Given a natural dataset, the first step is to de-correlate all the original features with the label, e.g., by randomly re-assigning the label for all examples. The second step is to inject the spurious feature in the data, such that it perfectly correlates with the new label. For example, for all the reviews with a \texttt{positive} sentiment, we change every article in them to ``the''; otherwise, we change the articles to ``a''.  In this way, \textit{only} the article feature is (perfectly) indicative of the label, so any model that performs better than random on the dataset must have used this feature. Finally, to evaluate an explanation method, we measure the extent to which it can correctly identify the spurious feature as the contributor for models that achieve perfect performance. 
%Since the informative features are controlled, any model that performs well on the tasks must have relied on these features. 
% Therefore, the ground-truth explanation is known.

Transparent models provide another way to obtain ground-truth explanations. They are inherently interpretable models, e.g., Logistic Regression or Decision Tree. The ground-truth explanation of important features can be directly obtained through their weights or prediction structure \cite{ribeiro_why_2016, natesan_ramamurthy_model_2020}. To evaluate an explanation method, one can apply it to these transparent models and compare the explanation with the ground-truth feature importance.

% Thus, one can run an explanation method with the transparent task or the transparent model, and compare the resulting explanation with the ground truth. If they are clearly different, then the explanation method is unfaithful. 

Note that this test is still largely a sanity check, constituting a necessary instead of sufficient condition for faithfulness. Since the transparent setups are simplified, passing the white-box test does not guarantee that the explanation method can faithfully generalize to real-world scenarios.

\subsubsection{Human Perception Evaluation} \leavevmode\newline
\label{section:human_perception_evaluation}

Human perception evaluation assesses if the model explanation matches human intuition when they make predictions on the same task. For example, if the explanation is in the form of feature importance scores, to what extent does it align with human-annotated importance scores \cite{feng_pathologies_2018, deyoung_eraser_2020, clinciu_study_2021}? 

However, many previous studies of this type do not report which principle is evaluated. We argue that such tests only evaluate plausibility. For them to touch on faithfulness, we need to make the assumption that models reason in the same way as humans do. Obviously, this does not always hold; otherwise, we will not need explanations at all. As said at the beginning of Section~\ref{section:faithfulness_evaluation}, faithfulness evaluation should not involve human judgment on the explanation quality \cite{jacovi_towards_2020}. Again, this is not to say that human evaluation has no value. Similar to robustness, human perception is a crucial evaluation to perform since our ultimate goal is for the target audience to better understand the model \cite{zhou_exsum_2022}.

\subsubsection{Meta-evaluation of Faithfulness Evaluation Metrics} \leavevmode\newline

To compare the quality of different faithfulness metrics, there are several meta-evaluation standards.

One such standard is the ability to detect known unfaithful explanations. This means that a good evaluation metric should be able to distinguish explanations that are known to be unfaithful from ``relatively faithful'' ones. In practice, these unfaithful explanations are often randomly generated \cite{chan_comparative_2022} or artificially created \cite{sia_logical_2022}. It is less trivial to come up with those allegedly ``relatively faithful'' explanations, though; existing work typically uses its proposed explanation method or other widely used methods. But after all, if we know in advance which explanations are relatively more faithful, we would not have needed faithfulness evaluation metrics, let alone another evaluation metric of these metrics.

Another standard is ``solvability''. If an evaluation metric can be treated as an objective and an optimal explanation can be found algorithmically, the metric is said to be \textit{solvable} \cite{zhou_solvability_2022}. For example, it is demonstrated that two widely-used perturbation-based evaluation metrics, sufficiency and comprehensiveness, are solvable with beam-search.\footnote{This does not mean that they are bad metrics, since the search-based explainer also performs decently on several other metrics, so this explainer could be at least a strong baseline to compare explanation methods against.}

The third standard is time complexity. Specifically, \citet{chan_comparative_2022} propose to measure the average number of model forward passes needed to compute an evaluation metric, and a smaller number is favored as it requires fewer computational resources. 

\subsubsection{Summary} \leavevmode\newline

Based on the above discussion, we recommend axiomatic evaluation, predictive power evaluation (no simulator, or with an external model as the simulator), and white-box evaluation, as desired faithfulness evaluation methods, with caveats specified before. More ideally, more than one of these evaluations should be done, since many of them only test a necessary condition of faithfulness.

To complement the list of design principles provided by \citet{jacovi_towards_2020}, we additionally propose a few more towards a better evaluation of faithfulness.

We believe it is important to \textbf{define faithfulness in advance rather than ad-hoc.} 
Instead of using the same term to refer to different things, a clear definition at the beginning of the evaluation will greatly benefit comparability.

It is also crucial to \textbf{state the assumptions of the evaluation, where they do not hold and the resulting implications.} Especially, we caution against making assumptions about how models reason, e.g., ``they reason in the same way as humans do'', as this conflates plausibility with faithfulness.

In addition, it is helpful to \textbf{disentangle the capacity of the model and the quality of the explanation.} For example, a non-robust explanation can result from either the model relying on inconsistent features or the explanation being unfaithful, as mentioned in the discussion of robustness evaluation.

Finally, how to evaluate the quality of evaluation metrics is still an under-studied problem. So far, the only study that systematically compares a set of metrics is \citet{chan_comparative_2022}. We call for more attention to this fundamental problem, so as to deepen our understanding of faithfulness and measure the progress in interpretability.
\section{Families of Model Explanation Methods}
\label{section:attempts}

We group existing explanation methods in pursuit of faithfulness into five categories: similarity-based methods, analysis of model-internal structures, backpropagation-based methods, counterfactual intervention, and self-explanatory models. To give the reader a quick overview, we briefly explain the intuition behind each type of method with a motivating example. 

Consider a Sentiment Analysis task, where a model determines if the sentiment of a given piece of text (e.g., product/movie review) is \texttt{positive}, \texttt{negative}, or \texttt{neutral}. An example input can be: 

\begin{extract}
\textit{This movie is great. I love it.}
\end{extract}

Suppose the model prediction is \texttt{positive}, which matches the ground truth. Our goal, now, is to explain why the model makes such a prediction. Here is how each method answers the question on a high level:

Similarity-based methods provide explanations in terms of previously seen examples, similar to how humans justify their actions by analogy. Specifically, they identify training instances or concepts\footnote{See Section~\ref{section:properties}.} that are similar to the current test example in the model's induced representation space (e.g., \textit{The movie is awesome}, \textit{The TV show is great}) as an explanation, assuming that the reasoning mechanism for similar examples is intrinsically similar. 

Analysis of model-internal structures examines the activation patterns of nodes, layers, or other model-specific mechanisms (e.g., attention \cite{bahdanau_neural_2015}), and derives an explanation with techniques like visualization, clustering, correlation analysis, etc. For example, on a visualized heat map of attention weights, \textit{great} and \textit{love} may have the highest weight among all token positions in the final layer. This can be interpreted as these two tokens contributing the most to the prediction.

Backpropagation-based methods compute the gradient (or some variation of it) of the model prediction with respect to each feature (e.g., input token). Features with the largest absolute gradient value (say, \textit{great} and \textit{love}) are then considered most important to the prediction. The intuition behind this is that even a slight change in these features\footnote{For discrete inputs like tokens, the extent of the change is defined in terms of similarity metrics in the embedding vector space.} could have resulted in a large change in the model output. For example, if \textit{great} becomes \textit{good} and \textit{love} becomes \textit{like}, the model's confidence of \texttt{positive} will probably not be as high.

Counterfactual intervention perturbs a specific feature (e.g., input token) while controlling for other features and observes the resulting influence in the model prediction. For example, to test if the word \textit{great} is important for the model prediction, we can mask it out or replace it with another word (e.g., \textit{OK}) and see how the model prediction changes. If the probability of the \texttt{positive} class decreases considerably, then the word \textit{great} has been an important feature for the original prediction.

Self-explanatory models do not rely on post-hoc explanation methods but provide explanations as a byproduct of the inference. For example, a self-explanatory model can be trained to predict the sentiment label (\texttt{positive}) and justify its prediction at the same time, by producing a natural language explanation (``The words \textit{great} and \textit{love} indicate that the person feels positive about the movie''). 

Next, we introduce each family of methods in detail, discussing their representative studies, strengths, and weaknesses.

\subsection{Similarity-based Methods}
\label{section:similarity_methods}

Similarity-based methods provide explanations in terms of training examples. Specifically, in order to explain the model prediction on a test example, they find its most similar\footnote{In practice, commonly used similarity metrics include cosine, Euclidean, etc.} training examples in the learned representation space, as support for the current prediction. This is akin to how humans explain their actions by analogy, e.g., doctors make diagnoses based on past cases.

\citet{caruana_case-based_1999} theoretically formalize the earliest approach of this kind, named ``case-based explanation''. Based on the learned hidden activations of the trained model, it finds the test example's k-Nearest Neighbors (kNN) in the training set as an explanation. Note that the similarity is defined in terms of the  \textit{model's learned representation space} but not the \textit{input feature space}, since otherwise the explanation would be model-independent.

\citet{wallace_interpreting_2018} also use the kNN search algorithm; but instead of deriving a post-hoc explanation, they replace the model's final softmax classifier with a kNN classifier at test time. Concretely, during training, the model architecture is unmodified. Then, with the trained parameters fixed, each training example is passed through the model again, and their representations are saved. The inference is done with a modified architecture: a test example is classified based on the labels of its kNNs from the training examples in the learned representation space. However, the resulting explanations are only evaluated based on whether the explanations align with human perception of feature importance on qualitative examples, which is irrelevant to faithfulness. 

\citet{rajagopal_selfexplain_2021} introduce a self-explanatory classification model, where one component, the ``global interpretable layer'', also uses the idea of similarity-based explanation. This layer essentially identifies the most similar \textit{concepts} (phrases in this case) in the training data for a given test example. Their approach is mainly evaluated in terms of plausibility, i.e., how adequate/understandable/trustworthy an explanation is based on human judgment. One metric touches on faithfulness --- human simulatability --- however, the authors only report the relative difference with and without the explanation instead of the absolute scores, which makes it hard to determine how faithful the approach is. 

\subsubsection{Strengths and Weaknesses}\leavevmode\newline

Similarity-based methods exhibit several strengths.
First, they are \textbf{intuitive} to humans since the justification by analogy paradigm has long been established. 
Second, they also \textbf{easy to implement}, as no re-training or data manipulation is needed. The similarity scores are available by simply passing examples through the trained model to obtain the model's representation of them. 
Third, they are \textbf{highly model-agnostic}, since all kinds of Neural Networks have a representation space, and any similarity metric (cosine, Euclidean, etc.) can be easily applied. 
Finally, certain similarity-based explanations are rated by human subjects as more \textbf{understandable and trustworthy} \cite{rajagopal_selfexplain_2021} compared to several other baselines in the families of backpropagation-based methods and counterfactual intervention.\\

Nevertheless, there are also several weaknesses:
First, most similarity-based methods only provide the user with the \textbf{outcome} of the model's reasoning process (i.e., which examples are similar in the learned space), but do not shed light on \textit{how} the model reasons (i.e., how the space is learned) \cite{caruana_case-based_1999}. 
Second, existing work mostly evaluates the resulting explanations with plausibility-related metrics, including adequacy, relevance, understandability, etc., with human judgments. But as mentioned in Section~\ref{section:faithfulness}, plausibility does not entail \textbf{faithfulness}. Thus, it is questionable whether similarity-based methods can truly establish causality between model predictions and the explanation.
Additionally, the space of exploration is \textbf{confined to the training set}. This inherently limits the diversity and scope of the explanation, potentially leaving certain edge cases unexplained. It also implies that the behavior of similarity-based methods depends on the distribution of the training data. In other words, the explanation outcomes may vary considerably if the training set does not well represent the broader data distribution or is biased in some manner.
Finally, similarity-based methods inherently offer \textbf{instance-level explanations} and do not provide insights into the feature-level contribution to the prediction. This lack of granularity can limit the potential for actionable insights.

\subsection{Analysis of Model-Internal Structures}
\label{section:analysis_structure}

The analysis of model-internal structures (e.g., individual neurons, or specific mechanisms like convolution or attention \cite{bahdanau_neural_2015}) is believed to shed light on the inner workings of NLP models. Common analysis techniques include visualization (e.g., activation heatmaps, information flow) \cite{vig_visualizing_2019}, clustering (e.g., neurons with similar functions, inputs with similar activation patterns) \cite{brunner_natural_2018}, and correlation analysis (e.g., between neuron activations and linguistic properties) \cite{qian_analyzing_2016}.

We categorize previous research by the target of the analysis, \textbf{individual neurons} or the \textbf{attention mechanism}, as two main lines of work in this area. We focus on attention among all mechanisms, because it has become the basis of the most widely-adopted architectures in NLP systems nowadays and has significantly reshaped this line of work ever since.

\subsubsection{Analysis on Neurons} \leavevmode \newline

\begin{figure}[!t]
  \centering
  \includegraphics[width=\textwidth]{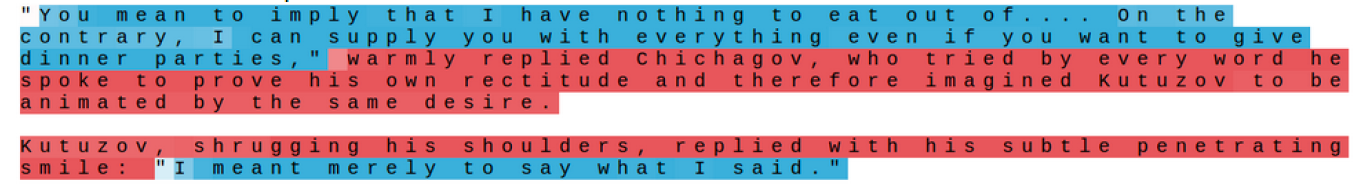}
  \caption{A neuron that ``turns on'' inside quotes (figure from \citet{karpathy_visualizing_2015}). Blue/red indicates positive/negative activations respectively, and a darker shade indicates larger magnitude.}\label{figure:karpathy_quotes}
\end{figure}

The initial success of neural models in NLP sparked interest in finding interpretable functions of individual neurons. 
\citet{karpathy_visualizing_2015} examine the activation patterns of neurons in a character-level LSTM language model. They find neurons with specific purposes, e.g., one that activates within quotes, inside if-statements, or toward the end of a line, respectively (Figure~\ref{figure:karpathy_quotes}). 

Using a similar approach, \citet{li_visualizing_2016} and \citet{strobelt_lstmvis_2018} then find individual LSTM neurons that specifically activate for certain compositional structures in language, such as negation, intensification, and adjective-noun composition. 
\citet{poerner_interpretable_2018} and \citet{hiebert_interpreting_2018} take the reverse direction: instead of analyzing which neurons fire for a given input pattern, they look for inputs that have similar neuron activations. Preliminary observations show that Gated Recurrent Units (GRU) and LSTM models can capture certain orthographic and grammatical patterns.

\subsubsection{Analysis on Attention Mechanism} \leavevmode \newline

Transformers \cite{vaswani_attention_2017} has become the foundation of the State-of-the-Art (SOTA) systems on many NLP tasks \cite{devlin_bert_2019, liu_roberta_2019, clark_electra_2020, brown_language_2020, raffel_exploring_2020, openai_gpt-4_2023}. The core of Transformers is an attention mechanism, called \textbf{self-attention}. Simply put, self-attention is a function that takes in a sequence of vectors $X=\langle x_1, x_2, ..., x_n \rangle$ (each $x_i \in \mathbb{R}^d$) as input and returns another sequence of vectors $Y=\langle y_1, y_2, ..., y_n \rangle$ (each $y_i \in \mathbb{R}^d$) of the same length. 
Each $y_i$ is a weighted average of a transformed version of all $x_i$'s, i.e., $y_j = \sum_{i=1}^{n} a_{ij} f(x_i)$, where $f: \mathbb{R}^d \rightarrow \mathbb{R}^d$ is an affine transformation. These weights $a_{ij}$ are called \textbf{attention weights}, intuitively representing how much the model ``attends to'' each input vector when computing the weighted average. In NLP Transformer models, we can think of the initial $X$ as token embeddings, i.e., each $x_i$ is a vector representation of a token in the input. Then, each $y_i$ can be viewed as a composite embedding. 

\begin{figure}[!tbp]
  \centering
  % \subfloat[A BERT sentiment classification model. The class label is predicted with the embedding of CLS (a special token for sentence classification tasks) in the final layer.]{
  %   \includegraphics[width=0.6\textwidth]{imgs/bert_classification.jpeg}
  % }
  % \hfill
  % \subfloat[Attention weights from CLS to all tokens. Each color represents an attention head. Lines represent averaged attention weights over all heads. Darker shades indicate higher weights.]
  \includegraphics[width=0.35\textwidth]{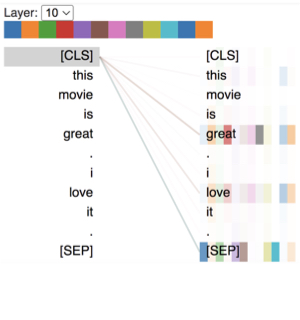}
\caption{
% A BERT sentiment classification model (left, adapted from \cite{devlin_bert_2019}) and its attention weights (right, created using BertViz \cite{vig_visualizing_2019}). 
% Given the input \textit{This movie is great. I love it.}, the model predicts \texttt{positive} using CLS. In the penultimate layer, \textit{great} and \textit{love} are among the tokens with highest attention weight from CLS.
Attention weights from [CLS] to all tokens in a BERT sentiment classification model, created using BertViz \cite{vig_visualizing_2019}. Each color represents an attention head. Lines represent averaged attention weights over all heads. Darker shades indicate higher weights.}
\label{figure:bert_attention}
\end{figure}

Given this structure, it may be tempting to interpret attention weights as the importance of input tokens to the output. Consider our running example: \textit{This movie is great. I love it}. A BERT-based classifier makes a prediction by passing the input tokens through a stack of 12 Transformer blocks, and then predicting the label \texttt{positive} using the representation of [CLS] (a special token for sentence classification tasks). Figure~\ref{figure:bert_attention} shows the attention weights from [CLS] to all tokens in the penultimate layer. Among all non-special input tokens, \textit{great} and \textit{love} receive the highest averaged weights over all heads. Intuitively, this is often understood as that they are the most important tokens for the prediction. This type of understanding has been used (implicitly or explicitly) as evidence for model interpretability in different tasks and domains, such as text classification \cite{martins_softmax_2016}, knowledge base induction \cite{xie_interpretable_2017}, and medical code prediction \cite{mullenbach_explainable_2018}. 

\begin{figure}[t]
  \centering
  \includegraphics[width=0.6\textwidth]{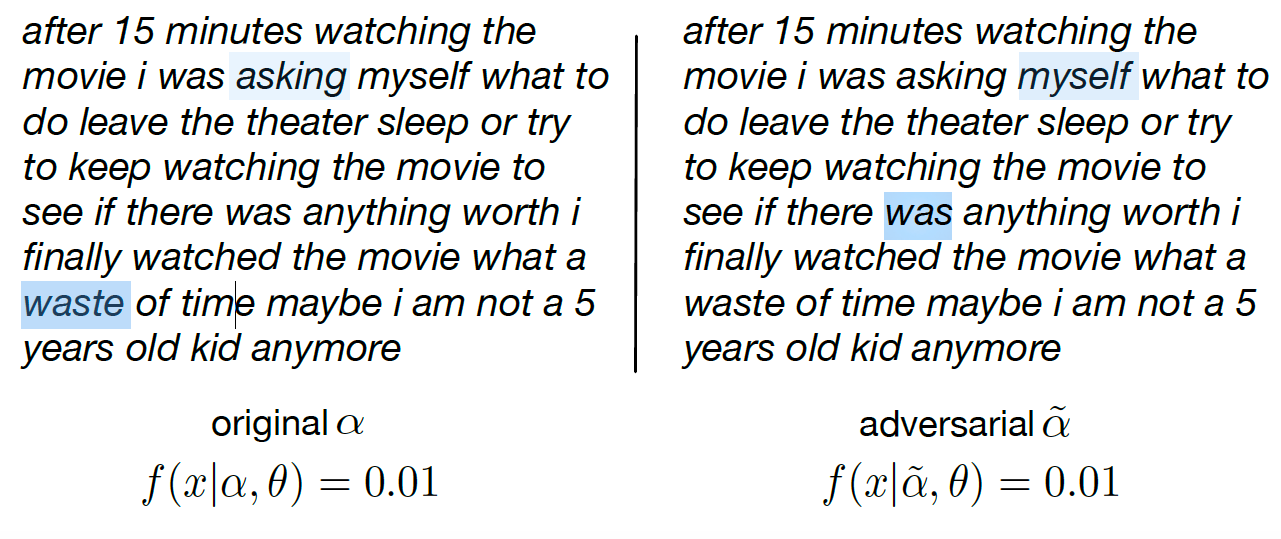}
  \caption{A sentiment analysis model's attention distribution over words in a negative movie review (figure from \citet{jain_attention_2019}). 
  The left part of the figure shows the observed attention weights, and the right part an adversarially constructed set of attention weights while controlling for all other parameters. Despite being quite dissimilar, they yield effectively the same prediction.  }\label{figure:jain_wallace_adversarial}
  \vspace{-0.15in}
\end{figure}

Nevertheless, there has been a long debate on whether attention constitutes a faithful model explanation. In their work ``Attention is Not Explanation'', \citet{jain_attention_2019} find that it is possible to construct an adversarial attention distribution, i.e., one that is maximally different from the original distribution but has little influence on the model output. For example, in Figure~\ref{figure:jain_wallace_adversarial}, a sentiment classification model predicts that a movie review is \texttt{negative}. Original attention weights suggest that \textit{waste} and \textit{asking} appear the most important, but the adversarial distribution shifts the model's attention onto uninformative words like \textit{myself} and \textit{was}, without changing its prediction. This indicates that attention weights do not always causally influence the prediction.

As a direct response, \citet{wiegreffe_attention_2019} offer a direct counter-argument in their paper titled ``Attention is \textbf{Not Not} Explanation'', emphasizing that 
adversarial distributions are not adversarial weights. The adversarial attention distributions are artificially constructed by humans, but not learned by models through training. In other words, a trained model would probably not naturally attend most to uninformative words like \textit{myself} and \textit{was}. In fact, even when the authors try to guide the model's attention towards such uninformative words using specially designed objectives, they seldom converge to these adversarial distributions after training.

\citet{pruthi_learning_2020} again refute their argument, showing that with a new training objective, they successfully guide the model to learn intended adversarial attention distributions. For example, when predicting the occupation of a person in the text, the model is trained to assign minimal attention weights to gender indicator tokens (e.g., ``he'' and ``she''), while it is still \textit{using} this signal for prediction. This implies that attention weights can be deceiving, i.e., a human user might find the model trustworthy since it is seemingly not relying on gender biases, yet it still does under the hood.

Nevertheless, in favor of the use of attention as explanation, other researchers argue that existing criticisms mainly target sequence classification tasks (e.g., Sentiment Analysis), but sequence-to-sequence tasks (e.g., MT) may be a more suitable use case for attention as explanation. 
It is found that modifying the \textbf{encoder-decoder} attention heads \textit{does} influence MT model generations substantially \cite{voita_analyzing_2019, vashishth_attention_2019}, in contrast to the findings of \citet{jain_attention_2019} on single-sequence tasks. Other attention heads, such as the encoder- or decoder- only ones, are less influential \cite{voita_analyzing_2019, raganato_fixed_2020}.

Concluding the debate, \citet{bastings_elephant_2020} still argue against attention and in favor of saliency methods\footnote{These include backpropagation-based methods (Section~\ref{section:backprop_methods}) and counterfactual intervention (\ref{section:counterfactual_intervention}) in our terms.} as faithful explanations. However, they acknowledge that understanding the role of the attention mechanism is still a valuable scientific question (e.g., what linguistic information it captures; which heads can be pruned without performance loss).

A natural question to ask next is: (how) can attention become a more faithful model explanation? To answer it, we first need to understand what might have caused attention to be \textit{unfaithful}. We summarize three potential factors:

(a) \textbf{Information mixing}: Attention weights are assigned to \textbf{hidden states} (in intermediate layers) instead of \textbf{input features} (in the initial layer), but we nevertheless interpret attention weights as the importance of the corresponding input feature. In fact, hidden states have already mixed in information from other input features \cite{tutek_staying_2020}. 

(b) \textbf{Locality}: Existing methods mostly focus on attention weights in a single layer (often the final) from a single position (often the special token [CLS]), but this fails to capture the big picture of how the information flows globally \cite[][i.a.]{pascual_telling_2021}.

(c) \textbf{Intrinsic lack of causality}: Attention weights are simply not designed to have any causal connection with the prediction, and thus cannot provide a faithful explanation alone, but need to be tied to other explanation methods \cite[][i.a.]{mylonas_attention_2022}.

Recent studies have started exploring ways to remedy each of these potential flaws in order to make attention more faithful. 

To address the issue of (a) information mixing, \citet{tutek_staying_2020} introduce two regularization techniques. The first is weight tying, which minimizes the distance between intermediate hidden states and their corresponding input features. The second is an auxiliary Masked Language Modeling (MLM) task, which decodes input representations from their corresponding hidden state. Both techniques aim to make hidden states more representative of input representations. Experiments show that they effectively increase the causal influence of attention modification on model predictions, thus improving faithfulness.

In response to (b) the locality issue, one solution is to characterize the information flow through the entire network, instead of only considering a single layer or token position. The earliest attempt is made by \citet{abnar_quantifying_2020}, who propose \textbf{Attention Rollout} and \textbf{Attention Flow}.
These are empirically shown to be more plausible compared to raw attention weights (evaluated with human perception), as well as claimed to be more faithful (in terms of correlation with gradient-based methods)\footnote{It is questionable if we can treat gradient-based methods as the gold standard for faithfulness though, as discussed in Section~\ref{section:backprop_methods}.}. In addition, \citet{ethayarajh_attention_2021} theoretically prove that Attention Flow can be Shapley values\footnote{We will revisit this in Section~\ref{section:counterfactual_intervention}.} --- a counterfactual explanation with provable faithfulness guarantees --- under certain conditions, whereas raw attention weights cannot. However, \citet{liu_rethinking_2022} point out that Attention Rollout performs almost as badly as raw attention weights in the polarity consistency test,\footnote{See Section~\ref{section:faithfulness_evaluation} for details.} again casting doubt on its faithfulness. 

Finally, to tackle (c) the intrinsic lack of causality, a number of studies propose to tie attention to other explanation methods, especially backpropagation-based ones (Section~\ref{section:backprop_methods}). This includes  \citet{hao_self-attention_2021}, \citet{lu_uence_2021}, \citet{pascual_telling_2021}, and \citet{mylonas_attention_2022}, which differ mainly in the explanation method that attention is connected with. 

Despite the controversy on faithfulness, recent work in cognitive science shows that attention can uncover interesting similarities between how the human brain and LMs work. Attention distributions in Transformers are found to partially converge with the activation patterns of the human brain in masked word prediction \cite{caucheteux_brains_2022}. Similarly, they are predictive of human eye fixation patterns during task-reading to some extent \cite{eberle_transformer_2022}.

\subsubsection{Strengths and Weaknesses}\leavevmode\newline

To summarize, analysis of model-internal structures, as a family of explanation methods, has several strengths.
First, the visualization of model-internal structures is \textbf{intuitive and readable} to humans, especially end-users.
Second, there are many \textbf{interactive} tools (see Appendix~\ref{appendix:structure_analysis}), which can help the user form hypotheses about their data and models and dynamically adjust them through minimal testing.
Third, the attention mechanism can capture the \textbf{interaction} between features, whereas many other methods can only capture the influence of individual features themselves.
Finally, model weights are \textbf{easily accessible and computationally efficient}, compared to other methods. \\

However, these methods also suffer from several key weaknesses:
First, it is questionable to what extent raw attention weights represent \textbf{causal contribution}, as mentioned in the debate.
Second, the lack of faithfulness may be due to our interpretation of attention weights on intermediate \textbf{hidden states} as the importance of input features; however, hidden states have already mixed in contextual information through previous self-attention layers, and therefore may not be representative of input features.
Finally, existing methods often focus on attention weights in a single layer and/or from a single token position. This may reflect how much the model attends to each input position \textbf{locally}, but without taking the whole computation path into account. Methods that characterize the global information flow may be a better alternative \cite[][i.a.]{abnar_quantifying_2020}.
\subsection{Backpropagation-based Methods}
\label{section:backprop_methods}

Backpropagation-based methods aim to identify the contribution of input features via a \textit{backward pass} through the model, propagating the \textit{importance} (or \textit{relevance}, used interchangeably in the literature) attribution scores from the output layer to the input features. They can be further distinguished into two categories, \textbf{gradient methods} and \textbf{propagation methods}. Gradient methods follow standard backpropagation rules. In other words, they directly compute the \textit{gradient} (or some variant of it) of the output w.r.t the input features via the chain rule, assuming features with larger gradient values are more influential to the model prediction. By contrast, propagation methods define custom backpropagation rules for different layer types and compute the relevance scores layer by layer until reaching the input. This is believed to better capture the redistribution of relevance through special layers, such as ReLU.

Most ideas in this family have been first proposed in Computer Vision (CV). In the following subsection, we will explain their origin in vision and adaptation in language.

To formally synthesize existing work, we will use the following notation throughout the rest of this section: An example $x$ (e.g., an image or a sentence) has features $x_i, i \in \{1, 2, ...n \}$ (e.g., a pixel, a region, or a token). A model $M$ takes $x$ as input and predicts $y = M(x)$ as output. Our goal is to explain the \textit{relevance} of each feature $x_i$ to $y$, denoted by $r_i(x)$. For some specific methods, we also define a \textit{baseline input} $\overline{x}$ (e.g., an all-black image, or a sentence with all-zero token embeddings) against which $x$ is compared. We will discuss each subsequent method using this formalization. 

\subsubsection{Gradient methods} \leavevmode \newline

As their name suggests, gradient methods treat the gradient (or some variant of it) of the model output w.r.t. each input feature as its relative importance. The feature can typically be a pixel in vision and a token in language. Intuitively, the gradient represents how much difference a tiny change in the input will make to the output. This idea comes from generalized linear models (e.g., Logistic Regression), where each feature has a linear coefficient as their importance to the output. In the case of deep NNs, a natural analog of such coefficients would be gradients, as they characterize the marginal effect of a feature change on the output.

Using the notation above, the core difference of existing gradient methods lies in how they calculate $r_i(x)$, the relevance of feature $x_i$.\footnote{In Appendix~\ref{appendix:bp_methods}, Table~\ref{table:gradient_methods} summarizes the formal definition of $r_i(x)$ in each method, and Figure~\ref{figure:grad_methods_visual} presents a visualization of all methods in the vision domain.}

The most straightforward idea is to take the gradient itself (referred to as \textbf{Simple Gradients} or \textbf{Vanilla Gradients}), $\frac{\partial M(x)}{\partial x_i}$, as the feature relevance \cite{baehrens_how_2010, simonyan_deep_2014}. The \textit{sign} of the gradient represents whether the feature is contributing positively or negatively to the output, e.g., increasing or decreasing the probability of a certain class in a classification task. The \textit{magnitude} of the gradient stands for the extent to which the feature influences the output. 
Simple Gradients are easy to implement, intuitive to understand, and flexible for extension -- beyond individual input features, \citet{kim_interpretability_2018} extend it to high-level concepts (color, pattern, gender ...) in their method called TCAV. 
However, Simple Gradients have two apparent problems related to faithfulness. First, a function can be \textit{saturated}. Consider common neuron activation functions like sigmoid ($y = \frac{1}{1+e^{-x}}$) and tanh ($y = \frac{e^{x} - e^{-x}}{e^{x}+e^{-x}}$). When $x \rightarrow \pm\infty$, we have $\frac{dy}{dx} \rightarrow 0$. In other words, when the absolute value of an input feature is large enough, it has a very small gradient locally, although the feature may have a large contribution to the output $y$ globally. Second, gradient only measures the \textit{responsiveness} of the output w.r.t. the feature (how much the output changes in response to an infinitesimal change in the feature), but not the \textit{contribution} of the feature (how much the current feature value contributes to the output value) \cite{bastings_elephant_2020}. Say, in an image classification model, we can interpret gradients as ``how to make an image more (or less) a cat'', but not ``what makes the image a cat''. More formally, taking a simple linear model $y =  \sum_{i=1}^n w_i x_i $ as an example, the gradient $w_i$ measures the responsiveness while $w_i x_i$ measures the contribution. If $w_i$ is small but $x_i$ is very large, the proportion of $w_i x_i$ in $y$ can still be large. This cannot be captured by the gradient alone.

\textbf{Gradient$\times$Input} \cite{denil_extraction_2015} is proposed as a natural solution to the latter issue. It computes the relevance score of a feature as the dot product of the input feature and the gradient, $x_i \odot \frac{\partial M(x)}{\partial x_i}$, analogous to $w_i x_i$ in a linear model. Intuitively, this takes into account the feature value itself. In CV, Gradient$\times$Input empirically reduces noise in feature relevance visualizations.
However, this only improves plausibility, and is not necessarily related to faithfulness.
Also, Gradient$\times$Input fails the Input Sensitivity test for faithfulness (cf. Section~\ref{section:principles}), i.e., if two inputs differ only at feature $x_i$ and lead to different predictions, then $x_i$ should have non-zero relevance. As a simple counterexample, when the differing feature $x_i$ has a zero gradient in both inputs, the product of the input and the gradient would also be zero in both cases, which fails to capture the difference in their contribution.

\begin{figure}[!t]
  \centering
  \includegraphics[width=0.9\textwidth]{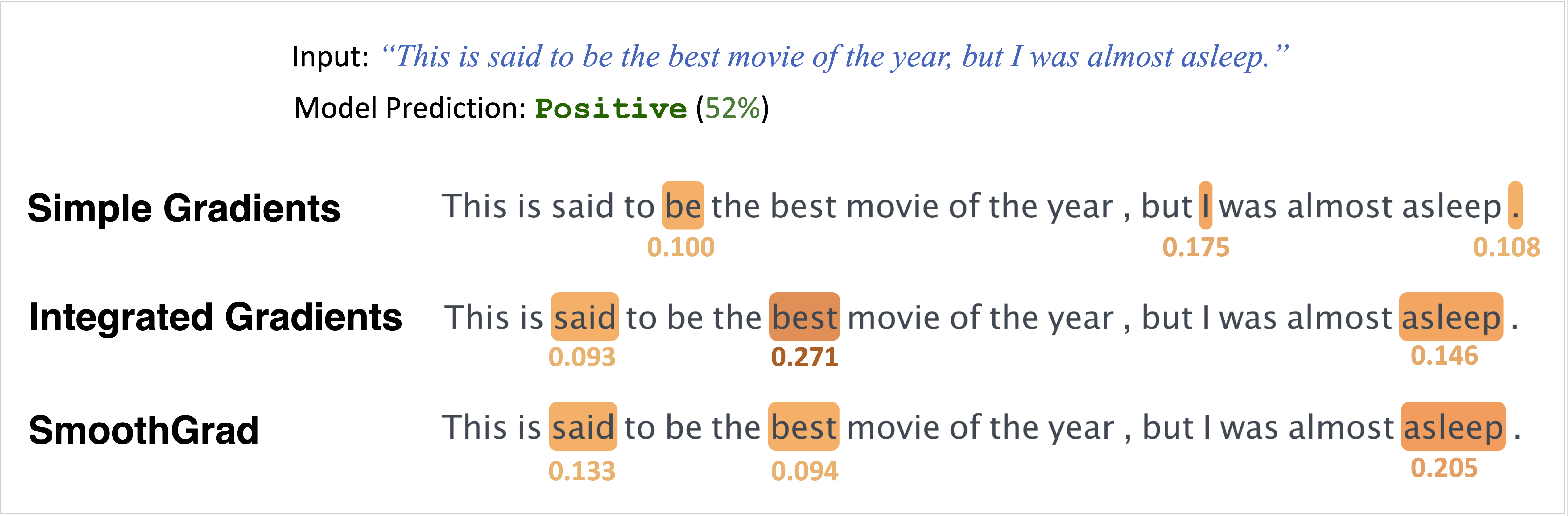}
  \caption{A visualization of different gradient methods on a sentiment classification example predicted as \texttt{positive} by an LSTM model, generated with AllenNLP Interpret. Darker shades indicate higher relevance for the prediction.}
  \label{figure:grad_methods_visual_NLP}
\end{figure}

To address the saturation and input sensitivity issues, \citet{sundararajan_axiomatic_2017} introduce \textbf{Integrated Gradients}. This method estimates the global relevance of a feature by comparing the input with a \textit{baseline input} $\overline{x}$. Typically, the baseline is chosen as an all-black image for vision and a sentence with all-zero token embeddings for language. Integrated Gradients satisfy the Input Sensitivity principle, as opposed to Gradient$\times$Input. Still, it is empirically observed to be visually noisy in CV, often resulting in blurry or unintelligible feature relevance maps \cite{smilkov_smoothgrad_2017}.\footnote{Of course, it is questionable if this noise comes from the deficiency of the explanation method or from the model reasoning mechanism itself.}

To this end, \textbf{SmoothGrad} is introduced \cite{smilkov_smoothgrad_2017}, aiming to ``remove noise by adding noise''. It argues that the Integrated Gradients method is visually noisy because the gradient can fluctuate rapidly with only subtle changes in the input.\footnote{For example, one of the most commonly used activation functions, ReLU ($y = \text{max}(0,x)$), is not continuously differentiable (at $x=0$, the gradient does not exist). Thus the fluctuation is ``infinite'' in some sense at $x=0$.} Such local fluctuations may lead to the apparent visual diffusion in relevance maps. To address this issue, SmoothGrad creates a few noisy copies of the original input, computes relevance maps for each copy with any existing gradient method, and averages all maps to obtain a less noisy map. 

This method proves effective in visually denoising the relevance maps. However, it is only qualitatively evaluated in terms of human readability; faithfulness is not assessed.

In NLP, both Simple Gradients and Integrated Gradients have been adopted, but mostly targeting sequence classification tasks. As an example, Figure~\ref{figure:grad_methods_visual_NLP} shows a visualization of these methods on sentiment classification. \citet{li_visualizing_2016} use Simple Gradients to explain token importance in RNN models on sentiment classification. More recently, targeting Transformer models, \citet{hao_self-attention_2021} and \citet{janizek_explaining_2021} adapt Integrated Gradients to capture token interactions on paraphrase detection, NLI, and sentiment classification. \\

\subsubsection{Propagation methods} \leavevmode \newline

While gradient methods follow standard backpropagation rules, propagation methods define a custom backward pass, using purposely designed local propagation rules for special layer types, such as ReLU, to reflect the relevance redistribution patterns that cannot be captured by gradients. 

\begin{figure}[!t]
  \centering
  \includegraphics[width=0.9\textwidth]{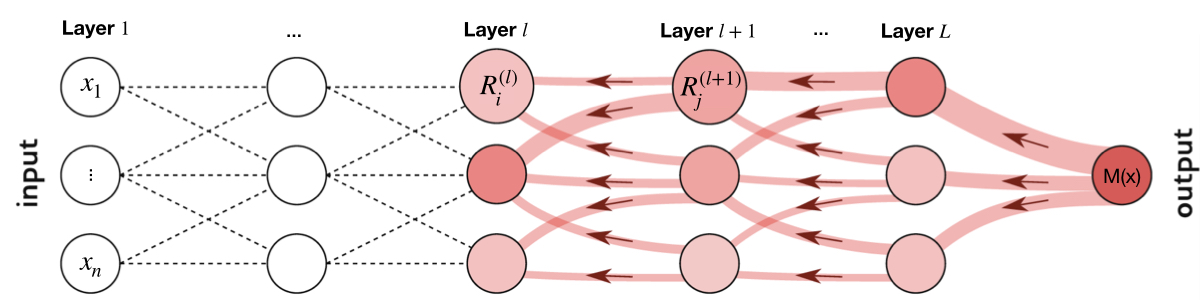}
  \caption{A schematic visualization of propagation methods. This figure is adapted from \cite{montavon_layer-wise_2019}.}
  \label{figure:propagation_methods}
\end{figure}

We now formalize the process using a feed-forward network as an example, as shown in Figure~\ref{figure:propagation_methods}. Using the previous notations, let $x$ represent the input and $M(x)$ represent the output of model $M$ after a forward pass. Denote any layer in $M$ by $l$ ($l = 1, 2, ..., L$), the dimension of which is $d_l$.
Define $R_i^{(l)}$ as the relevance score of any neuron $i$ in layer $l$. Our goal is to find $R_i^{(1)}$, the relevance of a given feature $x_i$ in the input layer, which also equals $r_i(x)$.

Unlike gradient methods, propagation methods do not have a closed-form expression for $r_i(x)$. Instead, they start with the output $M(x)$, which is considered the top-level relevance, $R_i^{(L)}$. Next, $R_i^{(L)}$ is propagated from layer $L$ to $L-1$ based on layer-specific rules, such that each neuron in $L-1$ receives a proportion of it. This process then proceeds layer by layer. Between any two adjacent layers $l$ and $l+1$, a generic function $D()$ determines how $R_j^{(l+1)}$ (the relevance of any neuron $j$ in $l+1$) is \textit{recursively distributed} to $R_i^{(l)}$ (the relevance of any neuron $i$ in $l$), where $i$ and $j$ are connected. Formally,

\begin{align}
    R_i^{(l)} =
    \begin{cases}
        M(x), & \text{for }l=L;\\
        D(R_j^{(l+1)}), & \text{for } 1 \leq l < L.\\
   \end{cases}
\end{align}

The recursion terminates once reaching $l=1$. All subsequently introduced propagation methods follow the same procedure, while differing in the definition of $D()$.\footnote{Table~\ref{table:propagation_methods} in Appendix~\ref{appendix:bp_methods} provides a formalization of how each method defines $D()$.} 
We will illustrate each method individually and provide visualizations when possible.

Among the earliest methods in this family, \textbf{DeconvNet} \cite{fleet_visualizing_2014} and \textbf{Guided BackPropagation (GBP)} \cite{springenberg_striving_2015} both design custom rules for ReLU units ($y=\text{max}(0,x)$) in particular, since it is then the most commonly used non-linear activation. Formally, suppose a ReLU unit $j$ in layer $l+1$ is connected to a set of neurons $i=0, 1, ... d_l$ in layer $l$, and let $a_i$ denote the activation of any such neuron $i$. Then we have $a_j = \text{max}(\sum_{i=0}^{d_l}a_iw_{ij}, 0)$ by the definition of ReLU, where $w_{ij}$ is the weight connecting $i$ and $j$.\footnote{To conveniently incorporate the bias term $b$, we let $a_0 = 1$ and $w_{0j} = b$.} According to the standard backpropagation rule used by Simple Gradients, only positive inputs in the forward pass ($\sum_{i=0}^{d_l} a_iw_{ij} >0$) will have non-zero gradients in the backward pass. This rule essentially loses all the relevance information from $R_j^{(l+1)}$ when the inputs are negative ($\sum_{i=0}^{d_l} a_iw_{ij} \leq 0$). 

By contrast, DeconvNet proposes to zero out the redistributed relevance \textbf{only if} the incoming relevance $R_j^{(l+1)}$ is non-positive, regardless of the input $\sum_{i=0}^{d_l} a_iw_{ij}$.
On the other hand, Guided BackPropagation combines the two rules above, zeroing out the redistributed relevance if \textit{either} the incoming relevance or the input is non-positive. Compared with Simple Gradients, both DeconvNet and Guided BackPropagation produce cleaner feature relevance visualizations as perceived by humans. However, they share several shortcomings pertaining to faithfulness. First, they fail the Input Sensitivity test (Section~\ref{section:principles}) and similarly suffer from the saturation issue \cite{sundararajan_axiomatic_2017, shrikumar_learning_2017}, like certain gradient methods mentioned before. Second, because of zeroing out negative inputs and/or negative incoming relevance, both methods cannot highlight features that contribute \textit{negatively} to the predicted class \cite{shrikumar_learning_2017}. 
Third, it is shown that both methods are essentially doing (partial) input recovery, which is unrelated to the network's decision \cite{nie_theoretical_2018}. Whichever label class is predicted, the feature attribution is almost invariant. Even with a network of random weights, Guided BackPropagation can still generate human-understandable visualizations. Thus, it is suspected that the visualization has little to do with the model's reasoning process.

While the previous two methods only treat ReLU specifically, \textbf{Layerwise Relevance Propagation (LRP)} has been proposed as a more generalized solution \cite{bach_pixel-wise_2015}. Instead of handcrafting rules directly, it first proposes a high-level \textit{Relevance Conservation} constraint, i.e., the total incoming relevance into a neuron should equal the total outgoing relevance from it. 

Any propagation rule conforming to this constraint can be called an instance of LRP. 

\citet{bach_pixel-wise_2015} quantitatively assess the faithfulness of LRP via a perturbation-based evaluation (a pixel flipping experiment in a digit classification task), yet no comparison is provided with other explanation methods. Also, LRP still suffers from the saturation problem \cite{shrikumar_not_2017} and violates Implementation Invariance (Section~\ref{section:principles}) \cite{shrikumar_learning_2017}.

In NLP, LRP has been applied and extended to sentence classification tasks \cite{arras_explaining_2016, arras_explaining_2017}. Regarding faithfulness, similar to pixel flipping in vision, a perturbation-based evaluation on the level of input tokens is used: a fixed number of tokens are deleted from the input according to the relevance score assigned by an explanation (LRP, Simple Gradients, and a random baseline), and then we measure the impact on classification performance. 
It is observed that for \textit{correctly} classified inputs, when deleting the \textit{most relevant} tokens first, LRP leads to the most rapid classification performance drop; but for \textit{incorrectly} classified inputs, when deleting the \textit{most irrelevant} tokens first, LRP can most effectively boost the performance. This indicates that LRP does provide more faithful insights into the model's reasoning mechanism compared to Simple Gradients and the random baseline.

To address LRP's failure with saturation and the Input Sensitivity test, two \textit{reference-based} methods, \textbf{DeepLift} \cite{shrikumar_learning_2017} and \textbf{Deep-Taylor Decomposition (DTD)} \cite{montavon_explaining_2017}, have been introduced. Analogous to Integrated Gradients, they aim to measure the \textit{global} contribution of input features by finding a reference point, or baseline, $\overline{x}$, to compare with the input $x$. Ideally, the baseline $\overline{x}$ should represent some ``neutral'' input, i.e., satisfying $M(\overline{x}) = 0$, so we can attribute all positive contribution to the presence of $x$. In practice, it needs to be chosen with domain-specific knowledge. The two methods differ in how the baseline input is chosen. DeepLift empirically chooses a baseline input which results in a neutral output, but DTD additionally requires the baseline to lie in the vicinity of the original input.\footnote{Specifically, this is due to its theoretical perspective, treating the explanation as an \textit{approximation} of the upper-level relevance $R_j^{(l+1)}$ through Taylor decomposition. Only when the baseline input lies in the vicinity of the original input, the Taylor decomposition can be accurate.} 
By using a baseline, they do not suffer from the issues of saturation and satisfy the Input Sensitivity principle. When evaluated on a similar pixel flipping test for faithfulness, DeepLift proves the most effective in manipulating the prediction toward the target class, compared to Integrated Gradients, Gradient$\times$Input, Simple Gradients, and Guided BackPropagation. However, DeepLift still fails the Implementation Invariance test \cite{sundararajan_axiomatic_2017}.
In terms of evaluation for DTD, only qualitative results on plausibility are reported, while faithfulness is unverified. 

In NLP, \citet{chefer_transformer_2021} extend DTD to explain the decision of Transformer models on sentiment classification. However, the explanations are evaluated against human-annotated token relevance, therefore also unrelated to faithfulness.

\subsubsection{Strengths and Weaknesses}\leavevmode\newline

Summarizing the discussion above, backpropagation-based methods have several key strengths.
First, they generate a spectrum of feature relevance scores, which is easily \textbf{understandable} for all kinds of target users. 
Second, the \textbf{computational cost} of these methods can vary significantly, but in general, they are relatively efficient to compute. Gradient-based techniques only require a handful of calls to the model's backward function. On the other hand, propagation methods involve a customized implementation of the backward pass, allowing for precise control over the relevance redistribution process while necessitating more complex computation.\footnote{In general, their computational cost is lower than counterfactual intervention (Section~\ref{appendix:counterfactual_intervention}), which typically requires multiple forward passes in addition, but higher than internal-structure analysis (Section~\ref{section:analysis_structure}), which typically requires no additional model calls. It's worth noting that these computational costs exist along a continuum rather than being binary categories.}
Third, in terms of \textbf{faithfulness}, gradients (and variants) are intrinsically tied to the influence of input features on the prediction. Empirically, certain recently proposed methods (e.g., Layerwise Relevance Propagation, DeepLift, Deep-Taylor Decomposition) are shown to be more faithful than previous baselines via perturbation-based evaluation, as mentioned before.
Finally, unlike most methods for the analysis of model-internal structures (e.g., raw attention weights), backpropagation-based methods take the \textbf{entire computation path} into account. \\

At the same time, these methods are far from perfect due to a number of weaknesses.
First, most existing backpropagation-based methods target \textbf{low-level features} only, e.g., pixels in vision and input tokens in language. It is unclear how to compute any sort of gradient w.r.t. higher-level features like case, gender, part-of-speech, semantic role, syntactic dependency, coreference, discourse relations, and so on.
Second, it is questionable how to apply such methods to \textbf{non-classification tasks}, especially when there is no single output of the model, e.g., text generation or structured prediction.
Additionally, as detailed before, certain methods \textbf{violate axiomatic principles} of faithfulness, e.g., Input Sensitivity and Model Sensitivity \cite{sundararajan_axiomatic_2017}.
Lastly, the explanation can be \textbf{unstable}, i.e., minimally different inputs can lead to drastically different relevance maps \cite{ghorbani_interpretation_2019, feng_pathologies_2018}.
Most methods are \textbf{not empirically evaluated on faithfulness} when they are first proposed, with only a few exceptions mentioned above. Moreover, subsequent researchers find many systematic deficiencies of them in ad-hoc evaluations. As mentioned before, Guided BackPropagation and DeconvNet are shown to be only doing partial input recovery, regardless of the model's behavior \cite{nie_theoretical_2018}. In addition, certain explanations (including Simple Gradients, Integrated Gradients, and SmoothGrad) can be adversarially manipulated, i.e., one can construct entirely different gradient distributions with little influence on the prediction \cite{wang_gradient-based_2020}. 
\subsection{Counterfactual Intervention}
\label{section:counterfactual_intervention}

The notion of counterfactual reasoning stems from the causality literature in social science: ``Given two co-occurring events $A$ and $B$, $A$ is said to cause $B$ if, under some hypothetical counterfactual case that $A$ did not occur, $B$ would not have occurred'' \cite{roese_counterfactual_1995, winship_estimation_1999, lipton_mythos_2017}. In the context of machine learning, counterfactual intervention methods explain the causal effect between a feature/concept/example and the prediction by \textit{erasing} or \textit{perturbing} it and observing the change in the prediction. A larger change indicates stronger importance.

One axis along which we can categorize existing studies is \textit{where the intervention happens}, in \textbf{inputs} or in \textbf{model representations}. The former manipulates the input and passes it through the original model; by contrast, the latter directly intervenes in the model-internal representations, e.g., neurons or layers. The rest of this section will elaborate on the two categories accordingly.

\subsubsection{Input Intervention} \leavevmode \newline

Input intervention methods can be further categorized along two dimensions: the \textit{target} and the \textit{operation}. The target refers to ``what is affected by the intervention'', normally input \textbf{features} (e.g., tokens) or \textbf{examples} (the entire input instance). The operation stands for the specific intervention method, which can be \textbf{erasure} (masking out or deleting the target) or \textbf{perturbation} (changing the value of the target). 

We will first classify existing work based on the target, and then on the operation.\\

\minisection{Feature-targeted intervention.} Earliest work mostly relies on \textbf{erasure}, since it is relatively straightforward to implement.

One intuitive idea is \textbf{leave-one-out}, which erases a single feature at a time and assesses the resulting change in the prediction. The feature can be input tokens \cite{li_understanding_2017} or vector dimensions \cite{kadar_representation_2017, li_understanding_2017}. In these studies, only plausibility is examined based on human perception of qualitative examples, and no faithfulness evaluation is reported. Also crucially, leave-one-out captures the linear contribution of single features, but cannot handle higher-order feature interactions. 

To address this issue, researchers propose explanation methods that erase \textbf{subsets of features} instead of individual ones. Several studies \cite[e.g.,][]{li_understanding_2017} aim to find the minimum subset of input tokens to \textbf{erase} such that the model's decision is \textbf{flipped}. Others \cite{ribeiro_anchors_2018} look for the contrary -- the minimum subset of input tokens to \textbf{keep} such that the model's decision is \textbf{unchanged} (called ``Anchors''). No matter which objective is taken, finding the exact desired subset of tokens is intractable, and thus both studies rely on approximated search. As a more efficient alternative, \citet{de_cao_how_2020} propose DiffMask, a method that trains a classifier on top of input representations to decide which subset of tokens to mask. In terms of faithfulness, Anchors is evaluated with human simulatability: compared to a popular baseline, LIME \cite{ribeiro_why_2016}, it allows users to more accurately predict model decisions on unseen examples. DiffMask is shown to be more faithful than several other baselines including Integrated Gradients \cite{sundararajan_axiomatic_2017}, but only with white-box evaluation using synthetic tasks.

\begin{figure}[!t]
  \centering
  \includegraphics[width=0.5\textwidth]{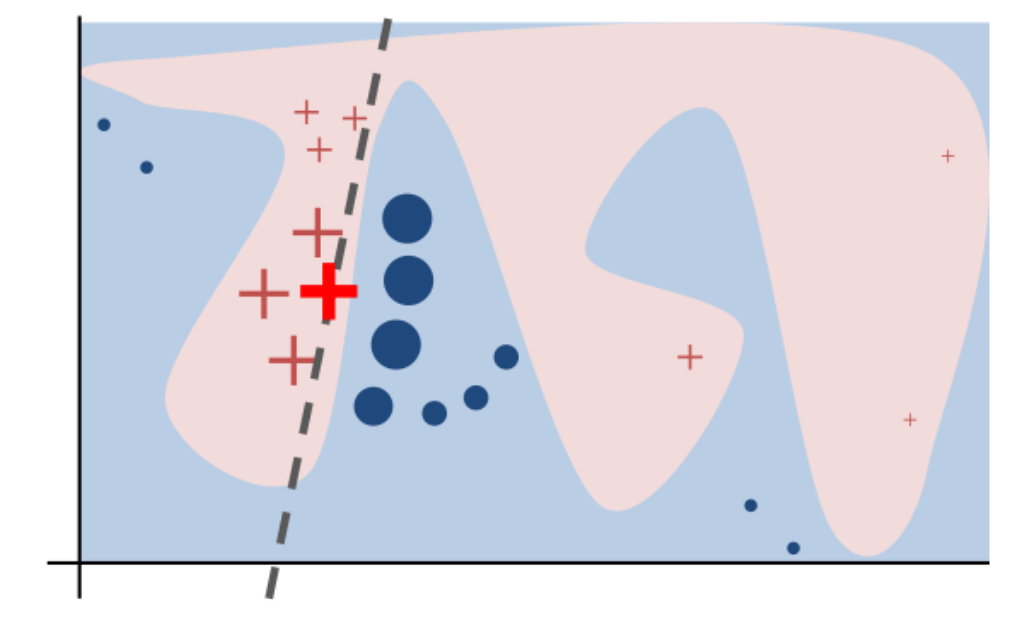}
  \caption{An illustration of LIME (figure from \citet{ribeiro_why_2016}). The complicated decision boundary of the black-box model (light blue/pink) is locally approximated by an interpretable linear model (dashed line), in the proximity of the current prediction to be explained (bold red cross).}\label{figure:LIME}
\end{figure}

Generalizing the idea of feature erasure, a novel family of methods based on \textbf{surrogate models} is proposed. The intuition is to \textit{locally} approximate a black-box model with a white-box surrogate model as an explanation of the current prediction. LIME \cite{ribeiro_why_2016}, or  Local Interpretable Model-agnostic Explanations, is a representative method of this type. Suppose the black-box model to be explained has a complicated decision boundary, as shown by light blue/pink in Figure~\ref{figure:LIME}. Our goal is to provide a local explanation for a given prediction (bold red cross). LIME first samples instances in the neighborhood of the current example by masking out different subsets of its features. Next, it obtains the model prediction on these instances and weighs them by their proximity to the current example (represented by size). Then, it approximates the model's local decision boundary by learning an interpretable model, e.g., a sparse linear regression (dashed line), on the input features, which constitutes the explanation. 

Another widely adopted surrogate-model-based method, SHAP \cite{lundberg_unified_2017}, or SHapley Additive exPlanations, can be thought of as using additive surrogate models as an explanation. Originating in the game theory, Shapley values \cite{kuhn_17_1953} are initially used to determine how to fairly distribute the ``payout'' among the ``players''. In machine learning, it is adapted to explain the contribution of input features (players) to the prediction (payout). 
Think of an input sentence $x$ as containing a set of binary features $x_i$, i.e., the presence of a token. 
The Shapley value computes the marginal contribution of $x_i$, averaged across all token subsets that include $x_i$.
When the number of features is large, the above process is computationally expensive. Therefore, the SHAP paper introduces an efficient approximation, which obviates re-sampling a combinatorial number of subsets. Followup work such as \citet{yeh_completeness-aware_2020} further extends SHAP from input tokens to higher-level concepts, represented as high-dimensional vectors. Interested readers can see \citet{mosca_shap-based_2022} for a targeted survey on SHAP-based methods.

In terms of faithfulness, LIME is evaluated with white-box tests: when used to explain models that are themselves interpretable (e.g., Decision Tree), LIME successfully recovers 90\% of important features. On the other hand, Shapley values are theoretically shown to be locally faithful, but there is no empirical evidence on whether this property is maintained after the SHAP approximation. Subsequent work also finds other limitations: (i) The choice of neighborhood is critical for such surrogate-model-based methods \cite{laugel_defining_2018}; (ii) Linear surrogate models have limited expressivity. For example, if the decision boundary is a circle and the target example is inside the circle, it is impossible to derive a locally faithful linear approximation. 

Beyond simply erasing features or feature subsets, recent work also explicitly models the contribution of \textbf{feature interactions}. For example, Archipelago \cite{tsang_how_2020} measures the contribution of the interaction between a pair of features by erasing \textit{all other} features and recomputing the prediction. It is shown to be faithful via white-box evaluation on synthetic tasks, such as function reconstruction. However, on realistic tasks like sentiment classification, only plausibility is evaluated. 

A common problem with all the above feature erasure methods is that they can produce out-of-distribution (OOD) inputs, for example, ungrammatical or nonsensical sentences after tokens are masked out. Exploiting this weakness, \citet{slack_fooling_2020} design an adversarial model to fool popular erasure-based explanation methods. Suppose a task (e.g., loan application decision) involves sensitive features (e.g., gender or race), and the explanation method is used for model auditing -- examining if a model is relying on these features. The adversarial model has two modules: a classifier to identify if an input is pre or post-erasure, and a predictor to handle the end task based on this classification. For in-distribution inputs, the predictor will rely solely on sensitive features, while for OOD inputs, it uses only insensitive features. In other words, this model is indeed biased on all in-distribution examples. However, any explanation method that works by sampling neighboring examples via feature erasure (like LIME and SHAP) will report that the model is unbiased, because they are designed to characterize the model's behavior only using these neighboring instances.

This leads us to the other operation -- \textbf{perturbation} -- as another type of feature-targeted intervention. Compared to simple erasure, perturbing the value of the feature is less likely to result in OOD inputs. Consider again our running example: \textit{This movie is great. I love it}. To study the importance of \textit{great} to the prediction, one can replace it with some other word (e.g., \textit{good}, \textit{OK}, ...) instead of just deleting it altogether, and observe the probability change of \texttt{positive}. The outcome of such perturbations is called \textit{counterfactual examples}, which resemble \textit{adversarial examples} in the robustness literature. They differ in three aspects: (i) the goal of the former is to explain the model's reasoning mechanism, while that of the latter is to examine model robustness; (ii) the former should be meaningfully different in the perturbed feature to the original example (e.g., \textit{This movie is great} $\rightarrow$ \textit{This movie is \textbf{not so} great}), while the latter should be similar to or even indistinguishable from it (e.g., \textit{This movie is great} $\rightarrow$ \textit{This movie is \textbf{G}reat}); (iii) the former can lead to changes in the ground truth label, whereas the latter should not \cite{kaushik_learning_2020}.

Generating high-quality counterfactual examples is non-trivial, as they need to simultaneously accord with the counterfactual target label, be semantically coherent, and only differ from the original example in the intended feature. In prior work, the most reliable (yet expensive) approach to collect counterfactual examples is still manual creation, as in \citet{kaushik_learning_2020} and \citet{abraham_cebab_2022}. However, recent studies propose promising ways to automate the generation, focusing on different aspects of perturbation: domain adaptation \cite{calderon_docogen_2022}, morpho-syntactic features \cite{zmigrod_counterfactual_2019, amini_naturalistic_2022}, or general-purpose \cite{wu_polyjuice_2021}.\\

\minisection{Example-targeted intervention.} In addition to feature-targeted intervention, counterfactual input intervention can also directly happen on the level of examples. 

A representative method is called \textbf{influence functions} \cite{koh_understanding_2017}, which is designed to explain which training examples are most influential in the prediction of a test example. This may remind us of similarity-based methods in Section~\ref{section:similarity_methods} \cite{caruana_case-based_1999}. Although both methods share the same goal, they rely on different mechanisms. Similarity-based methods identify the most influential training examples via similarity search, whereas influence functions are based on counterfactual reasoning -- if a training example were absent or slightly changed, how would the prediction change? Since it is impractical to retrain the model after erasing/perturbing every single training example, influence functions provide an approximation by directly recomputing the loss function. 
After the invention in vision \cite{koh_understanding_2017}, influence functions have been adapted to NLP \cite{han_explaining_2020}. Although they are claimed to be ``inherently faithful'', this is not well-supported empirically. Crucially, the approximation relies on the assumption that the loss function is convex. \citet{koh_understanding_2017} examines the assumption in vision, showing that influence functions can still be a good approximation even when the assumption does not hold (specifically, on CNNs for image classification). In NLP, though, only a qualitative sanity check is performed (on BERT for text classification); moreover, no baseline is provided. In fact, \citet{basu_influence_2020} discover that influence functions can become fragile in the age of deep NNs. The approximation accuracy can vary significantly depending on a variety of factors: ``the network architecture, its depth and width, the extent of model parameterization and regularization techniques, and the examined test points''. The findings call for increased caution on the consequences of the assumption as models become more complex.

\subsubsection{Model Representation Intervention} \leavevmode \newline

Similar to input intervention methods, we also categorize model representation intervention methods according to the \textit{target} and the \textit{operation}. Here, the target can typically be individual \textbf{neurons} or high-level \textbf{feature representations}. The operation still involves \textbf{erasure} and \textbf{perturbation}. We will still introduce existing work along the line of target first and operation next. \\

\minisection{Neuron-targeted intervention.} By intervening in individual neurons in an NN, one can explain the importance of each neuron to the prediction. The intervention can still be either \textbf{erasure} or \textbf{perturbation}.

The simplest form of erasure is still \textbf{leave-one-out}. Using the same strategy as with input features, \citet{li_understanding_2017} study the effect of zeroing out a single dimension in hidden units on the prediction. 
\citet{bau_identifying_2019} adapt the method to Machine Translation models and improve its efficiency, by searching for important neurons in a guided fashion instead of brute-force enumeration.

Apart from erasure, perturbation is another form of neuron-targeted intervention. One representative example is \textbf{causal mediation analysis} \cite{vig_investigating_2020}, which measures the effect of a \textit{control variable} on a \textit{response variable}, mediated by an intermediate variable (or \textit{mediator}). 
=In machine learning, we can think of the input example as the control variable, the model output as the response variable, and an internal neuron as the mediator. \citet{vig_investigating_2020} use this framework to analyze gender bias in LMs. 
Through a case study on GPT-2, the authors show that gender bias is concentrated in a relatively small proportion of neurons, especially in the middle layers.

The causal mediation analysis approach is further applied to tasks such as subject-verb agreement in English \cite{finlayson_causal_2021} and multilingual scenarios \cite{mueller_causal_2022}. However, all the above studies only intervene in one neuron at a time, failing to capture feature interactions.  \citet{de_cao_sparse_2022} address this issue by proposing a differentiable relaxation technique that allows efficient search through the combinatorial space of possible neuron combinations, identifying a small subset of neurons responsible for particular linguistic phenomena. In terms of faithfulness, only \citet{de_cao_sparse_2022} report the result of a perturbation-based evaluation, whereas the remaining three studies do not report empirical faithfulness evaluation results.\\

\minisection{Feature-representation-targeted intervention.} Beyond intervening in neurons, directly targeting feature representations in the model allows us to answer more insightful questions like ``Is some high-level feature, e.g., syntax tree, used in prediction?''. This is particularly meaningful to the line of work on \textit{what knowledge a model encodes} (Section~\ref{section:explainability_definition}), which oftentimes discovers linguistic features in model representations, but it is unclear whether these features are used by the model when making predictions. 

Similar to neuron-targeted intervention, the most intuitive way to perform an intervention on feature representation is \textbf{erasure}. Two pieces of concurrent work, \textbf{Amnesic Probing} \cite{elazar_amnesic_2021} and \textbf{CausalLM} \cite{feder_causalm_2021}, are representative examples. They both aim to answer the following question: is a certain feature (e.g., POS, dependency tree, constituency boundary, ...) used by the model on a task (e.g., language modeling, sentiment classification, ...)? To answer the question, they exploit different algorithms to erase the target feature from the model representation, via either Iterative Null-space Projection (INLP) \cite{ravfogel_null_2020} or adversarial training. Then, with the new representation, they measure the change in the prediction. The larger the change, the more strongly it indicates that the feature has been used by the original model. In terms of faithfulness, only CausalLM is validated with a white-box evaluation, whereas no explicit evaluation is provided for Amnesic Probing.

Methods along this line also differ in the stage where the erasure happens: \textbf{post-hoc erasure} methods remove the feature after a model representation is trained, while \textbf{adversarial erasure} methods jointly train the model with an adversarial predictor to forget the target feature. INLP \cite{ravfogel_null_2020} mentioned above is an example of post-hoc erasure, which removes the feature via a series of iterative linear projections, such that the target feature cannot be predicted by any linear classifier in the end, thus ``linearly guarded''. \citet{haghighatkhah_better_2022} improve INLP by proposing a variant that only needs a single projection. On the contrary, CausalLM \cite{feder_causal_2022} exemplifies adversarial erasure. However, neither of the two methods is perfect: \citet{ravfogel_linear_2022} discover that linear guardedness \textit{does not} guarantee that linear classifiers do not use the target features. At the same time, \citet{kumar_probing_2022} find that both post-hoc and adversarial erasure methods cannot guarantee to remove the target feature entirely; even worse, they may end up destroying other task-relevant features.

Taking a step back, even with perfect erasure techniques, a higher-level problem with the feature representation erasure methodology lies in unrealistic representations, similar to OOD inputs in the case of input erasure. For instance, since syntax is such a fundamental component of language, what does it mean if an LM \textit{is entirely ignorant of syntax}? Is it still an LM at all? 

To address this issue, \textbf{perturbation}-based methods targeting feature representations are proposed. \citet{ravfogel_counterfactual_2021} introduce AlterRep, an algorithm to manipulate the target feature value in model representations. Specifically, they investigate the task of subject-verb agreement prediction. For example, given the sentence 
\begin{quote}
    \textit{The man that they see [MASK] here.} 
\end{quote}
\noindent with an embedded relative clause (\textit{that they see}), the correct verb to fill in the mask should be \textit{is} as opposed to \textit{are}, since it should agree with \textit{man}. They now ask the question: does the model use syntactic information (e.g., the relative clause boundary) in making such predictions? To answer this, they first probe for syntactic knowledge in the model representation. If the model ``thinks'' that the \textit{[MASK]} token is outside the relative clause (factual), AlterRep would flip this knowledge via linear projection techniques like INLP, such that \textit{[MASK]} is now inside the relative clause according to the new model representation (counterfactual). They find that the probability ratio of \textit{are} versus \textit{is} increases significantly, as the model is potentially tempted to associate the verb with \textit{they} in the relative clause. Such findings suggest that syntactic information is indeed used by the model in predicting the verb. 

\citet{tucker_what_2021} study the same task and feature, but improve the method by providing syntactically ambiguous contexts, e.g., 

\begin{quote}
    \textit{I saw the boy and the girl [MASK] tall.}
\end{quote}

\noindent which can be interpreted as either \textit{[I saw the boy] and [the girl [MASK] tall].}, or \textit{I saw [the boy and the girl [MASK] tall]}. Therefore, \textit{[MASK]} can be either singular or plural. Likewise, they first probe for the model's syntactic representation, and then flip the structure of the syntax tree from one of the above interpretations to the other. 
Similar findings are reported, suggesting BERT-based models are using syntax in agreement prediction. However, no extrinsic evaluation of faithfulness is provided.

\subsubsection{Strengths and Weaknesses}\leavevmode\newline
\label{section:counterfactual_intervention_pros_cons}

Overall, there are several advantages unique to counterfactual intervention methods.
First, having its root in the causality literature, counterfactual intervention is therefore designed to \textbf{capture causal instead of mere correlational effects} between inputs and outputs.
Second, compared to other methods, counterfactual intervention methods are more often explicitly evaluated in terms of \textbf{faithfulness} \cite[e.g.,][]{ribeiro_anchors_2018, de_cao_how_2020, ribeiro_why_2016, lundberg_unified_2017, tsang_how_2020, feder_causalm_2021}. They are also shown to outperform existing baselines, mostly with predictive power or white-box tests for faithfulness.\\

However, they also share a number of disadvantages.
First, compared to other methods, counterfactual intervention is relatively more expensive in \textbf{computational cost}, normally requiring multiple forward passes and modifications to the model representation. Searching for the right targets to intervene in can also be costly.
Second, as explained before, erasure-based intervention can result in \textbf{nonsensical inputs or representations}, which sometimes allow adversaries to manipulate the explanation \cite{slack_fooling_2020}.
Third, intervening in a single feature relies on the assumption that features are \textbf{independent}. Consider the sentence \textit{This movie is mediocre, maybe even bad} \cite{wallace_interpreting_2020}. If we mask out \textit{mediocre} or \textit{bad} individually, the predicted sentiment will probably not change much (still \texttt{negative}). Hence, an explanation method that relies on single-feature erasure might report that neither token is important for model prediction. However, such a method does not capture feature interactions, like the OR relationship between \textit{mediocre} or \textit{bad} here -- as long as one of them is present, the sentiment is likely \texttt{negative}. More examples are provided in \citet{shrikumar_learning_2017}.
Additionally, interventions are often overly \textbf{specific to the particular example} \cite{wallace_interpreting_2020}. This calls for more insights into the scale of such explanations (i.e., if we discover a problem, is it only about one example or a bigger issue?) and general takeaways (i.e., what do we know about the model from this explanation?).
Finally, counterfactual intervention may suffer from \textbf{hindsight bias} \cite{de_cao_how_2020}, which questions the foundation of counterfactual reasoning. Specifically, the fact that a feature can be dropped without influencing the prediction \textit{might not mean} that the model ``knows'' that it can be dropped and has not used it in the original prediction. 
\citet{de_cao_how_2020} illustrates this point with an intuitive example of the Reading Comprehension task, where a model is given a paragraph and a question, and should identify an answer span in the paragraph. Now, using counterfactual intervention, if we mask out everything except the answer in the paragraph, the model will for sure predict the gold span. Nonetheless, this does not imply that everything else is unimportant for the model's original prediction. The issue again calls for a rethinking of the fundamental assumptions of counterfactual reasoning. 
\subsection{Self-Explanatory Models}
\label{section:self-explantory_models}

In contrast with all the above post-hoc methods, self-explanatory models provide built-in explanations. Typically, explanations can be in the form of feature importance scores, natural language, causal graphs, or the network architecture itself.

Prior work on self-explanatory models can be broadly categorized into two lines based on \textit{how the explanation is formed}: \textbf{explainable architecture} or \textbf{generating explanations}. The former relies on a transparent model architecture, such that no extra explanation is necessary. The latter, though, may still involve opaque architectures, but generates explicit explanations as a byproduct of the inference process. 

\subsubsection{Explainable Architecture} \leavevmode\newline 

While end-to-end NNs are a black-box, classic machine learning models like Decision Trees and linear regression have a highly interpretable reasoning mechanism. Drawing inspiration from them, researchers attempt to design neural models with more structural transparency while maintaining their performance.\\

\noindent \textbf{Neural Module Networks (NMNs)} are one representative example, specifically in the context of Question Answering (QA) tasks. Given a complex question (e.g., \textit{Are there more donuts than bagels in the image?}), humans naturally decompose it into a sequence of steps (e.g., look for donuts and bagels, count them, and compare the counts). With the same motivation, NMNs parse the input question into \textit{a program of learnable modules} (e.g., \texttt{compare(count(donuts), count(bagels))}), which is then executed to derive the answer. 

There are three potentially learnable components in NMNs: (i) the question parser (question $\rightarrow$ syntax tree), (ii) the network layout predictor (syntax tree $\rightarrow$ program), and (iii) the module parameters (program $\rightarrow$ answer). Previous studies differ in whether and how each component is learned. \citet{andreas_neural_2016} introduce the earliest version of NMN, where only module parameters are learned and the other two components are pretrained or deterministic. In a follow-up study \cite{andreas_learning_2016}, they extend the framework to also jointly learn the network layout specific to each question, which is then named Dynamic Neural Module Network (DNMN). \citet{hu_learning_2017} further extend the method to learn the question parser as well, resulting in their End-to-End Module Network (N2NMN). 

\begin{figure}[!t]
  \centering
  \includegraphics[width=0.9\textwidth]{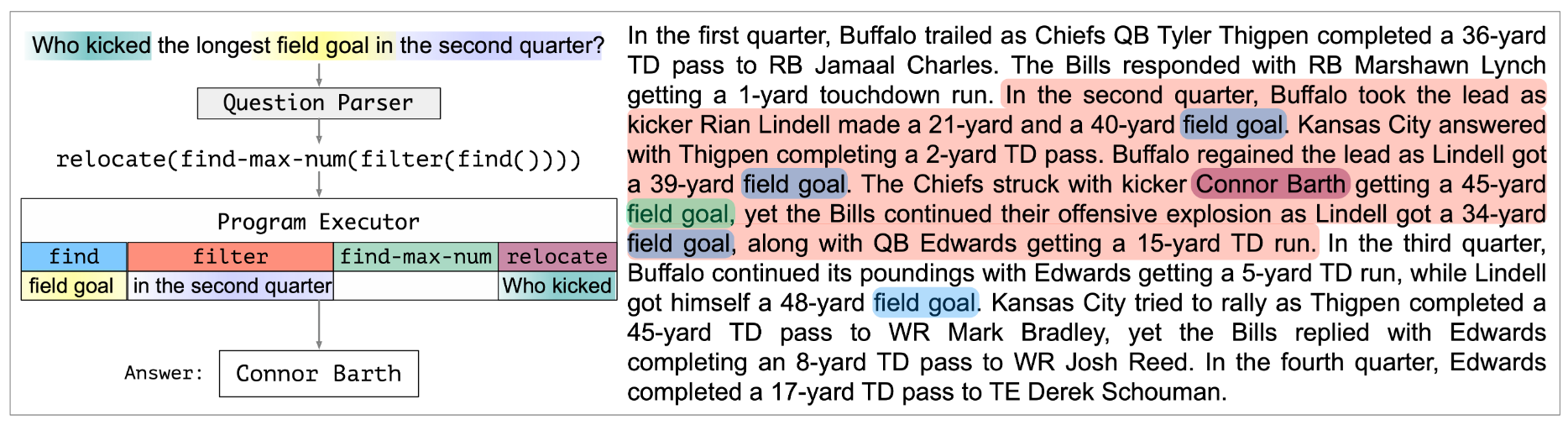}
  \caption{An illustration of Neural Module Network on the QA task (figure from \citet{gupta_neural_2019}). Given a paragraph of context and a question, the model parses the question into a program of learnable modules, which is then executed on the context to derive the answer. Colors represent the correspondence between spans in the question, the modules, and the intermediate outputs in the context. }\label{figure:gupta_nmn}
  \vspace{-0.2in}
\end{figure}

The above methods have been demonstrated to be effective on various visual QA tasks including VQA \cite{antol_vqa_2015}, SHAPES \cite{andreas_neural_2016}, GeoQA \cite{krishnamurthy_jointly_2013}, and CLEVR \cite{johnson_clevr_2017}, most of which are based on synthetic data, though. More recent studies investigate the application of NMNs on realistic data, especially in the language-only domain. \citet{jiang_explore_2019} apply NMN to HotpotQA \cite{yang_hotpotqa_2018}, a multiple-document QA dataset. However, their model is incapable of symbolic reasoning, such as counting and sorting, so it can only solve questions with directly retrievable answers in the context. To address this issue, \citet{gupta_neural_2019} introduce a set of modules for each symbolic operation, e.g., \texttt{count}, \texttt{find-max-num}, \texttt{compare-num}. Their model is capable of answering questions involving discrete reasoning in the DROP dataset \cite{dua_drop_2019}, for example, \textit{Who kicked the longest field goal in the second quarter?}, given a long description of the match. See Figure~\ref{figure:gupta_nmn} for an illustration. 

Despite their presumably transparent structure, there are two main faithfulness-related problems with NMNs: First, the modules' actual behavior may not be faithful to their intended function. Most NMNs pre-define modules only in terms of input/output interface. Their actual behavior -- module parameters -- are then typically learned from end-to-end supervision on the entire pipeline, i.e. only the question, context, and the final answer. Thus, there is no control over the intermediate output of individual modules. Consider our previous example \texttt{compare(count(donuts), count(bagels))}. Though we name the module \texttt{compare}, it may not perform the comparison function. Theoretically, it is possible that \texttt{compare} alone outputs the final answer, whereas \texttt{count} is ignored. \citet{subramanian_obtaining_2020} empirically confirm such failure cases and experiment with several remedies, such as introducing intermediate supervision and limiting the module complexity. Nevertheless, improved faithfulness comes at the cost of end-task accuracy. The second problem is that symbolic modules may not be expressive enough for the flexible semantics of natural language. For example, \citet{gupta_neural_2019} note that questions like \textit{Which quarterback threw the most touchdown passes?} would necessitate modules with some key-value representation (\texttt{\{quarterback: count\}}), which are non-obvious to design. Other subtle semantic phenomena like context-conditional parsing and coreference pose similar challenges.

Nonetheless, the compositional nature of NMNs brings about promising research opportunities. In particular, it is possible to pretrain modules independently on synthetic tasks and then use them with fixed parameters when training the QA pipeline. This allows us to ensure their faithfulness as well as exploit transferable knowledge.

NMNs can be thought of as a variant of \textbf{Neural-Symbolic Models} (NSMs), a more generic concept loosely defined as neural models that integrate symbolic reasoning \cite[][i.a.]{yi_neural-symbolic_2018, mao_neuro-symbolic_2019, bogin_latent_2021}. Interested readers can refer to \citet{Hamilton_2022} for a focused review on neuro-symbolic NLP.\\

\noindent \textbf{Models with constraints} are another example of explainable architectures. The idea is to incorporate constraints into neural networks from classic interpretable models, like generalized linear regression \cite{alvarez_melis_towards_2018} and finite-state automata \cite{schwartz_bridging_2018, deutsch_general-purpose_2019, jiang_cold-start_2020}. Still, a major challenge lies in the trade-off between interpretability and performance.

\begin{figure}[!t]
  \centering
  \includegraphics[width=0.9\textwidth]{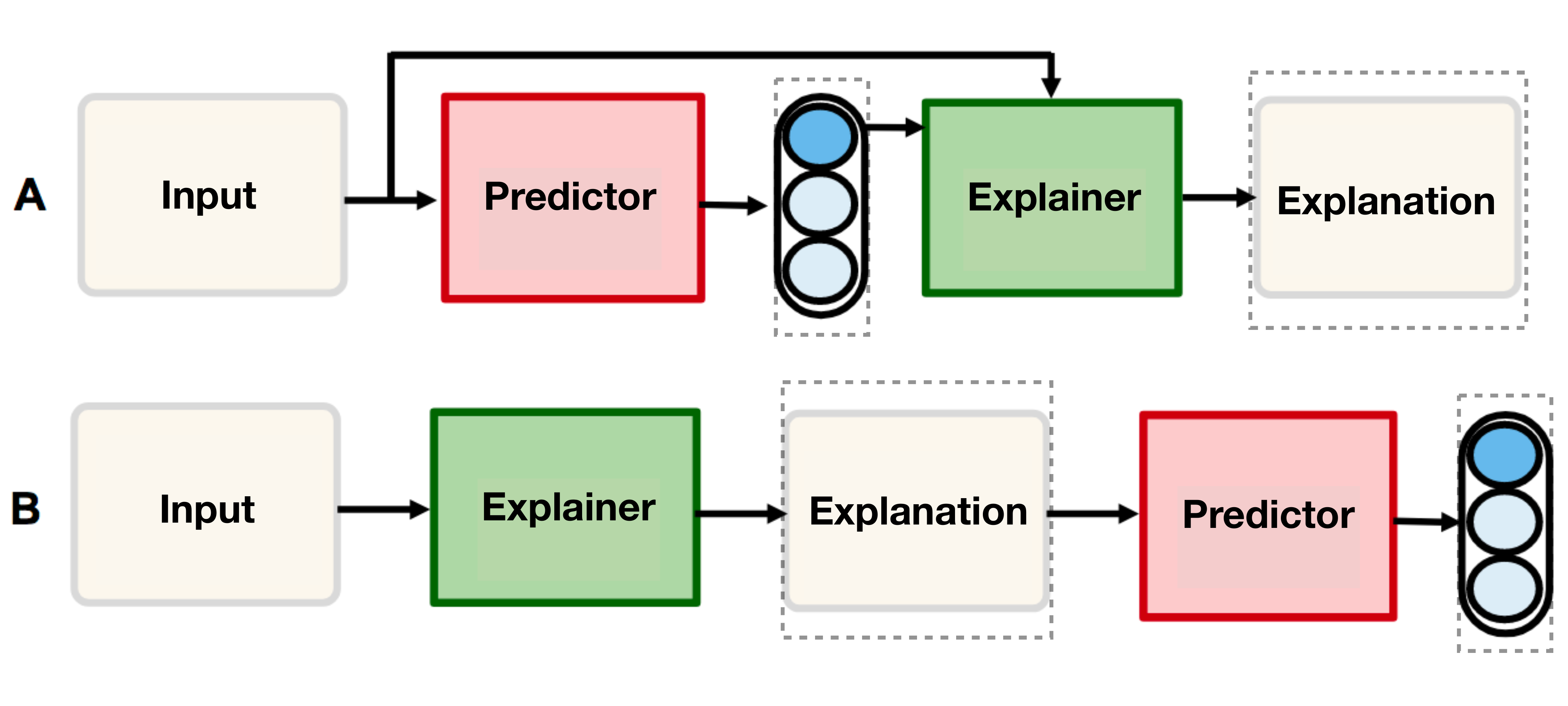}
  \caption{A comparison of frameworks of generating explanations (figure adapted from \citet{kumar_nile_2020}): predict-then-explain (A), explain-then-predict (B), and jointly-predict-and-explain (no particular dependency between explainer and predictor, thus not visualized). The training signals apply to everything in the gray boundary (the explanation and the prediction).}\label{figure:generating_explanation}
\end{figure}

\subsubsection{Generating Explanations} \leavevmode\newline  

Besides using architecture as an implicit explanation, another type of self-explanatory model learns to generate an explicit explanation as an additional task, aside from making the prediction. For supervision, human-written explanations are often used as an additional training signal, along with the end-task label. According to the dependency relationship between the \textit{predictor} and the \textit{explainer}, we can classify existing work into three categories: \textbf{predict-then-explain}, \textbf{explain-then-predict}, and \textbf{jointly-predict-and-explain}, as illustrated in Figure~\ref{figure:generating_explanation}.\\

\noindent \textbf{Predict-then-explain} models first make a prediction with a standard black-box predictor, and then justify the prediction with an explainer (Figure~\ref{figure:generating_explanation}A). This is analogous to previous post-hoc explanation methods (from Section~\ref{section:similarity_methods} to \ref{section:counterfactual_intervention}). This framework has been applied to many domains, including vision \cite{hendricks_generating_2016}, language \cite{camburu_e-snli_2018}, and multimodal tasks \cite{park_multimodal_2018}. However, it suffers from the same faithfulness challenge as all other post-hoc methods: since the predictor does not depend on the explainer, there is no guarantee that the explanation accurately reflects the reasoning process behind the prediction. Moreover, as the supervision comes from human-provided explanations, the explainer is only explicitly optimized in terms of plausibility, but not faithfulness.\\

\noindent \textbf{Explain-then-predict} methods (Figure~\ref{figure:generating_explanation}B) have been introduced in response to this issue. In this framework, the explainer first generates an explanation, which is then provided as the \textit{only} input to the predictor. In other words, the predictor can only access the explanation, but not the original input example. The intuition is that the prediction can only be made based on the explanation, which renders the predictor ``faithful by construction''.

Methods within this framework mainly differ in the \textit{form} of explanation, which is typically either \textbf{an extract from the input} or \textbf{natural language}, analogous to extractive and abstractive summarization respectively.

The former is also known as rationale-based methods, where a \textit{rationale} is defined as a part of the input that is short but sufficient for the prediction \cite{zaidan_using_2007}.\footnote{This is analogous to the notion of ``anchors'' mentioned in Section~\ref{section:counterfactual_intervention}.} For example, in sentiment classification, seeing the phrase \textit{not good} is probably enough for predicting \texttt{negative}. The job of the explainer is to extract such a rationale, and thus it is also called \textit{extractor} in this scenario. Then, the predictor will make the prediction using only the extracted rationale.

One difficulty lies in how to effectively find the rationale span, given the formidably large search space. \citet{lei_rationalizing_2016} propose to guide the search with reinforcement learning. \citet{bastings_interpretable_2019} introduce a re-parameterization technique as an alternative, which makes the learning differentiable. \citet{jain_learning_2020} discard searching and directly obtain candidate rationales from existing post-hoc explanation methods (e.g., backpropagation-based ones) instead. 

Although rationale-based models seem to be ``faithful by construction'', they are not necessarily so: (i) Clearly, only the rationale is used in the prediction, but this does not tell us anything about \textit{how} it is used. For example, the predictor might only be looking at superficial patterns in the rationale (e.g., the number of tokens that are kept). \citet{jacovi_aligning_2021} confirm the existence of these so-called ``trojan'' explanations in practice. (ii) The rationales are shown to be unstable to minimal meaning-preserving perturbations on the input and hard to understand by users, even those with advanced machine learning knowledge \cite{zheng_irrationality_2021}. 
(iii) Finally, whether rationales can ever be a sufficient explanation for the prediction is highly task-specific. For instance, it might make sense to classify the sentiment only based on a subset of tokens, but what about numerous tasks that are intrinsically context-dependent, such as NLI, coreference resolution, relation extraction, etc.? 

Alternative to extracted rationales, a more flexible form of explanation is natural language (also called \textit{free-text} explanation). Consider the NLI task as an example. Given a hypothesis (e.g., \textit{An adult dressed in black holds a stick}) and a premise as input (e.g., \textit{An adult is walking away, empty-handed}), the explainer first generates an explanation (\textit{Holds a stick implies using hands so it is not empty-handed}), and then the predictor makes a prediction (\texttt{Contradiction}) only based on the explanation. When experimenting with this model on the SNLI dataset \cite{bowman_large_2015}, \citet{camburu_e-snli_2018} discover a trade-off between the task accuracy and the plausibility of the explanation. It is also found that the model can generate self-inconsistent explanations \cite{camburu_make_2020}, e.g., \textit{Dogs are animals} and \textit{Dogs are not animals}. Moreover, the explanation might contain cues to the label, e.g., patterns like \textit{X implies Y} / \textit{X is a type of Y} oftentimes indicate \texttt{Entailment}, while \textit{X is not the same as Y} is a strong signal of \texttt{Contradiction}. To overcome this issue, \citet{kumar_nile_2020} propose the Natural language Inference over Label-specific Explanations (NILE) model, where every class label has a corresponding explainer. Given an input, an explanation is generated for each label (\texttt{Entailment}, \texttt{Neutral}, and \texttt{Contradiction}). Then, all three explanations are fed to the predictor, which makes a decision after comparing them. This precludes the possibility of the predictor exploiting cues in the explanation pattern. NILE is shown to have comparable accuracy with SOTA models on SNLI, as well as better transferability to OOD datasets. Through an extensive evaluation, the authors compare the faithfulness of a few variants of NILE.\\

\noindent \textbf{Jointly-predict-and-explain} methods have two possible structures: (i) there are still an explainer and a predictor, but the predictor can access both the explanation and the input example;\footnote{In contrast, the above-mentioned explain-then-predict methods do not allow the predictor to access the input example. This is why we categorize (i) as jointly-predict-and-explain methods.}  (ii) there are no separate explainer and predictor at all --- everything is produced jointly.

Approaches in (i) suffer from a similar faithfulness challenge as predict-and-explain methods do, because the predictor can make its decision \textit{only} based on the input using whatever reasoning mechanism, while entirely ignoring the explanation. For example, \citet{rajani_explain_2019} introduce a QA model that takes in a question, generates an explanation first, and then produces an answer based on \textit{both} the question and the explanation. Another example is a variant of the NILE model \cite{kumar_nile_2020} previously discussed in explain-then-predict, which allows the predictor to look at both the premise-hypothesis pair and the explanation. 

For works of type (ii) (where there are no separate explainer and predictor), given the input example as the prompt, a generation model outputs a continuation including both the explanation and the prediction in some designated order. This is analogous to any other generation task. Existing studies along this line differ in the choice of the generation model and the end task. For example, \citet{ling_program_2017} use LSTMs to solve algebraic problems and provide intermediate steps, and \citet{narang_wt5_2020} train T5 to generate predictions with explanations for NLI, sentiment classification, and QA. 

In particular, we elaborate on a series of studies on generating structured proofs for deductive reasoning. This is especially interesting since most previous work only aims at generating single-step explanations (e.g., a single sentence), but complex tasks would require a structured reasoning chain as explanations. 
As one of the earliest studies of this kind, \citet{tafjord_proofwriter_2021} develop ProofWriter, which takes in a set of premises and a hypothesis, and then decides if the hypothesis is true by providing a structured proof. 
Specifically, there are two versions of ProofWriter, all-at-once (generating the entire proof in one shot) and iterative (generating one step at a time). The all-at-once ProofWriter cannot guarantee that the proof is faithful, since it may not ``believe'' in the proof that it has generated.
The iterative ProofWriter bridges this gap by deriving the proof one step at a time. All steps are already verified during generation. Therefore, it is faithful by construction. However, compared to the all-at-once version, it suffers from efficiency and input length limitations.

After ProofWriter, \citet{dalvi_explaining_2021} generalize the idea to real-world data, developing the EntailmentWriter to verify hypotheses about scientific questions. Nevertheless, only an all-at-once version is implemented, indicating that the same faithfulness risk exists. To address this issue, \citet{hong_metgen_2022} propose METGEN (Module-based Entailment Tree GENeration), a modularized version of EntailmentWriter. They define multiple single-step entailment modules each performing a certain type of reasoning, such as substitution, conjunction, and if-then, like in Neural Module Networks (NMNs). Then, a high-level reasoning controller takes in the given facts and selects which module to use at each step to eventually arrive at the hypothesis. Notably, to encourage faithfulness, each module is trained with well-formed synthetic data and then finetuned on EntailmentBank, and their weights are frozen from then on.
The improvement in faithfulness also leads to higher performance of METGEN compared to previous versions of EntailmentWriter.

With the advances in in-context learning \cite{brown_language_2020}, recent work has started to explore the possibility of generating explanations with few-shot prompting. Earliest studies including \citet{wiegreffe_reframing_2022} and \citet{marasovic_few-shot_2022} find that Large Language Models (LLMs) like GPT-3 show potential for generating decently plausible free-text explanations with only a few examples, though the quality is still far from human-provided ones. 
On the other hand, \citet{ye_unreliability_2022} find that such few-shot model-generated explanations are often non-factual (i.e., not correctly grounded in the input) and inconsistent (i.e., not entailing the prediction). Nonetheless, these explanations are still helpful for users to calibrate the confidence of model predictions, since explanations rated by humans as plausible are more likely to co-occur with accurate predictions.

Another line of work under few-shot explanation generation is Chain-of-Thought-style (CoT) prompting, which is specifically effective for complex reasoning tasks like Math Word Problems and Multi-hop QA. Given a complex question $Q$, an LM is prompted to generate a reasoning chain $C$ along with the final answer $A$. Specifically, the prompt consists of a few instances of $(Q, C, A)$ triples as in-context exemplars. This allows pre-trained LLMs to solve unseen questions with much higher accuracy than standard prompting \cite{brown_language_2020}, where the exemplars do not contain the reasoning chain $C$. We categorize existing CoT-style prompting methods into three types: all-at-once, ensemble-based, and modularized. \textbf{All-at-once} prompting means that the LM produces $C$ and $A$ as one continuous string, without any dependencies or constraints in between. Scratchpad \cite{nye_show_2021}, CoT \cite{wei_chain_2022}, and ``Let's think step by step'' \cite{kojima_large_2022}, are all examples of this kind. \textbf{Ensemble-based} prompting is designed to overcome the locality issue of one-shot generation in previous methods by sampling multiple $(C,A)$ pairs and choosing the best answer via strategies like majority voting. Examples include Self-Consistent CoT \cite{wang_self-consistency_2022}, Minerva \cite{lewkowycz_solving_2022}, and DIVERSE \cite{li_advance_2022}, which differ mainly in the granularity of voting and the underlying LM. \textbf{Modularized} methods break down $Q$ into subproblems and then conquer them individually \cite{hong_metgen_2022, creswell_faithful_2022, qian_limitations_2022, jung_maieutic_2022}. In particular, Least-to-Most Prompting \cite{zhou_least--most_2022} uses an LLM to first reduce the question to subquestions, and then sequentially answers them conditioned on its answers to previous subquestions. However, for all the above CoT-style prompting methods, the generated reasoning chain is entirely in NL, so there is no guarantee of faithfulness: the answer does not need to causally follow from the reasoning chain. To bridge this gap, recent work attempts to generate the reasoning chain in some Symbolic Language (SL) (e.g., Python) and calls an  external solver (e.g., a Python interpreter) to derive the answer by deterministically executing the reasoning chain. In Program-of-Thought (PoT) \cite{chen_program_2022} and Program-Aided Language Models (PAL) \cite{gao_pal_2023}, the reasoning chain is entirely in Python, which allows the same underlying LM to outperform vanilla CoT with NL by a large margin on a wide range of arithmetic and symbolic reasoning tasks. \citet{lyu_faithful_2023} propose Faithful CoT, which interleaves NL comments and SL programs for users to better understand and potentially interact with the model. 
These symbolic CoT prompting methods guarantee that the reasoning chain is a faithful explanation of how the model derives the answer. However, \textit{how the model generates the reasoning chain} is still an opaque process, so there is no full interpretability of the entire pipeline.

\subsubsection{Strengths and Weaknesses}\leavevmode\newline

In summary, self-explanatory models have several strengths.
First, by definition, self-explanatory models provide built-in explanations, so there is \textbf{no need for post-hoc explanations}.
Second, \textbf{the form of explanation is flexible}, e.g., model architecture, input features, natural language, or causal graphs.
Third, it is possible to \textbf{supervise the explainer} with human-provided explanations. This is helpful for learning more plausible explanations, as well as encouraging the model to rely on desired human-like reasoning mechanisms instead of spurious cues.
Finally, certain self-explanatory models \cite[e.g.,][]{tafjord_proofwriter_2021, hong_metgen_2022, chen_program_2022, gao_pal_2023, lyu_faithful_2023}, are \textbf{faithful by construction} (we should be extra cautious about this claim, though).\\

Self-explanatory models, however, also present a few key weaknesses.
First, many such models cannot guarantee \textbf{faithfulness}, e.g., Neural Module Networks without intermediate supervision, predict-then-explain models, rationale-based explain-then-predict models, and certain jointly-predict-and-explain models.
Second, the influence of explanations on \textbf{task performance} is mixed in self-explanatory models. Many studies discover a trade-off between performance and interpretability \cite[][i.a.]{narang_wt5_2020, subramanian_obtaining_2020, hase_leakage-adjusted_2020}, while others observe a positive impact on the performance from including explanations (e.g., CoT-style prompting). The effect highly depends on the task, the model family and size, the format of explanations, whether they are tuned, and how they are used (as inputs, targets, or priors) \cite{hase_when_2022, lampinen_can_2022}. To make the process less mysterious, \citet{ye_complementary_2022} and \citet{ye_explanation_2023} develop principled methods to select exemplars with explanations for few-shot prompts, in order for interpretability to benefit performance.
Finally, large-scale human supervision on explanations can be \textbf{costly and noisy} \cite{dalvi_explaining_2021}. Also, it is \textbf{hard to automatically evaluate} the quality of model-generated explanations given reference human explanations, since there can be multiple ways to explain a prediction. 
\section{Summary and Discussion}
\label{section:summary_discussion}

After presenting existing explanation methods in detail, we now summarize all five method families in terms of faithfulness from both theoretical and empirical perspectives.

\textbf{Similarity-based methods} rely on the theoretical assumption that models use similar reasoning mechanisms for examples with similar representations in the learned space. However, this assumption might be invalid if the model is not robust to subtle changes in this space. Empirically, similarity-based methods are rarely evaluated with regard to faithfulness. 

\textbf{Analysis of model-internal structures} has sparked a long debate regarding the empirical faithfulness of raw attention weights as explanations. Theoretically, the lack of faithfulness can be attributed to issues like information mixing, locality, and/or an intrinsic lack of causality in attention weights. Recent approaches, including regularization, global characterization, and integration with other explanation methods, show promise in addressing these concerns. 

\textbf{Backpropagation-based methods} are faithfulness-driven by definition, but they still encounter theoretical challenges such as saturation, input sensitivity, and implementation variance. Subsequent variants have addressed these issues by considering the input alongside the gradient or by incorporating a baseline for comparison. Empirically, these methods are more frequently assessed for faithfulness, mainly through perturbation-based evaluation, than earlier methods. However, certain methods in this family are shown to be only doing partial input recovery, regardless of the model's prediction. Meanwhile, much progress has been made in improving the plausibility of these methods, especially in reducing the visual noise in relevance maps. 

\textbf{Counterfactual intervention} is theoretically grounded in the causality literature, yet it still faces nuanced pitfalls such as the feature independence assumption and hindsight bias. Empirically, methods of this family are most often evaluated with predictive power or white-box tests, which show their improved faithfulness compared to earlier baselines. Still, counterfactual intervention might produce OOD inputs, potentially exploitable by adversaries. Perturbation-based interventions are less prone to generating OOD inputs than erasure-based ones but are more complex to automate. 

\textbf{Self-explanatory models}, designed for intrinsic interpretability, might encounter theoretical challenges like the lack of intermediate supervision, label leakage, and social misalignment. As a result, these can empirically undermine their ``faithfulness by construction'' claim.

Next, we discuss the common virtues and challenges of existing methods, as well as identify future work directions towards faithful interpretability in NLP.

\subsection{Virtues}

Many studies are conducive to \textbf{bridging the gap between \emph{competence} and \emph{performance}} in language models. The two terms originate from linguistics: competence describes humans' (unconscious) knowledge of a language, whereas performance refers to their actual use of the knowledge \cite{chomsky_aspects_1965}. For humans, there is a gap between competence and performance, e.g., we can theoretically utter a sentence with infinitely many embedded clauses, but in practice, it is impossible to do so. Similarly, for language models, \textit{what they know} can be different from \textit{what they use (in a task)}, as discussed in Section~\ref{section:explainability_definition}. Previous work on interpretability predominantly focuses on competence, whereas more recent studies (e.g., all five methods discussed in this survey) aim at answering the performance question. This allows us to better understand whether the same gap exists in models, and if so, how we can bridge it.

It is also noteworthy that there has been \textbf{increasing awareness of faithfulness and other principles} of model explanation methods, especially since the seminal opinion piece by \citet{jacovi_towards_2020}. A number of evaluation methods have been proposed; see Section~\ref{section:principles} for details. Though each of them depends on assumptions and application scenarios, this is a good starting point for quantitatively assessing the quality of explanations.

In addition, explanations produced by most above-mentioned methods are \textbf{intuitive to understand}, even for lay people. This is because the form of explanation is simple, mostly feature importance scores, visualization, natural language, or causal graphs. Though the model and the explanation method may be opaque, the explanation itself is easily understandable. 

Finally, in terms of applicability, many available explanation tactics are \textbf{model-agnostic}, especially for classification tasks. Also, numerous \textbf{toolkits} have been developed to help users apply explanation methods to their own models. See Appendix~\ref{appendix:technical_details} for more details.

\subsection{Challenges and Future Work}
Despite the remarkable advances, the area of NLP interpretability still faces several major challenges, which also provide exciting opportunities for future research.

So far, a large number of explanation methods still \textbf{lack objective quality evaluation}, especially in terms of \textbf{faithfulness}. There has not been any established consensus on how to measure faithfulness. Different evaluations are often not directly comparable and yield inconsistent results. This necessitates the need for a universal evaluation framework (and maybe even a meta-evaluation framework of existing evaluation frameworks), which is fundamental to measuring the progress of any research in this area.

Next, most existing methods provide explanations in terms of \textbf{surface-level features}, e.g., pixels in vision and tokens in language. Future work can focus more on how to capture the contribution of \textbf{higher-level features} in a task, including linguistic (case, gender, part-of-speech, semantic role, syntax dependency, coreference, discourse relations, ...), and extra-linguistic (demographic features, commonsense and world knowledge, ...) ones. Several studies on counterfactual intervention provide inspiring examples \cite{ravfogel_null_2020, elazar_amnesic_2021, tucker_what_2021}; see Section~\ref{section:counterfactual_intervention} for details.

Another challenge is that most existing methods capture the \textbf{contribution of individual features to the prediction}, but not that of higher-order feature interactions. See Section~\ref{section:counterfactual_intervention_pros_cons} (c) for an illustration of why this is a problem. Future work can address the issue by developing more \textbf{flexible forms of explanation} instead of flat importance scores, e.g., feature subsets as in certain counterfactual intervention methods \cite[e.g.,][]{ribeiro_anchors_2018} and causal graphs as in several self-explanatory methods \cite[e.g.,][]{tafjord_proofwriter_2021, dalvi_explaining_2021}.

In addition, existing work mostly focuses on \textbf{limited task formats}, e.g., classification and span identification. This limits the downstream applicability of these methods to real-world scenarios. Future work can study \textbf{alternative task formats} such as language generation and structured prediction, or even better, develop explanation methods that are generalizable across tasks. Recent work such as \citet{yin_interpreting_2022} makes promising initial progress in this direction.

Meanwhile, it is not always obvious whether insights from model explanations are \textbf{actionable}. For example, given the explanation of the model's decision on one test example, the user finds that the model is not using the desired features. Then how should they go about fixing it -- through the data, model architecture, training procedure, hyper-parameters, or something else? How does the user \textbf{communicate} with the model? Consequently, \textbf{interactive} explanations will be a fruitful area for future study. A few studies on knowledge editing have shown the plausibility of the idea \cite{madaan_improving_2021, kassner_beliefbank_2021}.

Interestingly, the relationship between \textbf{model performance and interpretability} is not always predictable. Sometimes there is a synergy, but sometimes there is tension. This issue is especially evident in self-explanatory models; see Section~\ref{section:self-explantory_models} for more details. It will be greatly helpful to have a \textbf{theoretical understanding} of when and how explanations can help with model performance, as demonstrated by several recent studies \cite{ye_complementary_2022, hase_when_2022}. In the meantime, we need to cautiously balance between performance and interpretability depending on application scenarios.

Finally, we want to emphasize that \textbf{faithfulness is not the only desideratum}. After all, explanations are meant to help the target audience better understand the model, and faithfulness is only one (but fundamental) condition to this end. Furthermore, explanations should be \textit{useful} in helping the target audience with certain goals, such as decision making, model debugging, knowledge discovery, and so on. Findings from existing studies are still quite disappointing in this respect: for example, according to \citet{bansal_does_2021}, when human decision makers collaborate with a model (e.g., in computer-assisted diagnosis), current explanation methods rarely help them make more accurate decisions, but instead exacerbate their over-trust in model predictions even when they are wrong. As a result, more work should be done to investigate faithfulness \textbf{under the context of real-world applications}, especially its relationship with user-oriented desiderata like \textit{utility}.
\section{Conclusion}
\label{section:conclusion}

This survey provides an extensive tour of recent advances in NLP explainability, through the lens of faithfulness. Despite being a fundamental principle of model explanation methods, faithfulness does not have a universally accepted technical definition or evaluation framework. This absence makes it challenging to compare different methods based on faithfulness, and many methods do not provide quantitative faithfulness evaluation results.

We critically review five families of existing model explanation methods: similarity-based methods, analysis of model-internal structures, backpropagation-based methods, counterfactual intervention, and self-explanatory models. We introduce each category in terms of their representative work, strengths, and weaknesses, with a special focus on faithfulness. 

In summary, similarity-based methods are not primarily driven by faithfulness and are rarely assessed on this basis. For the analysis of model-internal structures, the early attempt to treat raw attention weights as explanations has faced strong criticism regarding faithfulness. However, there has been promising new progress in improving the faithfulness of attention by addressing these concerns. Backpropagation-based methods are intuitively faithfulness-motivated due to the nature of gradients, but theoretical and empirical evidence often points to various faithfulness issues, such as saturation and input sensitivity. Still, later variants in this family have shown gradual improvements in some of these aspects. Counterfactual intervention methods, rooted in causal inference, are also faithfulness-motivated. Nevertheless, certain types of intervention, such as perturbation, are more likely to be faithful than others like erasure, due to the OOD issue. There are also nuanced practical concerns such as feature independence and hindsight bias to consider. Self-explanatory models, which do not rely on any post-hoc explanation methods, often claim to be ``faithful by construction'', yet many fall short due to obstacles like lack of label leakage. As such, we need to be extra cautious about such claims. In essence, when deciding which explanation methods to use in practice, we advocate for those that are intrinsically motivated and empirically validated in terms of faithfulness, while still remaining aware of the potential pitfalls highlighted above. 

Finally, we discuss the common virtues and challenges of all methods and suggest potential directions for future research. In particular, we are eager to see future work on establishing a universal standard for faithfulness evaluation, exploring the relationship between interpretability and performance, and developing explanation methods that consider high-level features, flexible forms, and alternative task formats. We hope that this survey serves as an overview of the area for researchers interested in interpretability, and provides a practical guide for users seeking to better understand their models.
\appendix

\appendixsection{Additional Details}
\label{appendix:technical_details}

To complement the discussion of model explanation methods in Section~\ref{section:attempts}, here we provide interested readers with additional details about certain families of methods, including mathematical formalization, visualization of examples, and existing tools for implementation.

\subsection{Analysis of Model-Internal Structures}
\label{appendix:structure_analysis}

\minisection{Tools} For neuron visualization, a variety of visualization tools have been developed, including RNNvis\footnote{\url{https://www.myaooo.com/projects/rnnvis/}} \cite{ming_understanding_2017}, LSTMVis\footnote{\url{http://lstm.seas.harvard.edu/}} \cite{strobelt_lstmvis_2018}, and Seq2Seq-Vis\footnote{\url{https://seq2seq-vis.io/}} \cite{strobelt_seq2seq-vis_2019}. For attention visualization, readers can look into the following tools: BertViz\footnote{\url{https://github.com/jessevig/bertviz}} \cite{vig_visualizing_2019} and LIT\footnote{\url{https://pair-code.github.io/lit/}} \cite{tenney_language_2020}.

\subsection{Backpropagation-based Methods}
\label{appendix:bp_methods}

\begin{figure}[!t]
  \centering
  \includegraphics[width=\textwidth]{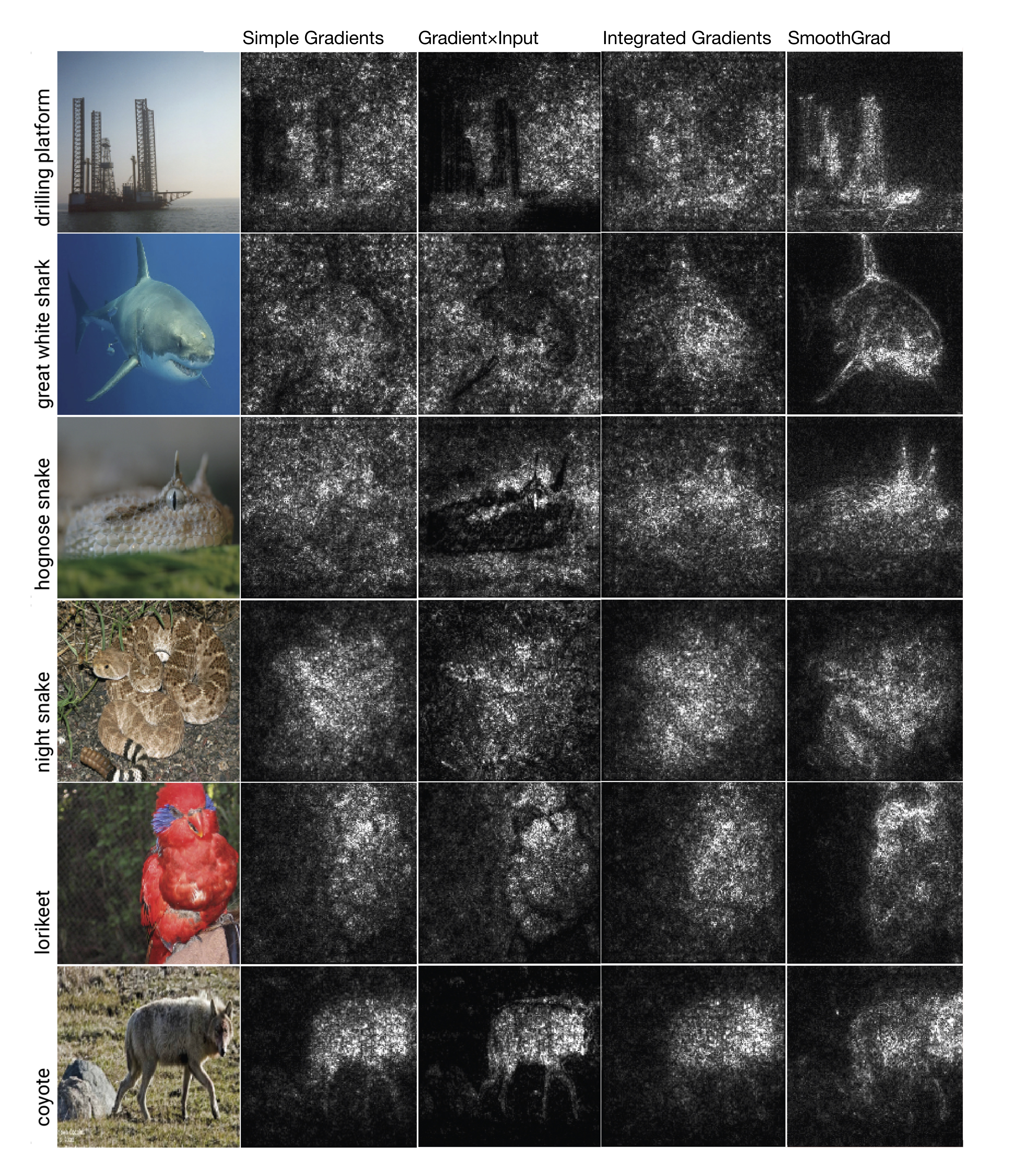}
  \caption{A visualization of different gradient methods on image classification examples (figure adapted from \cite{smilkov_smoothgrad_2017}). Brighter shades indicate higher feature relevance for the prediction.}
  \label{figure:grad_methods_visual}
\end{figure}

\minisection{Technical Details} Table~\ref{table:gradient_methods} and Table~\ref{table:propagation_methods} summarize the mathematical formalization of gradient methods and propagation methods respectively. Figure~\ref{figure:grad_methods_visual} shows a visualization of different gradient methods on image classification.

\begin{table*}[!t]
\small
\caption{Summary of different \textbf{gradient methods} in terms of how they compute $r_i(x)$, the relevance of feature $x_i$.  See Section~\ref{section:backprop_methods} -- Gradient methods for details on notations.}
\centering
\scalebox{1.0}{
\renewcommand{\arraystretch}{2}
\begin{tabular}{ll}
    \hline\hline \textbf{Method} & \textbf{Computation of $r_i(x)$} \\
    \hline 
    
    Simple Gradients & $ 
    \frac{\partial M(x)}{\partial x_i}$, $\norm{\frac{\partial M(x)}{\partial x_i}}_1 $, or $\norm{\frac{\partial M(x)}{\partial x_i}}_2$  \\
    
    Gradient$\times$Input & $x_i \odot \frac{\partial M(x)}{\partial x_i} $ \\
    
    \multirow{2}{*}{Integrated Gradients} & $ (x_i - \overline{x}_i) \odot \int_{\alpha=0}^{1} \frac{\partial M(\overline{x} + \alpha(x - \overline{x}))}{\partial x_i} d\alpha $ \\
    & approximated by $ (x_i - \overline{x}_i) \odot \sum_{\alpha=0}^{1} \frac{\partial M(\overline{x} + \alpha(x - \overline{x}))}{\partial x_i} $ \\
    
    \multirow{2}{*}{SmoothGrad} & $ \frac{1}{m} \sum_1^{m} \hat{r}_i(x) (x + \mathcal{N}(0, \sigma^2)) $ \\
    & where $\hat{r}_i(x)$ is any other relevance computation \\
    \hline\hline
\end{tabular}
}
\label{table:gradient_methods}
\end{table*}

\begin{table*}[!t]
\small
\caption[]{Summary of different propagation methods in terms of how they define the recursive function $D()$, as in $R_i^{(l)} = D(R_j^{(l+1)})$.\footnotemark\ Simple Gradients from gradient methods is included for comparison. See Section~\ref{section:backprop_methods} -- Propagation methods for details on notations.}
\centering
\renewcommand{\arraystretch}{2.6}
\scalebox{1.0}{
\begin{tabular}{ll}
    \hline\hline 
    \textbf{Method} & \textbf{Definition of $D()$} \\
    \hline
    Simple Gradients & $R_i^{(l)} = \sum_{j=0}^{d_{l+1}} \mathbbm{1}_{\sum_{i=0}^{d_l} a_iw_{ij} >0} \cdot R_j^{(l+1)}$ \vspace{0.05in} \\ 
    
    \hdashline
    DeconvNet & $ R_i^{(l)} = \sum_{j=0}^{d_{l+1}} \mathbbm{1}_{R_j^{(l+1)}>0} \cdot R_j^{(l+1)}$ \\
    
    Guided BackPropagation & $R_i^{(l)} = \sum_{j=0}^{d_{l+1}} \mathbbm{1}_{\sum_{i=0}^{d_l} a_iw_{ij} >0} \cdot \mathbbm{1}_{R_j^{(l+1)}>0} \cdot R_j^{(l+1)}$ \\
    
    Layerwise Relevance Propagation & $ R_i^{(l)} = \sum_{j=0}^{d_{l+1}} \frac{c_{ij}}{\sum_{i=0}^{d_{l}}c_{ij}} R_j^{(l+1)} $\\
    
    DeepLift & $R_i^{(l)} = \sum_{j=0}^{d_{l+1}} \frac{a_iw_{ij} - \overline{a_i}w_{ij}}{\sum_{i=0}^{d_l} (a_iw_{ij} - \overline{a_i}w_{ij})} R_j^{(l+1)} $\\
    
    Deep-Taylor Decomposition & $R_i^{(l)} = \sum_{j=0}^{d_{l+1}} \frac{\partial R_j^{(l+1)}}{\partial a_i} |_{\{\overline{a_i}\}^{(j)}} (a_i - \overline{a_i}^{(j)}) $\\
    \hline\hline

\end{tabular}
}
\renewcommand{\arraystretch}{1}
\label{table:propagation_methods}
\end{table*}

\footnotetext{Since $D()$ is layer-specific, we only show one or more representative rules for each method here: the ReLU unit propagation rule for Simple Gradients, DeconvNet, and GBP; the general-form rule for LRP; the Rescale rule For DeepLift; and the general-form rule for Deep-Taylor Expansion.}

\minisection{Tools} Readers interested in using backpropagation-based methods can consider the following packages: AllenNLP Interpret\footnote{\url{https://allenai.github.io/allennlp-website/interpret}} \cite{wallace_allennlp_2019}, Captum\footnote{\url{https://captum.ai/}} \cite{kokhlikyan_captum_2020}, RNNbow\footnote{\url{https://www.eecs.tufts.edu/~dcashm01/rnn_vis/d3_code/}} \cite{cashman_rnnbow_2018}, and DeepExplain\footnote{\url{https://github.com/marcoancona/DeepExplain}}.

\subsection{Counterfactual Intervention}
\label{appendix:counterfactual_intervention}

\minisection{Tools}  The following tools implement certain type(s) of counterfactual intervention: Captum\footnote{\url{https://captum.ai}}, LIT\footnote{\url{https://pair-code.github.io/lit}} \cite{tenney_language_2020}, LIME\footnote{\url{https://github.com/marcotcr/lime}} \cite{ribeiro_why_2016}, SHAP\footnote{\url{https://github.com/slundberg/shap}} \cite{lundberg_unified_2017}, Anchors\footnote{\url{https://github.com/marcotcr/anchor}} \cite{ribeiro_anchors_2018}, Seq2Seq-Vis\footnote{\url{https://seq2seq-vis.io}} \cite{strobelt_seq2seq-vis_2019}, and the What-if Tool\footnote{\url{https://pair-code.github.io/what-if-tool}} \cite{wexler_what-if_2020}.

\begin{acknowledgments}
This research is based upon work supported in part by the Air Force Research Laboratory (contract FA8750-23-C-0507), the DARPA KAIROS Program (contract FA8750-19-2-1004), the IARPA HIATUS Program (contract 2022-22072200005), and the NSF (Award 1928631). Approved for Public Release, Distribution Unlimited. The views and conclusions contained herein are those of the authors and should not be interpreted as necessarily representing the official policies, either expressed or implied, of AFRL, DARPA, IARPA, NSF, or the U.S. Government.
\end{acknowledgments}

\starttwocolumn    
\bibliography{custom}

\begin{thebibliography}{245}
\expandafter\ifx\csname natexlab\endcsname\relax\def\natexlab#1{#1}\fi

\bibitem[{Abnar and Zuidema(2020)}]{abnar_quantifying_2020}
Abnar, Samira and Willem Zuidema. 2020.
\newblock Quantifying attention flow in transformers.
\newblock In \emph{Proceedings of the 58th Annual Meeting of the Association for Computational Linguistics}, pages 4190--4197, Association for Computational Linguistics, Online.

\bibitem[{Abraham et~al.(2022)Abraham, D'Oosterlinck, Feder, Gat, Geiger, Potts, Reichart, and Wu}]{abraham_cebab_2022}
Abraham, Eldar~David, Karel D'Oosterlinck, Amir Feder, Yair~Ori Gat, Atticus Geiger, Christopher Potts, Roi Reichart, and Zhengxuan Wu. 2022.
\newblock {CEBaB}: {Estimating} the {Causal} {Effects} of {Real}-{World} {Concepts} on {NLP} {Model} {Behavior}.

\bibitem[{Adebayo et~al.(2018)Adebayo, Gilmer, Muelly, Goodfellow, Hardt, and Kim}]{adebayo_sanity_2018}
Adebayo, Julius, Justin Gilmer, Michael Muelly, Ian~J. Goodfellow, Moritz Hardt, and Been Kim. 2018.
\newblock Sanity checks for saliency maps.
\newblock In \emph{Advances in Neural Information Processing Systems 31: Annual Conference on Neural Information Processing Systems 2018, NeurIPS 2018, December 3-8, 2018, Montr{\'{e}}al, Canada}, pages 9525--9536.

\bibitem[{Adebayo et~al.(2020)Adebayo, Muelly, Liccardi, and Kim}]{adebayo_debugging_2020}
Adebayo, Julius, Michael Muelly, Ilaria Liccardi, and Been Kim. 2020.
\newblock Debugging tests for model explanations.
\newblock In \emph{Advances in Neural Information Processing Systems 33: Annual Conference on Neural Information Processing Systems 2020, NeurIPS 2020, December 6-12, 2020, virtual}.

\bibitem[{Adi et~al.(2017)Adi, Kermany, Belinkov, Lavi, and Goldberg}]{adi_fine-grained_2017}
Adi, Yossi, Einat Kermany, Yonatan Belinkov, Ofer Lavi, and Yoav Goldberg. 2017.
\newblock Fine-grained analysis of sentence embeddings using auxiliary prediction tasks.
\newblock In \emph{5th International Conference on Learning Representations, {ICLR} 2017, Toulon, France, April 24-26, 2017, Conference Track Proceedings}, OpenReview.net.

\bibitem[{Alvarez-Melis and Jaakkola(2018)}]{alvarez-melis_robustness_2018}
Alvarez-Melis, David and Tommi~S. Jaakkola. 2018.
\newblock On the {Robustness} of {Interpretability} {Methods}.
\newblock \emph{ArXiv preprint}, abs/1806.08049.

\bibitem[{Alvarez{-}Melis and Jaakkola(2018)}]{alvarez_melis_towards_2018}
Alvarez{-}Melis, David and Tommi~S. Jaakkola. 2018.
\newblock Towards robust interpretability with self-explaining neural networks.
\newblock In \emph{Advances in Neural Information Processing Systems 31: Annual Conference on Neural Information Processing Systems 2018, NeurIPS 2018, December 3-8, 2018, Montr{\'{e}}al, Canada}, pages 7786--7795.

\bibitem[{Amini et~al.(2022)Amini, Pimentel, Meister, and Cotterell}]{amini_naturalistic_2022}
Amini, Afra, Tiago Pimentel, Clara Meister, and Ryan Cotterell. 2022.
\newblock Naturalistic {Causal} {Probing} for {Morpho}-{Syntax}.

\bibitem[{Andreas et~al.(2016{\natexlab{a}})Andreas, Rohrbach, Darrell, and Klein}]{andreas_learning_2016}
Andreas, Jacob, Marcus Rohrbach, Trevor Darrell, and Dan Klein. 2016{\natexlab{a}}.
\newblock Learning to compose neural networks for question answering.
\newblock In \emph{Proceedings of the 2016 Conference of the North {A}merican Chapter of the Association for Computational Linguistics: Human Language Technologies}, pages 1545--1554, Association for Computational Linguistics, San Diego, California.

\bibitem[{Andreas et~al.(2016{\natexlab{b}})Andreas, Rohrbach, Darrell, and Klein}]{andreas_neural_2016}
Andreas, Jacob, Marcus Rohrbach, Trevor Darrell, and Dan Klein. 2016{\natexlab{b}}.
\newblock Neural module networks.
\newblock In \emph{2016 {IEEE} Conference on Computer Vision and Pattern Recognition, {CVPR} 2016, Las Vegas, NV, USA, June 27-30, 2016}, pages 39--48, {IEEE} Computer Society.

\bibitem[{Antol et~al.(2015)Antol, Agrawal, Lu, Mitchell, Batra, Zitnick, and Parikh}]{antol_vqa_2015}
Antol, Stanislaw, Aishwarya Agrawal, Jiasen Lu, Margaret Mitchell, Dhruv Batra, C.~Lawrence Zitnick, and Devi Parikh. 2015.
\newblock {VQA:} visual question answering.
\newblock In \emph{2015 {IEEE} International Conference on Computer Vision, {ICCV} 2015, Santiago, Chile, December 7-13, 2015}, pages 2425--2433, {IEEE} Computer Society.

\bibitem[{Arras et~al.(2016)Arras, Horn, Montavon, M{\"u}ller, and Samek}]{arras_explaining_2016}
Arras, Leila, Franziska Horn, Gr{\'e}goire Montavon, Klaus-Robert M{\"u}ller, and Wojciech Samek. 2016.
\newblock Explaining predictions of non-linear classifiers in {NLP}.
\newblock In \emph{Proceedings of the 1st Workshop on Representation Learning for {NLP}}, pages 1--7, Association for Computational Linguistics, Berlin, Germany.

\bibitem[{Arras et~al.(2017)Arras, Montavon, M{\"u}ller, and Samek}]{arras_explaining_2017}
Arras, Leila, Gr{\'e}goire Montavon, Klaus-Robert M{\"u}ller, and Wojciech Samek. 2017.
\newblock Explaining recurrent neural network predictions in sentiment analysis.
\newblock In \emph{Proceedings of the 8th Workshop on Computational Approaches to Subjectivity, Sentiment and Social Media Analysis}, pages 159--168, Association for Computational Linguistics, Copenhagen, Denmark.

\bibitem[{Atanasova et~al.(2020)Atanasova, Simonsen, Lioma, and Augenstein}]{atanasova_diagnostic_2020}
Atanasova, Pepa, Jakob~Grue Simonsen, Christina Lioma, and Isabelle Augenstein. 2020.
\newblock A diagnostic study of explainability techniques for text classification.
\newblock In \emph{Proceedings of the 2020 Conference on Empirical Methods in Natural Language Processing (EMNLP)}, pages 3256--3274, Association for Computational Linguistics, Online.

\bibitem[{Bach et~al.(2015)Bach, Binder, Montavon, Klauschen, M{\"u}ller, and Samek}]{bach_pixel-wise_2015}
Bach, Sebastian, Alexander Binder, Gr{\'e}goire Montavon, Frederick Klauschen, Klaus-Robert M{\"u}ller, and Wojciech Samek. 2015.
\newblock On {Pixel}-{Wise} {Explanations} for {Non}-{Linear} {Classifier} {Decisions} by {Layer}-{Wise} {Relevance} {Propagation}.
\newblock \emph{PLOS ONE}, 10(7):e0130140.
\newblock Publisher: Public Library of Science.

\bibitem[{Baehrens et~al.(2010)Baehrens, Schroeter, Harmeling, Kawanabe, and Hansen}]{baehrens_how_2010}
Baehrens, David, Timon Schroeter, Stefan Harmeling, Motoaki Kawanabe, and Katja Hansen. 2010.
\newblock How to {Explain} {Individual} {Classification} {Decisions}.
\newblock \emph{Journal of Machine Learning Research}, page~29.

\bibitem[{Bahdanau, Cho, and Bengio(2015)}]{bahdanau_neural_2015}
Bahdanau, Dzmitry, Kyunghyun Cho, and Yoshua Bengio. 2015.
\newblock Neural machine translation by jointly learning to align and translate.
\newblock In \emph{3rd International Conference on Learning Representations, {ICLR} 2015, San Diego, CA, USA, May 7-9, 2015, Conference Track Proceedings}.

\bibitem[{Bansal et~al.(2021)Bansal, Wu, Zhou, Fok, Nushi, Kamar, Ribeiro, and Weld}]{bansal_does_2021}
Bansal, Gagan, Tongshuang Wu, Joyce Zhou, Raymond Fok, Besmira Nushi, Ece Kamar, Marco~Tulio Ribeiro, and Daniel Weld. 2021.
\newblock Does the {Whole} {Exceed} its {Parts}? {The} {Effect} of {AI} {Explanations} on {Complementary} {Team} {Performance}.
\newblock In \emph{Proceedings of the 2021 {CHI} {Conference} on {Human} {Factors} in {Computing} {Systems}}, {CHI} '21, pages 1--16, Association for Computing Machinery, New York, NY, USA.

\bibitem[{Barredo~Arrieta et~al.(2020)Barredo~Arrieta, D{\'i}az-Rodr{\'i}guez, Del~Ser, Bennetot, Tabik, Barbado, Garcia, Gil-Lopez, Molina, Benjamins, Chatila, and Herrera}]{barredo_arrieta_explainable_2020}
Barredo~Arrieta, Alejandro, Natalia D{\'i}az-Rodr{\'i}guez, Javier Del~Ser, Adrien Bennetot, Siham Tabik, Alberto Barbado, Salvador Garcia, Sergio Gil-Lopez, Daniel Molina, Richard Benjamins, Raja Chatila, and Francisco Herrera. 2020.
\newblock Explainable {Artificial} {Intelligence} ({XAI}): {Concepts}, taxonomies, opportunities and challenges toward responsible {AI}.
\newblock \emph{Information Fusion}, 58:82--115.

\bibitem[{Bastings, Aziz, and Titov(2019)}]{bastings_interpretable_2019}
Bastings, Jasmijn, Wilker Aziz, and Ivan Titov. 2019.
\newblock Interpretable neural predictions with differentiable binary variables.
\newblock In \emph{Proceedings of the 57th Annual Meeting of the Association for Computational Linguistics}, pages 2963--2977, Association for Computational Linguistics, Florence, Italy.

\bibitem[{Bastings et~al.(2022)Bastings, Ebert, Zablotskaia, Sandholm, and Filippova}]{bastings_will_2021}
Bastings, Jasmijn, Sebastian Ebert, Polina Zablotskaia, Anders Sandholm, and Katja Filippova. 2022.
\newblock {``}will you find these shortcuts?{''} a protocol for evaluating the faithfulness of input salience methods for text classification.
\newblock In \emph{Proceedings of the 2022 Conference on Empirical Methods in Natural Language Processing}, pages 976--991, Association for Computational Linguistics, Abu Dhabi, United Arab Emirates.

\bibitem[{Bastings and Filippova(2020)}]{bastings_elephant_2020}
Bastings, Jasmijn and Katja Filippova. 2020.
\newblock The elephant in the interpretability room: Why use attention as explanation when we have saliency methods?
\newblock In \emph{Proceedings of the Third BlackboxNLP Workshop on Analyzing and Interpreting Neural Networks for NLP}, pages 149--155, Association for Computational Linguistics, Online.

\bibitem[{Basu, Pope, and Feizi(2021)}]{basu_influence_2020}
Basu, Samyadeep, Phillip Pope, and Soheil Feizi. 2021.
\newblock Influence functions in deep learning are fragile.
\newblock In \emph{9th International Conference on Learning Representations, {ICLR} 2021, Virtual Event, Austria, May 3-7, 2021}, OpenReview.net.

\bibitem[{Bau et~al.(2019)Bau, Belinkov, Sajjad, Durrani, Dalvi, and Glass}]{bau_identifying_2019}
Bau, Anthony, Yonatan Belinkov, Hassan Sajjad, Nadir Durrani, Fahim Dalvi, and James~R. Glass. 2019.
\newblock Identifying and controlling important neurons in neural machine translation.
\newblock In \emph{7th International Conference on Learning Representations, {ICLR} 2019, New Orleans, LA, USA, May 6-9, 2019}, OpenReview.net.

\bibitem[{Belinkov et~al.(2020)Belinkov, Durrani, Dalvi, Sajjad, and Glass}]{belinkov_linguistic_2020}
Belinkov, Yonatan, Nadir Durrani, Fahim Dalvi, Hassan Sajjad, and James Glass. 2020.
\newblock On the linguistic representational power of neural machine translation models.
\newblock \emph{Computational Linguistics}, 46(1):1--52.

\bibitem[{Belinkov and Glass(2019)}]{belinkov_analysis_2019}
Belinkov, Yonatan and James Glass. 2019.
\newblock Analysis methods in neural language processing: A survey.
\newblock \emph{Transactions of the Association for Computational Linguistics}, 7:49--72.

\bibitem[{Bogin et~al.(2021)Bogin, Subramanian, Gardner, and Berant}]{bogin_latent_2021}
Bogin, Ben, Sanjay Subramanian, Matt Gardner, and Jonathan Berant. 2021.
\newblock Latent compositional representations improve systematic generalization in grounded question answering.
\newblock \emph{Transactions of the Association for Computational Linguistics}, 9:195--210.

\bibitem[{Bommasani et~al.(2021)Bommasani, Hudson, Adeli, Altman, Arora, von Arx, Bernstein, Bohg, Bosselut, Brunskill, Brynjolfsson, Buch, Card, Castellon, Chatterji, Chen, Creel, Davis, Demszky, Donahue, Doumbouya, Durmus, Ermon, Etchemendy, Ethayarajh, Fei-Fei, Finn, Gale, Gillespie, Goel, Goodman, Grossman, Guha, Hashimoto, Henderson, Hewitt, Ho, Hong, Hsu, Huang, Icard, Jain, Jurafsky, Kalluri, Karamcheti, Keeling, Khani, Khattab, Koh, Krass, Krishna, Kuditipudi, Kumar, Ladhak, Lee, Lee, Leskovec, Levent, Li, Li, Ma, Malik, Manning, Mirchandani, Mitchell, Munyikwa, Nair, Narayan, Narayanan, Newman, Nie, Niebles, Nilforoshan, Nyarko, Ogut, Orr, Papadimitriou, Park, Piech, Portelance, Potts, Raghunathan, Reich, Ren, Rong, Roohani, Ruiz, Ryan, R{\'e}, Sadigh, Sagawa, Santhanam, Shih, Srinivasan, Tamkin, Taori, Thomas, Tram{\`e}r, Wang, Wang, Wu, Wu, Wu, Xie, Yasunaga, You, Zaharia, Zhang, Zhang, Zhang, Zhang, Zheng, Zhou, and Liang}]{bommasani_opportunities_2021}
Bommasani, Rishi, Drew~A. Hudson, Ehsan Adeli, Russ Altman, Simran Arora, Sydney von Arx, Michael~S. Bernstein, Jeannette Bohg, Antoine Bosselut, Emma Brunskill, Erik Brynjolfsson, Shyamal Buch, Dallas Card, Rodrigo Castellon, Niladri Chatterji, Annie Chen, Kathleen Creel, Jared~Quincy Davis, Dora Demszky, Chris Donahue, Moussa Doumbouya, Esin Durmus, Stefano Ermon, John Etchemendy, Kawin Ethayarajh, Li~Fei-Fei, Chelsea Finn, Trevor Gale, Lauren Gillespie, Karan Goel, Noah Goodman, Shelby Grossman, Neel Guha, Tatsunori Hashimoto, Peter Henderson, John Hewitt, Daniel~E. Ho, Jenny Hong, Kyle Hsu, Jing Huang, Thomas Icard, Saahil Jain, Dan Jurafsky, Pratyusha Kalluri, Siddharth Karamcheti, Geoff Keeling, Fereshte Khani, Omar Khattab, Pang~Wei Koh, Mark Krass, Ranjay Krishna, Rohith Kuditipudi, Ananya Kumar, Faisal Ladhak, Mina Lee, Tony Lee, Jure Leskovec, Isabelle Levent, Xiang~Lisa Li, Xuechen Li, Tengyu Ma, Ali Malik, Christopher~D. Manning, Suvir Mirchandani, Eric Mitchell, Zanele Munyikwa, Suraj Nair,
  Avanika Narayan, Deepak Narayanan, Ben Newman, Allen Nie, Juan~Carlos Niebles, Hamed Nilforoshan, Julian Nyarko, Giray Ogut, Laurel Orr, Isabel Papadimitriou, Joon~Sung Park, Chris Piech, Eva Portelance, Christopher Potts, Aditi Raghunathan, Rob Reich, Hongyu Ren, Frieda Rong, Yusuf Roohani, Camilo Ruiz, Jack Ryan, Christopher R{\'e}, Dorsa Sadigh, Shiori Sagawa, Keshav Santhanam, Andy Shih, Krishnan Srinivasan, Alex Tamkin, Rohan Taori, Armin~W. Thomas, Florian Tram{\`e}r, Rose~E. Wang, William Wang, Bohan Wu, Jiajun Wu, Yuhuai Wu, Sang~Michael Xie, Michihiro Yasunaga, Jiaxuan You, Matei Zaharia, Michael Zhang, Tianyi Zhang, Xikun Zhang, Yuhui Zhang, Lucia Zheng, Kaitlyn Zhou, and Percy Liang. 2021.
\newblock On the {Opportunities} and {Risks} of {Foundation} {Models}.
\newblock \emph{ArXiv preprint}, abs/2108.07258.

\bibitem[{Bowman et~al.(2015)Bowman, Angeli, Potts, and Manning}]{bowman_large_2015}
Bowman, Samuel~R., Gabor Angeli, Christopher Potts, and Christopher~D. Manning. 2015.
\newblock A large annotated corpus for learning natural language inference.
\newblock In \emph{Proceedings of the 2015 Conference on Empirical Methods in Natural Language Processing}, pages 632--642, Association for Computational Linguistics, Lisbon, Portugal.

\bibitem[{Brown et~al.(2020)Brown, Mann, Ryder, Subbiah, Kaplan, Dhariwal, Neelakantan, Shyam, Sastry, Askell, Agarwal, Herbert{-}Voss, Krueger, Henighan, Child, Ramesh, Ziegler, Wu, Winter, Hesse, Chen, Sigler, Litwin, Gray, Chess, Clark, Berner, McCandlish, Radford, Sutskever, and Amodei}]{brown_language_2020}
Brown, Tom~B., Benjamin Mann, Nick Ryder, Melanie Subbiah, Jared Kaplan, Prafulla Dhariwal, Arvind Neelakantan, Pranav Shyam, Girish Sastry, Amanda Askell, Sandhini Agarwal, Ariel Herbert{-}Voss, Gretchen Krueger, Tom Henighan, Rewon Child, Aditya Ramesh, Daniel~M. Ziegler, Jeffrey Wu, Clemens Winter, Christopher Hesse, Mark Chen, Eric Sigler, Mateusz Litwin, Scott Gray, Benjamin Chess, Jack Clark, Christopher Berner, Sam McCandlish, Alec Radford, Ilya Sutskever, and Dario Amodei. 2020.
\newblock Language models are few-shot learners.
\newblock In \emph{Advances in Neural Information Processing Systems 33: Annual Conference on Neural Information Processing Systems 2020, NeurIPS 2020, December 6-12, 2020, virtual}.

\bibitem[{Brunner et~al.(2018)Brunner, Wang, Wattenhofer, and Weigelt}]{brunner_natural_2018}
Brunner, Gino, Yuyi Wang, Roger Wattenhofer, and Michael Weigelt. 2018.
\newblock Natural {Language} {Multitasking}: {Analyzing} and {Improving} {Syntactic} {Saliency} of {Hidden} {Representations}.

\bibitem[{Calderon et~al.(2022)Calderon, Ben-David, Feder, and Reichart}]{calderon_docogen_2022}
Calderon, Nitay, Eyal Ben-David, Amir Feder, and Roi Reichart. 2022.
\newblock {D}o{C}o{G}en: {D}omain counterfactual generation for low resource domain adaptation.
\newblock In \emph{Proceedings of the 60th Annual Meeting of the Association for Computational Linguistics (Volume 1: Long Papers)}, pages 7727--7746, Association for Computational Linguistics, Dublin, Ireland.

\bibitem[{Camburu et~al.(2018)Camburu, Rockt{\"{a}}schel, Lukasiewicz, and Blunsom}]{camburu_e-snli_2018}
Camburu, Oana{-}Maria, Tim Rockt{\"{a}}schel, Thomas Lukasiewicz, and Phil Blunsom. 2018.
\newblock e-snli: Natural language inference with natural language explanations.
\newblock In \emph{Advances in Neural Information Processing Systems 31: Annual Conference on Neural Information Processing Systems 2018, NeurIPS 2018, December 3-8, 2018, Montr{\'{e}}al, Canada}, pages 9560--9572.

\bibitem[{Camburu et~al.(2020)Camburu, Shillingford, Minervini, Lukasiewicz, and Blunsom}]{camburu_make_2020}
Camburu, Oana-Maria, Brendan Shillingford, Pasquale Minervini, Thomas Lukasiewicz, and Phil Blunsom. 2020.
\newblock Make up your mind! adversarial generation of inconsistent natural language explanations.
\newblock In \emph{Proceedings of the 58th Annual Meeting of the Association for Computational Linguistics}, pages 4157--4165, Association for Computational Linguistics, Online.

\bibitem[{Caruana et~al.(1999)Caruana, Kangarloo, Dionisio, Sinha, and Johnson}]{caruana_case-based_1999}
Caruana, R., H.~Kangarloo, J.~D. Dionisio, U.~Sinha, and D.~Johnson. 1999.
\newblock Case-based explanation of non-case-based learning methods.
\newblock \emph{Proceedings of the AMIA Symposium}, pages 212--215.

\bibitem[{Cashman et~al.(2018)Cashman, Patterson, Mosca, Watts, Robinson, and Chang}]{cashman_rnnbow_2018}
Cashman, Dylan, Genevi{\`e}ve Patterson, Abigail Mosca, Nathan Watts, Shannon Robinson, and Remco Chang. 2018.
\newblock {RNNbow}: {Visualizing} {Learning} {Via} {Backpropagation} {Gradients} in {RNNs}.
\newblock \emph{IEEE Computer Graphics and Applications}, 38(6):39--50.
\newblock Conference Name: IEEE Computer Graphics and Applications.

\bibitem[{Caucheteux and King(2022)}]{caucheteux_brains_2022}
Caucheteux, Charlotte and Jean-R{\'e}mi King. 2022.
\newblock Brains and algorithms partially converge in natural language processing.
\newblock \emph{Communications Biology}, 5(1):134.

\bibitem[{Chan, Kong, and Guanqing(2022)}]{chan_comparative_2022}
Chan, Chun~Sik, Huanqi Kong, and Liang Guanqing. 2022.
\newblock A comparative study of faithfulness metrics for model interpretability methods.
\newblock In \emph{Proceedings of the 60th Annual Meeting of the Association for Computational Linguistics (Volume 1: Long Papers)}, pages 5029--5038, Association for Computational Linguistics, Dublin, Ireland.

\bibitem[{Chefer, Gur, and Wolf(2021)}]{chefer_transformer_2021}
Chefer, Hila, Shir Gur, and Lior Wolf. 2021.
\newblock Transformer interpretability beyond attention visualization.
\newblock In \emph{{IEEE} Conference on Computer Vision and Pattern Recognition, {CVPR} 2021, virtual, June 19-25, 2021}, pages 782--791, Computer Vision Foundation / {IEEE}.

\bibitem[{Chen et~al.(2022{\natexlab{a}})Chen, Brahman, Ren, Ji, Choi, and Swayamdipta}]{chen_rev_2022}
Chen, Hanjie, Faeze Brahman, Xiang Ren, Yangfeng Ji, Yejin Choi, and Swabha Swayamdipta. 2022{\natexlab{a}}.
\newblock {REV}: {Information}-{Theoretic} {Evaluation} of {Free}-{Text} {Rationales}.

\bibitem[{Chen et~al.(2018)Chen, Song, Wainwright, and Jordan}]{chen_learning_2018}
Chen, Jianbo, Le~Song, Martin~J. Wainwright, and Michael~I. Jordan. 2018.
\newblock Learning to explain: An information-theoretic perspective on model interpretation.
\newblock In \emph{Proceedings of the 35th International Conference on Machine Learning, {ICML} 2018, Stockholmsm{\"{a}}ssan, Stockholm, Sweden, July 10-15, 2018}, volume~80 of \emph{Proceedings of Machine Learning Research}, pages 882--891, {PMLR}.

\bibitem[{Chen et~al.(2022{\natexlab{b}})Chen, Ma, Wang, and Cohen}]{chen_program_2022}
Chen, Wenhu, Xueguang Ma, Xinyi Wang, and William~W. Cohen. 2022{\natexlab{b}}.
\newblock Program of {Thoughts} {Prompting}: {Disentangling} {Computation} from {Reasoning} for {Numerical} {Reasoning} {Tasks}.

\bibitem[{Choenni, Shutova, and van Rooij(2021)}]{choenni_stepmothers_2021}
Choenni, Rochelle, Ekaterina Shutova, and Robert van Rooij. 2021.
\newblock Stepmothers are mean and academics are pretentious: What do pretrained language models learn about you?
\newblock In \emph{Proceedings of the 2021 Conference on Empirical Methods in Natural Language Processing}, pages 1477--1491, Association for Computational Linguistics, Online and Punta Cana, Dominican Republic.

\bibitem[{Chomsky(1965)}]{chomsky_aspects_1965}
Chomsky, Noam. 1965.
\newblock \emph{Aspects of the theory of syntax}.
\newblock Aspects of the theory of syntax. M.I.T. Press, Oxford, England.

\bibitem[{Clark et~al.(2020)Clark, Luong, Le, and Manning}]{clark_electra_2020}
Clark, Kevin, Minh{-}Thang Luong, Quoc~V. Le, and Christopher~D. Manning. 2020.
\newblock {ELECTRA:} pre-training text encoders as discriminators rather than generators.
\newblock In \emph{8th International Conference on Learning Representations, {ICLR} 2020, Addis Ababa, Ethiopia, April 26-30, 2020}, OpenReview.net.

\bibitem[{Clark et~al.(2018)Clark, Cowhey, Etzioni, Khot, Sabharwal, Schoenick, and Tafjord}]{clark_think_2018}
Clark, Peter, Isaac Cowhey, Oren Etzioni, Tushar Khot, Ashish Sabharwal, Carissa Schoenick, and Oyvind Tafjord. 2018.
\newblock Think you have {Solved} {Question} {Answering}? {Try} {ARC}, the {AI2} {Reasoning} {Challenge}.

\bibitem[{Clinciu, Eshghi, and Hastie(2021)}]{clinciu_study_2021}
Clinciu, Miruna-Adriana, Arash Eshghi, and Helen Hastie. 2021.
\newblock A study of automatic metrics for the evaluation of natural language explanations.
\newblock In \emph{Proceedings of the 16th Conference of the European Chapter of the Association for Computational Linguistics: Main Volume}, pages 2376--2387, Association for Computational Linguistics, Online.

\bibitem[{Cobbe et~al.(2021)Cobbe, Kosaraju, Bavarian, Chen, Jun, Kaiser, Plappert, Tworek, Hilton, Nakano, Hesse, and Schulman}]{cobbe_training_2021}
Cobbe, Karl, Vineet Kosaraju, Mohammad Bavarian, Mark Chen, Heewoo Jun, Lukasz Kaiser, Matthias Plappert, Jerry Tworek, Jacob Hilton, Reiichiro Nakano, Christopher Hesse, and John Schulman. 2021.
\newblock Training {Verifiers} to {Solve} {Math} {Word} {Problems}.

\bibitem[{Conneau et~al.(2018)Conneau, Kruszewski, Lample, Barrault, and Baroni}]{conneau_what_2018}
Conneau, Alexis, German Kruszewski, Guillaume Lample, Lo{\"\i}c Barrault, and Marco Baroni. 2018.
\newblock What you can cram into a single {\$}{\&}!{\#}* vector: Probing sentence embeddings for linguistic properties.
\newblock In \emph{Proceedings of the 56th Annual Meeting of the Association for Computational Linguistics (Volume 1: Long Papers)}, pages 2126--2136, Association for Computational Linguistics, Melbourne, Australia.

\bibitem[{Creswell and Shanahan(2022)}]{creswell_faithful_2022}
Creswell, Antonia and Murray Shanahan. 2022.
\newblock Faithful {Reasoning} {Using} {Large} {Language} {Models}.

\bibitem[{Dalvi et~al.(2021)Dalvi, Jansen, Tafjord, Xie, Smith, Pipatanangkura, and Clark}]{dalvi_explaining_2021}
Dalvi, Bhavana, Peter Jansen, Oyvind Tafjord, Zhengnan Xie, Hannah Smith, Leighanna Pipatanangkura, and Peter Clark. 2021.
\newblock Explaining answers with entailment trees.
\newblock In \emph{Proceedings of the 2021 Conference on Empirical Methods in Natural Language Processing}, pages 7358--7370, Association for Computational Linguistics, Online and Punta Cana, Dominican Republic.

\bibitem[{Dalvi et~al.(2022)Dalvi, Khan, Alam, Durrani, Xu, and Sajjad}]{dalvi_discovering_2022}
Dalvi, Fahim, Abdul~Rafae Khan, Firoj Alam, Nadir Durrani, Jia Xu, and Hassan Sajjad. 2022.
\newblock Discovering latent concepts learned in {BERT}.
\newblock In \emph{The Tenth International Conference on Learning Representations, {ICLR} 2022, Virtual Event, April 25-29, 2022}, OpenReview.net.

\bibitem[{Danilevsky et~al.(2020)Danilevsky, Qian, Aharonov, Katsis, Kawas, and Sen}]{danilevsky-etal-2020-survey}
Danilevsky, Marina, Kun Qian, Ranit Aharonov, Yannis Katsis, Ban Kawas, and Prithviraj Sen. 2020.
\newblock A survey of the state of explainable {AI} for natural language processing.
\newblock In \emph{Proceedings of the 1st Conference of the Asia-Pacific Chapter of the Association for Computational Linguistics and the 10th International Joint Conference on Natural Language Processing}, pages 447--459, Association for Computational Linguistics, Suzhou, China.

\bibitem[{De~Cao et~al.(2020)De~Cao, Schlichtkrull, Aziz, and Titov}]{de_cao_how_2020}
De~Cao, Nicola, Michael~Sejr Schlichtkrull, Wilker Aziz, and Ivan Titov. 2020.
\newblock How do decisions emerge across layers in neural models? interpretation with differentiable masking.
\newblock In \emph{Proceedings of the 2020 Conference on Empirical Methods in Natural Language Processing (EMNLP)}, pages 3243--3255, Association for Computational Linguistics, Online.

\bibitem[{De~Cao et~al.(2022)De~Cao, Schmid, Hupkes, and Titov}]{de_cao_sparse_2022}
De~Cao, Nicola, Leon Schmid, Dieuwke Hupkes, and Ivan Titov. 2022.
\newblock Sparse interventions in language models with differentiable masking.
\newblock In \emph{Proceedings of the Fifth BlackboxNLP Workshop on Analyzing and Interpreting Neural Networks for NLP}, pages 16--27, Association for Computational Linguistics, Abu Dhabi, United Arab Emirates (Hybrid).

\bibitem[{Denil, Demiraj, and de~Freitas(2015)}]{denil_extraction_2015}
Denil, Misha, Alban Demiraj, and Nando de~Freitas. 2015.
\newblock Extraction of {Salient} {Sentences} from {Labelled} {Documents}.
\newblock \emph{arXiv:1412.6815 [cs]}.
\newblock ArXiv: 1412.6815.

\bibitem[{Deutsch, Upadhyay, and Roth(2019)}]{deutsch_general-purpose_2019}
Deutsch, Daniel, Shyam Upadhyay, and Dan Roth. 2019.
\newblock A general-purpose algorithm for constrained sequential inference.
\newblock In \emph{Proceedings of the 23rd Conference on Computational Natural Language Learning (CoNLL)}, pages 482--492, Association for Computational Linguistics, Hong Kong, China.

\bibitem[{Devlin et~al.(2019)Devlin, Chang, Lee, and Toutanova}]{devlin_bert_2019}
Devlin, Jacob, Ming-Wei Chang, Kenton Lee, and Kristina Toutanova. 2019.
\newblock {BERT}: Pre-training of deep bidirectional transformers for language understanding.
\newblock In \emph{Proceedings of the 2019 Conference of the North {A}merican Chapter of the Association for Computational Linguistics: Human Language Technologies, Volume 1 (Long and Short Papers)}, pages 4171--4186, Association for Computational Linguistics, Minneapolis, Minnesota.

\bibitem[{DeYoung et~al.(2020)DeYoung, Jain, Rajani, Lehman, Xiong, Socher, and Wallace}]{deyoung_eraser_2020}
DeYoung, Jay, Sarthak Jain, Nazneen~Fatema Rajani, Eric Lehman, Caiming Xiong, Richard Socher, and Byron~C. Wallace. 2020.
\newblock {ERASER}: {A} benchmark to evaluate rationalized {NLP} models.
\newblock In \emph{Proceedings of the 58th Annual Meeting of the Association for Computational Linguistics}, pages 4443--4458, Association for Computational Linguistics, Online.

\bibitem[{Ding and Koehn(2021)}]{ding_evaluating_2021}
Ding, Shuoyang and Philipp Koehn. 2021.
\newblock Evaluating saliency methods for neural language models.
\newblock In \emph{Proceedings of the 2021 Conference of the North American Chapter of the Association for Computational Linguistics: Human Language Technologies}, pages 5034--5052, Association for Computational Linguistics, Online.

\bibitem[{Doshi-Velez and Kim(2017)}]{doshi-velez_towards_2017}
Doshi-Velez, Finale and Been Kim. 2017.
\newblock Towards {A} {Rigorous} {Science} of {Interpretable} {Machine} {Learning}.
\newblock \emph{ArXiv preprint}, abs/1702.08608.

\bibitem[{Dua et~al.(2019)Dua, Wang, Dasigi, Stanovsky, Singh, and Gardner}]{dua_drop_2019}
Dua, Dheeru, Yizhong Wang, Pradeep Dasigi, Gabriel Stanovsky, Sameer Singh, and Matt Gardner. 2019.
\newblock {DROP}: A reading comprehension benchmark requiring discrete reasoning over paragraphs.
\newblock In \emph{Proceedings of the 2019 Conference of the North {A}merican Chapter of the Association for Computational Linguistics: Human Language Technologies, Volume 1 (Long and Short Papers)}, pages 2368--2378, Association for Computational Linguistics, Minneapolis, Minnesota.

\bibitem[{Eberle et~al.(2022)Eberle, Brandl, Pilot, and S{\o}gaard}]{eberle_transformer_2022}
Eberle, Oliver, Stephanie Brandl, Jonas Pilot, and Anders S{\o}gaard. 2022.
\newblock Do transformer models show similar attention patterns to task-specific human gaze?
\newblock In \emph{Proceedings of the 60th Annual Meeting of the Association for Computational Linguistics (Volume 1: Long Papers)}, pages 4295--4309, Association for Computational Linguistics, Dublin, Ireland.

\bibitem[{Ebrahimi et~al.(2018)Ebrahimi, Rao, Lowd, and Dou}]{ebrahimi_hotflip_2018}
Ebrahimi, Javid, Anyi Rao, Daniel Lowd, and Dejing Dou. 2018.
\newblock {H}ot{F}lip: White-box adversarial examples for text classification.
\newblock In \emph{Proceedings of the 56th Annual Meeting of the Association for Computational Linguistics (Volume 2: Short Papers)}, pages 31--36, Association for Computational Linguistics, Melbourne, Australia.

\bibitem[{Elazar et~al.(2021)Elazar, Ravfogel, Jacovi, and Goldberg}]{elazar_amnesic_2021}
Elazar, Yanai, Shauli Ravfogel, Alon Jacovi, and Yoav Goldberg. 2021.
\newblock Amnesic probing: Behavioral explanation with amnesic counterfactuals.
\newblock \emph{Transactions of the Association for Computational Linguistics}, 9:160--175.

\bibitem[{Ethayarajh(2019)}]{ethayarajh_how_2019}
Ethayarajh, Kawin. 2019.
\newblock How contextual are contextualized word representations? {C}omparing the geometry of {BERT}, {ELM}o, and {GPT}-2 embeddings.
\newblock In \emph{Proceedings of the 2019 Conference on Empirical Methods in Natural Language Processing and the 9th International Joint Conference on Natural Language Processing (EMNLP-IJCNLP)}, pages 55--65, Association for Computational Linguistics, Hong Kong, China.

\bibitem[{Ethayarajh and Jurafsky(2021)}]{ethayarajh_attention_2021}
Ethayarajh, Kawin and Dan Jurafsky. 2021.
\newblock Attention flows are shapley value explanations.
\newblock In \emph{Proceedings of the 59th Annual Meeting of the Association for Computational Linguistics and the 11th International Joint Conference on Natural Language Processing (Volume 2: Short Papers)}, pages 49--54, Association for Computational Linguistics, Online.

\bibitem[{Feder et~al.(2022)Feder, Keith, Manzoor, Pryzant, Sridhar, Wood-Doughty, Eisenstein, Grimmer, Reichart, Roberts, Stewart, Veitch, and Yang}]{feder_causal_2022}
Feder, Amir, Katherine~A. Keith, Emaad Manzoor, Reid Pryzant, Dhanya Sridhar, Zach Wood-Doughty, Jacob Eisenstein, Justin Grimmer, Roi Reichart, Margaret~E. Roberts, Brandon~M. Stewart, Victor Veitch, and Diyi Yang. 2022.
\newblock Causal inference in natural language processing: Estimation, prediction, interpretation and beyond.
\newblock \emph{Transactions of the Association for Computational Linguistics}, 10:1138--1158.

\bibitem[{Feder et~al.(2021)Feder, Oved, Shalit, and Reichart}]{feder_causalm_2021}
Feder, Amir, Nadav Oved, Uri Shalit, and Roi Reichart. 2021.
\newblock {C}ausa{LM}: Causal model explanation through counterfactual language models.
\newblock \emph{Computational Linguistics}, 47(2):333--386.

\bibitem[{Feng et~al.(2018)Feng, Wallace, Grissom~II, Iyyer, Rodriguez, and Boyd-Graber}]{feng_pathologies_2018}
Feng, Shi, Eric Wallace, Alvin Grissom~II, Mohit Iyyer, Pedro Rodriguez, and Jordan Boyd-Graber. 2018.
\newblock Pathologies of neural models make interpretations difficult.
\newblock In \emph{Proceedings of the 2018 Conference on Empirical Methods in Natural Language Processing}, pages 3719--3728, Association for Computational Linguistics, Brussels, Belgium.

\bibitem[{Finlayson et~al.(2021)Finlayson, Mueller, Gehrmann, Shieber, Linzen, and Belinkov}]{finlayson_causal_2021}
Finlayson, Matthew, Aaron Mueller, Sebastian Gehrmann, Stuart Shieber, Tal Linzen, and Yonatan Belinkov. 2021.
\newblock Causal analysis of syntactic agreement mechanisms in neural language models.
\newblock In \emph{Proceedings of the 59th Annual Meeting of the Association for Computational Linguistics and the 11th International Joint Conference on Natural Language Processing (Volume 1: Long Papers)}, pages 1828--1843, Association for Computational Linguistics, Online.

\bibitem[{Gao et~al.(2022)Gao, Madaan, Zhou, Alon, Liu, Yang, Callan, and Neubig}]{gao_pal_2023}
Gao, Luyu, Aman Madaan, Shuyan Zhou, Uri Alon, Pengfei Liu, Yiming Yang, Jamie Callan, and Graham Neubig. 2022.
\newblock {PAL}: {Program}-aided {Language} {Models}.

\bibitem[{Gardner et~al.(2020)Gardner, Artzi, Basmov, Berant, Bogin, Chen, Dasigi, Dua, Elazar, Gottumukkala, Gupta, Hajishirzi, Ilharco, Khashabi, Lin, Liu, Liu, Mulcaire, Ning, Singh, Smith, Subramanian, Tsarfaty, Wallace, Zhang, and Zhou}]{gardner_evaluating_2020}
Gardner, Matt, Yoav Artzi, Victoria Basmov, Jonathan Berant, Ben Bogin, Sihao Chen, Pradeep Dasigi, Dheeru Dua, Yanai Elazar, Ananth Gottumukkala, Nitish Gupta, Hannaneh Hajishirzi, Gabriel Ilharco, Daniel Khashabi, Kevin Lin, Jiangming Liu, Nelson~F. Liu, Phoebe Mulcaire, Qiang Ning, Sameer Singh, Noah~A. Smith, Sanjay Subramanian, Reut Tsarfaty, Eric Wallace, Ally Zhang, and Ben Zhou. 2020.
\newblock Evaluating models{'} local decision boundaries via contrast sets.
\newblock In \emph{Findings of the Association for Computational Linguistics: EMNLP 2020}, pages 1307--1323, Association for Computational Linguistics, Online.

\bibitem[{Geva, Goldberg, and Berant(2019)}]{geva_are_2019}
Geva, Mor, Yoav Goldberg, and Jonathan Berant. 2019.
\newblock Are we modeling the task or the annotator? an investigation of annotator bias in natural language understanding datasets.
\newblock In \emph{Proceedings of the 2019 Conference on Empirical Methods in Natural Language Processing and the 9th International Joint Conference on Natural Language Processing (EMNLP-IJCNLP)}, pages 1161--1166, Association for Computational Linguistics, Hong Kong, China.

\bibitem[{Ghorbani, Abid, and Zou(2019)}]{ghorbani_interpretation_2019}
Ghorbani, Amirata, Abubakar Abid, and James~Y. Zou. 2019.
\newblock Interpretation of neural networks is fragile.
\newblock In \emph{The Thirty-Third {AAAI} Conference on Artificial Intelligence, {AAAI} 2019, The Thirty-First Innovative Applications of Artificial Intelligence Conference, {IAAI} 2019, The Ninth {AAAI} Symposium on Educational Advances in Artificial Intelligence, {EAAI} 2019, Honolulu, Hawaii, USA, January 27 - February 1, 2019}, pages 3681--3688, {AAAI} Press.

\bibitem[{Gupta et~al.(2020)Gupta, Lin, Roth, Singh, and Gardner}]{gupta_neural_2019}
Gupta, Nitish, Kevin Lin, Dan Roth, Sameer Singh, and Matt Gardner. 2020.
\newblock Neural module networks for reasoning over text.
\newblock In \emph{8th International Conference on Learning Representations, {ICLR} 2020, Addis Ababa, Ethiopia, April 26-30, 2020}, OpenReview.net.

\bibitem[{Haghighatkhah et~al.(2022)Haghighatkhah, Fokkens, Sommerauer, Speckmann, and Verbeek}]{haghighatkhah_better_2022}
Haghighatkhah, Pantea, Antske Fokkens, Pia Sommerauer, Bettina Speckmann, and Kevin Verbeek. 2022.
\newblock Better hit the nail on the head than beat around the bush: Removing protected attributes with a single projection.
\newblock In \emph{Proceedings of the 2022 Conference on Empirical Methods in Natural Language Processing}, pages 8395--8416, Association for Computational Linguistics, Abu Dhabi, United Arab Emirates.

\bibitem[{Halpern and Pearl(2005)}]{halpern_causes_2005}
Halpern, Joseph~Y. and Judea Pearl. 2005.
\newblock Causes and {Explanations}: {A} {Structural}-{Model} {Approach}. {Part} {I}: {Causes}.
\newblock \emph{The British Journal for the Philosophy of Science}, 56(4):843--887.
\newblock Publisher: The University of Chicago Press.

\bibitem[{Han, Wallace, and Tsvetkov(2020)}]{han_explaining_2020}
Han, Xiaochuang, Byron~C. Wallace, and Yulia Tsvetkov. 2020.
\newblock Explaining black box predictions and unveiling data artifacts through influence functions.
\newblock In \emph{Proceedings of the 58th Annual Meeting of the Association for Computational Linguistics}, pages 5553--5563, Association for Computational Linguistics, Online.

\bibitem[{Hao et~al.(2021)Hao, Dong, Wei, and Xu}]{hao_self-attention_2021}
Hao, Yaru, Li~Dong, Furu Wei, and Ke~Xu. 2021.
\newblock Self-attention attribution: Interpreting information interactions inside transformer.
\newblock In \emph{Thirty-Fifth {AAAI} Conference on Artificial Intelligence, {AAAI} 2021, Thirty-Third Conference on Innovative Applications of Artificial Intelligence, {IAAI} 2021, The Eleventh Symposium on Educational Advances in Artificial Intelligence, {EAAI} 2021, Virtual Event, February 2-9, 2021}, pages 12963--12971, {AAAI} Press.

\bibitem[{Harrington et~al.(1985)Harrington, Morley, {\v S}cedrov, and Simpson}]{harrington_harvey_1985}
Harrington, L.~A., M.~D. Morley, A.~{\v S}cedrov, and S.~G. Simpson. 1985.
\newblock \emph{Harvey {Friedman}'s {Research} on the {Foundations} of {Mathematics}}.
\newblock Elsevier.
\newblock Google-Books-ID: 2plPRR4LDxIC.

\bibitem[{Hase and Bansal(2020)}]{hase_evaluating_2020}
Hase, Peter and Mohit Bansal. 2020.
\newblock Evaluating explainable {AI}: Which algorithmic explanations help users predict model behavior?
\newblock In \emph{Proceedings of the 58th Annual Meeting of the Association for Computational Linguistics}, pages 5540--5552, Association for Computational Linguistics, Online.

\bibitem[{Hase and Bansal(2022)}]{hase_when_2022}
Hase, Peter and Mohit Bansal. 2022.
\newblock When can models learn from explanations? a formal framework for understanding the roles of explanation data.
\newblock In \emph{Proceedings of the First Workshop on Learning with Natural Language Supervision}, pages 29--39, Association for Computational Linguistics, Dublin, Ireland.

\bibitem[{Hase et~al.(2020)Hase, Zhang, Xie, and Bansal}]{hase_leakage-adjusted_2020}
Hase, Peter, Shiyue Zhang, Harry Xie, and Mohit Bansal. 2020.
\newblock Leakage-adjusted simulatability: Can models generate non-trivial explanations of their behavior in natural language?
\newblock In \emph{Findings of the Association for Computational Linguistics: EMNLP 2020}, pages 4351--4367, Association for Computational Linguistics, Online.

\bibitem[{Hendricks et~al.(2016)Hendricks, Akata, Rohrbach, Donahue, Schiele, and Darrell}]{hendricks_generating_2016}
Hendricks, Lisa~Anne, Zeynep Akata, Marcus Rohrbach, Jeff Donahue, Bernt Schiele, and Trevor Darrell. 2016.
\newblock Generating {Visual} {Explanations}.
\newblock In \emph{Computer {Vision} {\textendash} {ECCV} 2016}, Lecture {Notes} in {Computer} {Science}, pages 3--19, Springer International Publishing, Cham.

\bibitem[{Hendrycks et~al.(2021)Hendrycks, Burns, Basart, Zou, Mazeika, Song, and Steinhardt}]{hendrycks_measuring_2021}
Hendrycks, Dan, Collin Burns, Steven Basart, Andy Zou, Mantas Mazeika, Dawn Song, and Jacob Steinhardt. 2021.
\newblock Measuring massive multitask language understanding.
\newblock In \emph{9th International Conference on Learning Representations, {ICLR} 2021, Virtual Event, Austria, May 3-7, 2021}, OpenReview.net.

\bibitem[{Herman(2017)}]{herman_promise_2019}
Herman, Bernease. 2017.
\newblock The {Promise} and {Peril} of {Human} {Evaluation} for {Model} {Interpretability}.
\newblock \emph{ArXiv preprint}, abs/1711.07414.

\bibitem[{Hewitt and Liang(2019)}]{hewitt_designing_2019}
Hewitt, John and Percy Liang. 2019.
\newblock Designing and interpreting probes with control tasks.
\newblock In \emph{Proceedings of the 2019 Conference on Empirical Methods in Natural Language Processing and the 9th International Joint Conference on Natural Language Processing (EMNLP-IJCNLP)}, pages 2733--2743, Association for Computational Linguistics, Hong Kong, China.

\bibitem[{Hiebert et~al.(2018)Hiebert, Peterson, Fyshe, and Mehta}]{hiebert_interpreting_2018}
Hiebert, Avery, Cole Peterson, Alona Fyshe, and Nishant Mehta. 2018.
\newblock Interpreting word-level hidden state behaviour of character-level {LSTM} language models.
\newblock In \emph{Proceedings of the 2018 {EMNLP} Workshop {B}lackbox{NLP}: Analyzing and Interpreting Neural Networks for {NLP}}, pages 258--266, Association for Computational Linguistics, Brussels, Belgium.

\bibitem[{Hong et~al.(2022)Hong, Zhang, Yu, and Zhang}]{hong_metgen_2022}
Hong, Ruixin, Hongming Zhang, Xintong Yu, and Changshui Zhang. 2022.
\newblock {METGEN}: A module-based entailment tree generation framework for answer explanation.
\newblock In \emph{Findings of the Association for Computational Linguistics: NAACL 2022}, pages 1887--1905, Association for Computational Linguistics, Seattle, United States.

\bibitem[{Hooker et~al.(2019)Hooker, Erhan, Kindermans, and Kim}]{hooker_benchmark_2019}
Hooker, Sara, Dumitru Erhan, Pieter{-}Jan Kindermans, and Been Kim. 2019.
\newblock A benchmark for interpretability methods in deep neural networks.
\newblock In \emph{Advances in Neural Information Processing Systems 32: Annual Conference on Neural Information Processing Systems 2019, NeurIPS 2019, December 8-14, 2019, Vancouver, BC, Canada}, pages 9734--9745.

\bibitem[{Hu et~al.(2017)Hu, Andreas, Rohrbach, Darrell, and Saenko}]{hu_learning_2017}
Hu, Ronghang, Jacob Andreas, Marcus Rohrbach, Trevor Darrell, and Kate Saenko. 2017.
\newblock Learning to reason: End-to-end module networks for visual question answering.
\newblock In \emph{{IEEE} International Conference on Computer Vision, {ICCV} 2017, Venice, Italy, October 22-29, 2017}, pages 804--813, {IEEE} Computer Society.

\bibitem[{Jacovi and Goldberg(2020)}]{jacovi_towards_2020}
Jacovi, Alon and Yoav Goldberg. 2020.
\newblock Towards faithfully interpretable {NLP} systems: How should we define and evaluate faithfulness?
\newblock In \emph{Proceedings of the 58th Annual Meeting of the Association for Computational Linguistics}, pages 4198--4205, Association for Computational Linguistics, Online.

\bibitem[{Jacovi and Goldberg(2021)}]{jacovi_aligning_2021}
Jacovi, Alon and Yoav Goldberg. 2021.
\newblock Aligning faithful interpretations with their social attribution.
\newblock \emph{Transactions of the Association for Computational Linguistics}, 9:294--310.

\bibitem[{Jacovi et~al.(2021)Jacovi, Swayamdipta, Ravfogel, Elazar, Choi, and Goldberg}]{jacovi_contrastive_2021}
Jacovi, Alon, Swabha Swayamdipta, Shauli Ravfogel, Yanai Elazar, Yejin Choi, and Yoav Goldberg. 2021.
\newblock Contrastive explanations for model interpretability.
\newblock In \emph{Proceedings of the 2021 Conference on Empirical Methods in Natural Language Processing}, pages 1597--1611, Association for Computational Linguistics, Online and Punta Cana, Dominican Republic.

\bibitem[{Jain and Wallace(2019)}]{jain_attention_2019}
Jain, Sarthak and Byron~C. Wallace. 2019.
\newblock {A}ttention is not {E}xplanation.
\newblock In \emph{Proceedings of the 2019 Conference of the North {A}merican Chapter of the Association for Computational Linguistics: Human Language Technologies, Volume 1 (Long and Short Papers)}, pages 3543--3556, Association for Computational Linguistics, Minneapolis, Minnesota.

\bibitem[{Jain et~al.(2020)Jain, Wiegreffe, Pinter, and Wallace}]{jain_learning_2020}
Jain, Sarthak, Sarah Wiegreffe, Yuval Pinter, and Byron~C. Wallace. 2020.
\newblock {L}earning to faithfully rationalize by construction.
\newblock In \emph{Proceedings of the 58th Annual Meeting of the Association for Computational Linguistics}, pages 4459--4473, Association for Computational Linguistics, Online.

\bibitem[{Janizek, Sturmfels, and Lee(2021)}]{janizek_explaining_2021}
Janizek, Joseph~D., Pascal Sturmfels, and Su{-}In Lee. 2021.
\newblock Explaining explanations: Axiomatic feature interactions for deep networks.
\newblock \emph{J. Mach. Learn. Res.}, 22:104:1--104:54.

\bibitem[{Jiang et~al.(2020)Jiang, Zhao, Chu, Shen, and Tu}]{jiang_cold-start_2020}
Jiang, Chengyue, Yinggong Zhao, Shanbo Chu, Libin Shen, and Kewei Tu. 2020.
\newblock Cold-start and interpretability: Turning regular expressions into trainable recurrent neural networks.
\newblock In \emph{Proceedings of the 2020 Conference on Empirical Methods in Natural Language Processing (EMNLP)}, pages 3193--3207, Association for Computational Linguistics, Online.

\bibitem[{Jiang et~al.(2019)Jiang, Joshi, Chen, and Bansal}]{jiang_explore_2019}
Jiang, Yichen, Nitish Joshi, Yen-Chun Chen, and Mohit Bansal. 2019.
\newblock Explore, propose, and assemble: An interpretable model for multi-hop reading comprehension.
\newblock In \emph{Proceedings of the 57th Annual Meeting of the Association for Computational Linguistics}, pages 2714--2725, Association for Computational Linguistics, Florence, Italy.

\bibitem[{Johnson et~al.(2017)Johnson, Hariharan, van~der Maaten, Fei{-}Fei, Zitnick, and Girshick}]{johnson_clevr_2017}
Johnson, Justin, Bharath Hariharan, Laurens van~der Maaten, Li~Fei{-}Fei, C.~Lawrence Zitnick, and Ross~B. Girshick. 2017.
\newblock {CLEVR:} {A} diagnostic dataset for compositional language and elementary visual reasoning.
\newblock In \emph{2017 {IEEE} Conference on Computer Vision and Pattern Recognition, {CVPR} 2017, Honolulu, HI, USA, July 21-26, 2017}, pages 1988--1997, {IEEE} Computer Society.

\bibitem[{Ju et~al.(2022)Ju, Zhang, Yang, Jiang, Liu, and Zhao}]{ju_logic_2022}
Ju, Yiming, Yuanzhe Zhang, Zhao Yang, Zhongtao Jiang, Kang Liu, and Jun Zhao. 2022.
\newblock Logic traps in evaluating attribution scores.
\newblock In \emph{Proceedings of the 60th Annual Meeting of the Association for Computational Linguistics (Volume 1: Long Papers)}, pages 5911--5922, Association for Computational Linguistics, Dublin, Ireland.

\bibitem[{Jung et~al.(2022)Jung, Qin, Welleck, Brahman, Bhagavatula, Le~Bras, and Choi}]{jung_maieutic_2022}
Jung, Jaehun, Lianhui Qin, Sean Welleck, Faeze Brahman, Chandra Bhagavatula, Ronan Le~Bras, and Yejin Choi. 2022.
\newblock Maieutic prompting: Logically consistent reasoning with recursive explanations.
\newblock In \emph{Proceedings of the 2022 Conference on Empirical Methods in Natural Language Processing}, pages 1266--1279, Association for Computational Linguistics, Abu Dhabi, United Arab Emirates.

\bibitem[{K{\'a}d{\'a}r, Chrupa{\l}a, and Alishahi(2017)}]{kadar_representation_2017}
K{\'a}d{\'a}r, {\'A}kos, Grzegorz Chrupa{\l}a, and Afra Alishahi. 2017.
\newblock Representation of linguistic form and function in recurrent neural networks.
\newblock \emph{Computational Linguistics}, 43(4):761--780.

\bibitem[{Karidi et~al.(2021)Karidi, Zhou, Schneider, Abend, and Srikumar}]{karidi_putting_2021}
Karidi, Taelin, Yichu Zhou, Nathan Schneider, Omri Abend, and Vivek Srikumar. 2021.
\newblock Putting words in {BERT}{'}s mouth: Navigating contextualized vector spaces with pseudowords.
\newblock In \emph{Proceedings of the 2021 Conference on Empirical Methods in Natural Language Processing}, pages 10300--10313, Association for Computational Linguistics, Online and Punta Cana, Dominican Republic.

\bibitem[{Karpathy, Johnson, and Fei-Fei(2015)}]{karpathy_visualizing_2015}
Karpathy, Andrej, Justin Johnson, and Li~Fei-Fei. 2015.
\newblock Visualizing and {Understanding} {Recurrent} {Networks}.
\newblock \emph{ArXiv preprint}, abs/1506.02078.

\bibitem[{Kassner et~al.(2021)Kassner, Tafjord, Sch{\"u}tze, and Clark}]{kassner_beliefbank_2021}
Kassner, Nora, Oyvind Tafjord, Hinrich Sch{\"u}tze, and Peter Clark. 2021.
\newblock {B}elief{B}ank: Adding memory to a pre-trained language model for a systematic notion of belief.
\newblock In \emph{Proceedings of the 2021 Conference on Empirical Methods in Natural Language Processing}, pages 8849--8861, Association for Computational Linguistics, Online and Punta Cana, Dominican Republic.

\bibitem[{Kaushik, Hovy, and Lipton(2020)}]{kaushik_learning_2020}
Kaushik, Divyansh, Eduard~H. Hovy, and Zachary~Chase Lipton. 2020.
\newblock Learning the difference that makes {A} difference with counterfactually-augmented data.
\newblock In \emph{8th International Conference on Learning Representations, {ICLR} 2020, Addis Ababa, Ethiopia, April 26-30, 2020}, OpenReview.net.

\bibitem[{Kaushik and Lipton(2018)}]{kaushik_how_2018}
Kaushik, Divyansh and Zachary~C. Lipton. 2018.
\newblock How much reading does reading comprehension require? a critical investigation of popular benchmarks.
\newblock In \emph{Proceedings of the 2018 Conference on Empirical Methods in Natural Language Processing}, pages 5010--5015, Association for Computational Linguistics, Brussels, Belgium.

\bibitem[{Kim et~al.(2018)Kim, Wattenberg, Gilmer, Cai, Wexler, Vi{\'{e}}gas, and Sayres}]{kim_interpretability_2018}
Kim, Been, Martin Wattenberg, Justin Gilmer, Carrie~J. Cai, James Wexler, Fernanda~B. Vi{\'{e}}gas, and Rory Sayres. 2018.
\newblock Interpretability beyond feature attribution: Quantitative testing with concept activation vectors {(TCAV)}.
\newblock In \emph{Proceedings of the 35th International Conference on Machine Learning, {ICML} 2018, Stockholmsm{\"{a}}ssan, Stockholm, Sweden, July 10-15, 2018}, volume~80 of \emph{Proceedings of Machine Learning Research}, pages 2673--2682, {PMLR}.

\bibitem[{Kindermans et~al.(2019)Kindermans, Hooker, Adebayo, Alber, Sch{\"u}tt, D{\"a}hne, Erhan, and Kim}]{samek_reliability_2019}
Kindermans, Pieter-Jan, Sara Hooker, Julius Adebayo, Maximilian Alber, Kristof~T. Sch{\"u}tt, Sven D{\"a}hne, Dumitru Erhan, and Been Kim. 2019.
\newblock The ({Un})reliability of {Saliency} {Methods}.
\newblock In Wojciech Samek, Gr{\'e}goire Montavon, Andrea Vedaldi, Lars~Kai Hansen, and Klaus-Robert M{\"u}ller, editors, \emph{Explainable {AI}: {Interpreting}, {Explaining} and {Visualizing} {Deep} {Learning}}, volume 11700. Springer International Publishing, Cham, pages 267--280.
\newblock Series Title: Lecture Notes in Computer Science.

\bibitem[{Kindermans et~al.(2018)Kindermans, Sch{\"{u}}tt, Alber, M{\"{u}}ller, Erhan, Kim, and D{\"{a}}hne}]{kindermans_learning_2017}
Kindermans, Pieter{-}Jan, Kristof~T. Sch{\"{u}}tt, Maximilian Alber, Klaus{-}Robert M{\"{u}}ller, Dumitru Erhan, Been Kim, and Sven D{\"{a}}hne. 2018.
\newblock Learning how to explain neural networks: Patternnet and patternattribution.
\newblock In \emph{6th International Conference on Learning Representations, {ICLR} 2018, Vancouver, BC, Canada, April 30 - May 3, 2018, Conference Track Proceedings}, OpenReview.net.

\bibitem[{Koh and Liang(2017)}]{koh_understanding_2017}
Koh, Pang~Wei and Percy Liang. 2017.
\newblock Understanding black-box predictions via influence functions.
\newblock In \emph{Proceedings of the 34th International Conference on Machine Learning, {ICML} 2017, Sydney, NSW, Australia, 6-11 August 2017}, volume~70 of \emph{Proceedings of Machine Learning Research}, pages 1885--1894, {PMLR}.

\bibitem[{Kojima et~al.(2022)Kojima, Gu, Reid, Matsuo, and Iwasawa}]{kojima_large_2022}
Kojima, Takeshi, Shixiang~Shane Gu, Machel Reid, Yutaka Matsuo, and Yusuke Iwasawa. 2022.
\newblock Large {Language} {Models} are {Zero}-{Shot} {Reasoners}.

\bibitem[{Kokhlikyan et~al.(2020)Kokhlikyan, Miglani, Martin, Wang, Alsallakh, Reynolds, Melnikov, Kliushkina, Araya, Yan, and Reblitz-Richardson}]{kokhlikyan_captum_2020}
Kokhlikyan, Narine, Vivek Miglani, Miguel Martin, Edward Wang, Bilal Alsallakh, Jonathan Reynolds, Alexander Melnikov, Natalia Kliushkina, Carlos Araya, Siqi Yan, and Orion Reblitz-Richardson. 2020.
\newblock Captum: {A} unified and generic model interpretability library for {PyTorch}.
\newblock \emph{ArXiv preprint}, abs/2009.07896.

\bibitem[{Krishnamurthy and Kollar(2013)}]{krishnamurthy_jointly_2013}
Krishnamurthy, Jayant and Thomas Kollar. 2013.
\newblock Jointly learning to parse and perceive: Connecting natural language to the physical world.
\newblock \emph{Transactions of the Association for Computational Linguistics}, 1:193--206.

\bibitem[{Kumar, Tan, and Sharma(2022)}]{kumar_probing_2022}
Kumar, Abhinav, Chenhao Tan, and Amit Sharma. 2022.
\newblock Probing {Classifiers} are {Unreliable} for {Concept} {Removal} and {Detection}.

\bibitem[{Kumar and Talukdar(2020)}]{kumar_nile_2020}
Kumar, Sawan and Partha Talukdar. 2020.
\newblock {NILE} : Natural language inference with faithful natural language explanations.
\newblock In \emph{Proceedings of the 58th Annual Meeting of the Association for Computational Linguistics}, pages 8730--8742, Association for Computational Linguistics, Online.

\bibitem[{Kunkel et~al.(2019)Kunkel, Donkers, Michael, Barbu, and Ziegler}]{kunkel_let_2019}
Kunkel, Johannes, Tim Donkers, Lisa Michael, Catalin{-}Mihai Barbu, and J{\"{u}}rgen Ziegler. 2019.
\newblock Let me explain: Impact of personal and impersonal explanations on trust in recommender systems.
\newblock In \emph{Proceedings of the 2019 {CHI} Conference on Human Factors in Computing Systems, {CHI} 2019, Glasgow, Scotland, UK, May 04-09, 2019}, page 487, {ACM}.

\bibitem[{Lakkaraju and Bastani(2020)}]{lakkaraju_how_2020}
Lakkaraju, Himabindu and Osbert Bastani. 2020.
\newblock "{How} do {I} fool you?": {Manipulating} {User} {Trust} via {Misleading} {Black} {Box} {Explanations}.
\newblock In \emph{Proceedings of the {AAAI}/{ACM} {Conference} on {AI}, {Ethics}, and {Society}}, pages 79--85, ACM, New York NY USA.

\bibitem[{Lampinen et~al.(2022)Lampinen, Dasgupta, Chan, Mathewson, Tessler, Creswell, McClelland, Wang, and Hill}]{lampinen_can_2022}
Lampinen, Andrew, Ishita Dasgupta, Stephanie Chan, Kory Mathewson, Mh~Tessler, Antonia Creswell, James McClelland, Jane Wang, and Felix Hill. 2022.
\newblock Can language models learn from explanations in context?
\newblock In \emph{Findings of the Association for Computational Linguistics: EMNLP 2022}, pages 537--563, Association for Computational Linguistics, Abu Dhabi, United Arab Emirates.

\bibitem[{Laugel et~al.(2018)Laugel, Renard, Lesot, Marsala, and Detyniecki}]{laugel_defining_2018}
Laugel, Thibault, Xavier Renard, Marie-Jeanne Lesot, Christophe Marsala, and Marcin Detyniecki. 2018.
\newblock Defining {Locality} for {Surrogates} in {Post}-hoc {Interpretablity}.

\bibitem[{Lei, Barzilay, and Jaakkola(2016)}]{lei_rationalizing_2016}
Lei, Tao, Regina Barzilay, and Tommi Jaakkola. 2016.
\newblock Rationalizing neural predictions.
\newblock In \emph{Proceedings of the 2016 Conference on Empirical Methods in Natural Language Processing}, pages 107--117, Association for Computational Linguistics, Austin, Texas.

\bibitem[{Levesque, Davis, and Morgenstern(2012)}]{levesque_winograd_2012}
Levesque, Hector, Ernest Davis, and Leora Morgenstern. 2012.
\newblock The {Winograd} {Schema} {Challenge}.
\newblock page~10.

\bibitem[{Lewkowycz et~al.(2022)Lewkowycz, Andreassen, Dohan, Dyer, Michalewski, Ramasesh, Slone, Anil, Schlag, Gutman-Solo, Wu, Neyshabur, Gur-Ari, and Misra}]{lewkowycz_solving_2022}
Lewkowycz, Aitor, Anders Andreassen, David Dohan, Ethan Dyer, Henryk Michalewski, Vinay Ramasesh, Ambrose Slone, Cem Anil, Imanol Schlag, Theo Gutman-Solo, Yuhuai Wu, Behnam Neyshabur, Guy Gur-Ari, and Vedant Misra. 2022.
\newblock Solving {Quantitative} {Reasoning} {Problems} with {Language} {Models}.

\bibitem[{Li et~al.(2020)Li, Liu, Li, Li, Huang, and Shi}]{li_evaluating_2020}
Li, Jierui, Lemao Liu, Huayang Li, Guanlin Li, Guoping Huang, and Shuming Shi. 2020.
\newblock Evaluating explanation methods for neural machine translation.
\newblock In \emph{Proceedings of the 58th Annual Meeting of the Association for Computational Linguistics}, pages 365--375, Association for Computational Linguistics, Online.

\bibitem[{Li et~al.(2016)Li, Chen, Hovy, and Jurafsky}]{li_visualizing_2016}
Li, Jiwei, Xinlei Chen, Eduard Hovy, and Dan Jurafsky. 2016.
\newblock Visualizing and understanding neural models in {NLP}.
\newblock In \emph{Proceedings of the 2016 Conference of the North {A}merican Chapter of the Association for Computational Linguistics: Human Language Technologies}, pages 681--691, Association for Computational Linguistics, San Diego, California.

\bibitem[{Li, Monroe, and Jurafsky(2016)}]{li_understanding_2017}
Li, Jiwei, Will Monroe, and Dan Jurafsky. 2016.
\newblock Understanding {Neural} {Networks} through {Representation} {Erasure}.
\newblock \emph{ArXiv preprint}, abs/1612.08220.

\bibitem[{Li et~al.(2022)Li, Lin, Zhang, Fu, Chen, Lou, and Chen}]{li_advance_2022}
Li, Yifei, Zeqi Lin, Shizhuo Zhang, Qiang Fu, Bei Chen, Jian-Guang Lou, and Weizhu Chen. 2022.
\newblock On the {Advance} of {Making} {Language} {Models} {Better} {Reasoners}.

\bibitem[{Ling et~al.(2017)Ling, Yogatama, Dyer, and Blunsom}]{ling_program_2017}
Ling, Wang, Dani Yogatama, Chris Dyer, and Phil Blunsom. 2017.
\newblock Program induction by rationale generation: Learning to solve and explain algebraic word problems.
\newblock In \emph{Proceedings of the 55th Annual Meeting of the Association for Computational Linguistics (Volume 1: Long Papers)}, pages 158--167, Association for Computational Linguistics, Vancouver, Canada.

\bibitem[{Lipton(2016)}]{lipton_mythos_2017}
Lipton, Zachary~C. 2016.
\newblock The {Mythos} of {Model} {Interpretability}.
\newblock \emph{ArXiv preprint}, abs/1606.03490.

\bibitem[{Liu et~al.(2022)Liu, Li, Guo, Kong, Li, and Wang}]{liu_rethinking_2022}
Liu, Yibing, Haoliang Li, Yangyang Guo, Chenqi Kong, Jing Li, and Shiqi Wang. 2022.
\newblock Rethinking attention-model explainability through faithfulness violation test.
\newblock In \emph{International Conference on Machine Learning, {ICML} 2022, 17-23 July 2022, Baltimore, Maryland, {USA}}, volume 162 of \emph{Proceedings of Machine Learning Research}, pages 13807--13824, {PMLR}.

\bibitem[{Liu et~al.(2019)Liu, Ott, Goyal, Du, Joshi, Chen, Levy, Lewis, Zettlemoyer, and Stoyanov}]{liu_roberta_2019}
Liu, Yinhan, Myle Ott, Naman Goyal, Jingfei Du, Mandar Joshi, Danqi Chen, Omer Levy, Mike Lewis, Luke Zettlemoyer, and Veselin Stoyanov. 2019.
\newblock {RoBERTa}: {A} {Robustly} {Optimized} {BERT} {Pretraining} {Approach}.
\newblock \emph{ArXiv preprint}, abs/1907.11692.

\bibitem[{Lovering et~al.(2020)Lovering, Jha, Linzen, and Pavlick}]{lovering_information-theoretic_2020}
Lovering, Charles, Rohan Jha, Tal Linzen, and Ellie Pavlick. 2020.
\newblock Information-theoretic {Probing} {Explains} {Reliance} on {Spurious} {Features}.

\bibitem[{Lu et~al.(2021)Lu, Wang, Mardziel, and Datta}]{lu_uence_2021}
Lu, Kaiji, Zifan Wang, Piotr Mardziel, and Anupam Datta. 2021.
\newblock Influence patterns for explaining information flow in {BERT}.
\newblock In \emph{Advances in Neural Information Processing Systems 34: Annual Conference on Neural Information Processing Systems 2021, NeurIPS 2021, December 6-14, 2021, virtual}, pages 4461--4474.

\bibitem[{Lundberg and Lee(2017)}]{lundberg_unified_2017}
Lundberg, Scott~M. and Su{-}In Lee. 2017.
\newblock A unified approach to interpreting model predictions.
\newblock In \emph{Advances in Neural Information Processing Systems 30: Annual Conference on Neural Information Processing Systems 2017, December 4-9, 2017, Long Beach, CA, {USA}}, pages 4765--4774.

\bibitem[{Lyu et~al.(2023)Lyu, Havaldar, Stein, Zhang, Rao, Wong, Apidianaki, and Callison-Burch}]{lyu_faithful_2023}
Lyu, Qing, Shreya Havaldar, Adam Stein, Li~Zhang, Delip Rao, Eric Wong, Marianna Apidianaki, and Chris Callison-Burch. 2023.
\newblock Faithful {Chain}-of-{Thought} {Reasoning}.

\bibitem[{Madaan et~al.(2021)Madaan, Tandon, Rajagopal, Yang, Clark, Sakaguchi, and Hovy}]{madaan_improving_2021}
Madaan, Aman, Niket Tandon, Dheeraj Rajagopal, Yiming Yang, Peter Clark, Keisuke Sakaguchi, and Ed~Hovy. 2021.
\newblock Improving {Neural} {Model} {Performance} through {Natural} {Language} {Feedback} on {Their} {Explanations}.
\newblock \emph{ArXiv preprint}, abs/2104.08765.

\bibitem[{Mao et~al.(2019)Mao, Gan, Kohli, Tenenbaum, and Wu}]{mao_neuro-symbolic_2019}
Mao, Jiayuan, Chuang Gan, Pushmeet Kohli, Joshua~B. Tenenbaum, and Jiajun Wu. 2019.
\newblock The neuro-symbolic concept learner: Interpreting scenes, words, and sentences from natural supervision.
\newblock In \emph{7th International Conference on Learning Representations, {ICLR} 2019, New Orleans, LA, USA, May 6-9, 2019}, OpenReview.net.

\bibitem[{Marasovic et~al.(2022)Marasovic, Beltagy, Downey, and Peters}]{marasovic_few-shot_2022}
Marasovic, Ana, Iz~Beltagy, Doug Downey, and Matthew Peters. 2022.
\newblock Few-shot self-rationalization with natural language prompts.
\newblock In \emph{Findings of the Association for Computational Linguistics: NAACL 2022}, pages 410--424, Association for Computational Linguistics, Seattle, United States.

\bibitem[{Martins and Astudillo(2016)}]{martins_softmax_2016}
Martins, Andr{\'{e}} F.~T. and Ram{\'{o}}n~Fernandez Astudillo. 2016.
\newblock From softmax to sparsemax: {A} sparse model of attention and multi-label classification.
\newblock In \emph{Proceedings of the 33nd International Conference on Machine Learning, {ICML} 2016, New York City, NY, USA, June 19-24, 2016}, volume~48 of \emph{{JMLR} Workshop and Conference Proceedings}, pages 1614--1623, JMLR.org.

\bibitem[{McCoy, Pavlick, and Linzen(2019)}]{mccoy_right_2019}
McCoy, Tom, Ellie Pavlick, and Tal Linzen. 2019.
\newblock Right for the wrong reasons: Diagnosing syntactic heuristics in natural language inference.
\newblock In \emph{Proceedings of the 57th Annual Meeting of the Association for Computational Linguistics}, pages 3428--3448, Association for Computational Linguistics, Florence, Italy.

\bibitem[{Miller(2017)}]{miller_explanation_2018}
Miller, Tim. 2017.
\newblock Explanation in {Artificial} {Intelligence}: {Insights} from the {Social} {Sciences}.
\newblock \emph{ArXiv preprint}, abs/1706.07269.

\bibitem[{Ming et~al.(2017)Ming, Cao, Zhang, Li, Chen, Song, and Qu}]{ming_understanding_2017}
Ming, Yao, Shaozu Cao, Ruixiang Zhang, Zhen Li, Yuanzhe Chen, Yangqiu Song, and Huamin Qu. 2017.
\newblock Understanding {Hidden} {Memories} of {Recurrent} {Neural} {Networks}.
\newblock In \emph{2017 {IEEE} {Conference} on {Visual} {Analytics} {Science} and {Technology} ({VAST})}, pages 13--24.

\bibitem[{Montavon et~al.(2019)Montavon, Binder, Lapuschkin, Samek, and M{\"u}ller}]{montavon_layer-wise_2019}
Montavon, Gr{\'e}goire, Alexander Binder, Sebastian Lapuschkin, Wojciech Samek, and Klaus-Robert M{\"u}ller. 2019.
\newblock Layer-{Wise} {Relevance} {Propagation}: {An} {Overview}.
\newblock In Wojciech Samek, Gr{\'e}goire Montavon, Andrea Vedaldi, Lars~Kai Hansen, and Klaus-Robert M{\"u}ller, editors, \emph{Explainable {AI}: {Interpreting}, {Explaining} and {Visualizing} {Deep} {Learning}}, Lecture {Notes} in {Computer} {Science}. Springer International Publishing, Cham, pages 193--209.

\bibitem[{Montavon et~al.(2017)Montavon, Lapuschkin, Binder, Samek, and M{\"u}ller}]{montavon_explaining_2017}
Montavon, Gr{\'e}goire, Sebastian Lapuschkin, Alexander Binder, Wojciech Samek, and Klaus-Robert M{\"u}ller. 2017.
\newblock Explaining nonlinear classification decisions with deep {Taylor} decomposition.
\newblock \emph{Pattern Recognition}, 65:211--222.

\bibitem[{Mosca et~al.(2022)Mosca, Szigeti, Tragianni, Gallagher, and Groh}]{mosca_shap-based_2022}
Mosca, Edoardo, Ferenc Szigeti, Stella Tragianni, Daniel Gallagher, and Georg Groh. 2022.
\newblock {SHAP}-based explanation methods: A review for {NLP} interpretability.
\newblock In \emph{Proceedings of the 29th International Conference on Computational Linguistics}, pages 4593--4603, International Committee on Computational Linguistics, Gyeongju, Republic of Korea.

\bibitem[{Mueller, Xia, and Linzen(2022)}]{mueller_causal_2022}
Mueller, Aaron, Yu~Xia, and Tal Linzen. 2022.
\newblock Causal analysis of syntactic agreement neurons in multilingual language models.
\newblock In \emph{Proceedings of the 26th Conference on Computational Natural Language Learning (CoNLL)}, pages 95--109, Association for Computational Linguistics, Abu Dhabi, United Arab Emirates (Hybrid).

\bibitem[{Mullenbach et~al.(2018)Mullenbach, Wiegreffe, Duke, Sun, and Eisenstein}]{mullenbach_explainable_2018}
Mullenbach, James, Sarah Wiegreffe, Jon Duke, Jimeng Sun, and Jacob Eisenstein. 2018.
\newblock Explainable prediction of medical codes from clinical text.
\newblock In \emph{Proceedings of the 2018 Conference of the North {A}merican Chapter of the Association for Computational Linguistics: Human Language Technologies, Volume 1 (Long Papers)}, pages 1101--1111, Association for Computational Linguistics, New Orleans, Louisiana.

\bibitem[{Murdoch et~al.(2019)Murdoch, Singh, Kumbier, Abbasi-Asl, and Yu}]{murdoch_definitions_2019}
Murdoch, W.~James, Chandan Singh, Karl Kumbier, Reza Abbasi-Asl, and Bin Yu. 2019.
\newblock Definitions, methods, and applications in interpretable machine learning.
\newblock \emph{Proceedings of the National Academy of Sciences}, 116(44):22071--22080.
\newblock Publisher: Proceedings of the National Academy of Sciences.

\bibitem[{Mylonas, Mollas, and Tsoumakas(2022)}]{mylonas_attention_2022}
Mylonas, Nikolaos, Ioannis Mollas, and Grigorios Tsoumakas. 2022.
\newblock An {Attention} {Matrix} for {Every} {Decision}: {Faithfulness}-based {Arbitration} {Among} {Multiple} {Attention}-{Based} {Interpretations} of {Transformers} in {Text} {Classification}.

\bibitem[{Narang et~al.(2020)Narang, Raffel, Lee, Roberts, Fiedel, and Malkan}]{narang_wt5_2020}
Narang, Sharan, Colin Raffel, Katherine Lee, Adam Roberts, Noah Fiedel, and Karishma Malkan. 2020.
\newblock {WT5}?! {Training} {Text}-to-{Text} {Models} to {Explain} their {Predictions}.
\newblock \emph{ArXiv preprint}, abs/2004.14546.

\bibitem[{Nie, Zhang, and Patel(2018)}]{nie_theoretical_2018}
Nie, Weili, Yang Zhang, and Ankit Patel. 2018.
\newblock A theoretical explanation for perplexing behaviors of backpropagation-based visualizations.
\newblock In \emph{Proceedings of the 35th International Conference on Machine Learning, {ICML} 2018, Stockholmsm{\"{a}}ssan, Stockholm, Sweden, July 10-15, 2018}, volume~80 of \emph{Proceedings of Machine Learning Research}, pages 3806--3815, {PMLR}.

\bibitem[{Nye et~al.(2021)Nye, Andreassen, Gur-Ari, Michalewski, Austin, Bieber, Dohan, Lewkowycz, Bosma, Luan, Sutton, and Odena}]{nye_show_2021}
Nye, Maxwell, Anders~Johan Andreassen, Guy Gur-Ari, Henryk Michalewski, Jacob Austin, David Bieber, David Dohan, Aitor Lewkowycz, Maarten Bosma, David Luan, Charles Sutton, and Augustus Odena. 2021.
\newblock Show {Your} {Work}: {Scratchpads} for {Intermediate} {Computation} with {Language} {Models}.

\bibitem[{OpenAI(2023)}]{openai_gpt-4_2023}
OpenAI. 2023.
\newblock {GPT}-4 {Technical} {Report}.

\bibitem[{Park et~al.(2018)Park, Hendricks, Akata, Rohrbach, Schiele, Darrell, and Rohrbach}]{park_multimodal_2018}
Park, Dong~Huk, Lisa~Anne Hendricks, Zeynep Akata, Anna Rohrbach, Bernt Schiele, Trevor Darrell, and Marcus Rohrbach. 2018.
\newblock Multimodal explanations: Justifying decisions and pointing to the evidence.
\newblock In \emph{2018 {IEEE} Conference on Computer Vision and Pattern Recognition, {CVPR} 2018, Salt Lake City, UT, USA, June 18-22, 2018}, pages 8779--8788, {IEEE} Computer Society.

\bibitem[{Pascual, Brunner, and Wattenhofer(2021)}]{pascual_telling_2021}
Pascual, Damian, Gino Brunner, and Roger Wattenhofer. 2021.
\newblock Telling {BERT}{'}s full story: from local attention to global aggregation.
\newblock In \emph{Proceedings of the 16th Conference of the European Chapter of the Association for Computational Linguistics: Main Volume}, pages 105--124, Association for Computational Linguistics, Online.

\bibitem[{Petroni et~al.(2019)Petroni, Rockt{\"a}schel, Riedel, Lewis, Bakhtin, Wu, and Miller}]{petroni_language_2019}
Petroni, Fabio, Tim Rockt{\"a}schel, Sebastian Riedel, Patrick Lewis, Anton Bakhtin, Yuxiang Wu, and Alexander Miller. 2019.
\newblock Language models as knowledge bases?
\newblock In \emph{Proceedings of the 2019 Conference on Empirical Methods in Natural Language Processing and the 9th International Joint Conference on Natural Language Processing (EMNLP-IJCNLP)}, pages 2463--2473, Association for Computational Linguistics, Hong Kong, China.

\bibitem[{Pezeshkpour et~al.(2021)Pezeshkpour, Jain, Wallace, and Singh}]{pezeshkpour_empirical_2021}
Pezeshkpour, Pouya, Sarthak Jain, Byron Wallace, and Sameer Singh. 2021.
\newblock An empirical comparison of instance attribution methods for {NLP}.
\newblock In \emph{Proceedings of the 2021 Conference of the North American Chapter of the Association for Computational Linguistics: Human Language Technologies}, pages 967--975, Association for Computational Linguistics, Online.

\bibitem[{Poerner, Roth, and Sch{\"u}tze(2018)}]{poerner_interpretable_2018}
Poerner, Nina, Benjamin Roth, and Hinrich Sch{\"u}tze. 2018.
\newblock Interpretable textual neuron representations for {NLP}.
\newblock In \emph{Proceedings of the 2018 {EMNLP} Workshop {B}lackbox{NLP}: Analyzing and Interpreting Neural Networks for {NLP}}, pages 325--327, Association for Computational Linguistics, Brussels, Belgium.

\bibitem[{Poerner, Sch{\"u}tze, and Roth(2018)}]{poerner_evaluating_2018}
Poerner, Nina, Hinrich Sch{\"u}tze, and Benjamin Roth. 2018.
\newblock Evaluating neural network explanation methods using hybrid documents and morphosyntactic agreement.
\newblock In \emph{Proceedings of the 56th Annual Meeting of the Association for Computational Linguistics (Volume 1: Long Papers)}, pages 340--350, Association for Computational Linguistics, Melbourne, Australia.

\bibitem[{Poliak et~al.(2018)Poliak, Naradowsky, Haldar, Rudinger, and Van~Durme}]{poliak_hypothesis_2018}
Poliak, Adam, Jason Naradowsky, Aparajita Haldar, Rachel Rudinger, and Benjamin Van~Durme. 2018.
\newblock Hypothesis only baselines in natural language inference.
\newblock In \emph{Proceedings of the Seventh Joint Conference on Lexical and Computational Semantics}, pages 180--191, Association for Computational Linguistics, New Orleans, Louisiana.

\bibitem[{Pruthi et~al.(2022)Pruthi, Bansal, Dhingra, Baldini~Soares, Collins, Lipton, Neubig, and Cohen}]{pruthi_evaluating_2022}
Pruthi, Danish, Rachit Bansal, Bhuwan Dhingra, Livio Baldini~Soares, Michael Collins, Zachary~C. Lipton, Graham Neubig, and William~W. Cohen. 2022.
\newblock Evaluating explanations: How much do explanations from the teacher aid students?
\newblock \emph{Transactions of the Association for Computational Linguistics}, 10:359--375.

\bibitem[{Pruthi et~al.(2020)Pruthi, Gupta, Dhingra, Neubig, and Lipton}]{pruthi_learning_2020}
Pruthi, Danish, Mansi Gupta, Bhuwan Dhingra, Graham Neubig, and Zachary~C. Lipton. 2020.
\newblock Learning to deceive with attention-based explanations.
\newblock In \emph{Proceedings of the 58th Annual Meeting of the Association for Computational Linguistics}, pages 4782--4793, Association for Computational Linguistics, Online.

\bibitem[{Qian et~al.(2022)Qian, Wang, Li, Li, and Yan}]{qian_limitations_2022}
Qian, Jing, Hong Wang, Zekun Li, Shiyang Li, and Xifeng Yan. 2022.
\newblock Limitations of {Language} {Models} in {Arithmetic} and {Symbolic} {Induction}.

\bibitem[{Qian, Qiu, and Huang(2016)}]{qian_analyzing_2016}
Qian, Peng, Xipeng Qiu, and Xuanjing Huang. 2016.
\newblock Analyzing linguistic knowledge in sequential model of sentence.
\newblock In \emph{Proceedings of the 2016 Conference on Empirical Methods in Natural Language Processing}, pages 826--835, Association for Computational Linguistics, Austin, Texas.

\bibitem[{Raffel et~al.(2020)Raffel, Shazeer, Roberts, Lee, Narang, Matena, Zhou, Li, and Liu}]{raffel_exploring_2020}
Raffel, Colin, Noam Shazeer, Adam Roberts, Katherine Lee, Sharan Narang, Michael Matena, Yanqi Zhou, Wei Li, and Peter~J. Liu. 2020.
\newblock Exploring the limits of transfer learning with a unified text-to-text transformer.
\newblock \emph{J. Mach. Learn. Res.}, 21:140:1--140:67.

\bibitem[{Raganato, Scherrer, and Tiedemann(2020)}]{raganato_fixed_2020}
Raganato, Alessandro, Yves Scherrer, and J{\"o}rg Tiedemann. 2020.
\newblock Fixed encoder self-attention patterns in transformer-based machine translation.
\newblock In \emph{Findings of the Association for Computational Linguistics: EMNLP 2020}, pages 556--568, Association for Computational Linguistics, Online.

\bibitem[{Rajagopal et~al.(2021)Rajagopal, Balachandran, Hovy, and Tsvetkov}]{rajagopal_selfexplain_2021}
Rajagopal, Dheeraj, Vidhisha Balachandran, Eduard~H Hovy, and Yulia Tsvetkov. 2021.
\newblock {SELFEXPLAIN}: A self-explaining architecture for neural text classifiers.
\newblock In \emph{Proceedings of the 2021 Conference on Empirical Methods in Natural Language Processing}, pages 836--850, Association for Computational Linguistics, Online and Punta Cana, Dominican Republic.

\bibitem[{Rajani et~al.(2019)Rajani, McCann, Xiong, and Socher}]{rajani_explain_2019}
Rajani, Nazneen~Fatema, Bryan McCann, Caiming Xiong, and Richard Socher. 2019.
\newblock Explain yourself! leveraging language models for commonsense reasoning.
\newblock In \emph{Proceedings of the 57th Annual Meeting of the Association for Computational Linguistics}, pages 4932--4942, Association for Computational Linguistics, Florence, Italy.

\bibitem[{Ramamurthy et~al.(2020)Ramamurthy, Vinzamuri, Zhang, and Dhurandhar}]{natesan_ramamurthy_model_2020}
Ramamurthy, Karthikeyan~Natesan, Bhanukiran Vinzamuri, Yunfeng Zhang, and Amit Dhurandhar. 2020.
\newblock Model agnostic multilevel explanations.
\newblock In \emph{Advances in Neural Information Processing Systems 33: Annual Conference on Neural Information Processing Systems 2020, NeurIPS 2020, December 6-12, 2020, virtual}.

\bibitem[{Ravfogel et~al.(2020)Ravfogel, Elazar, Gonen, Twiton, and Goldberg}]{ravfogel_null_2020}
Ravfogel, Shauli, Yanai Elazar, Hila Gonen, Michael Twiton, and Yoav Goldberg. 2020.
\newblock Null it out: Guarding protected attributes by iterative nullspace projection.
\newblock In \emph{Proceedings of the 58th Annual Meeting of the Association for Computational Linguistics}, pages 7237--7256, Association for Computational Linguistics, Online.

\bibitem[{Ravfogel, Goldberg, and Cotterell(2022)}]{ravfogel_linear_2022}
Ravfogel, Shauli, Yoav Goldberg, and Ryan Cotterell. 2022.
\newblock Linear {Guardedness} and its {Implications}.

\bibitem[{Ravfogel et~al.(2021)Ravfogel, Prasad, Linzen, and Goldberg}]{ravfogel_counterfactual_2021}
Ravfogel, Shauli, Grusha Prasad, Tal Linzen, and Yoav Goldberg. 2021.
\newblock Counterfactual interventions reveal the causal effect of relative clause representations on agreement prediction.
\newblock In \emph{Proceedings of the 25th Conference on Computational Natural Language Learning}, pages 194--209, Association for Computational Linguistics, Online.

\bibitem[{Ravichander, Belinkov, and Hovy(2021)}]{ravichander_probing_2021}
Ravichander, Abhilasha, Yonatan Belinkov, and Eduard Hovy. 2021.
\newblock Probing the probing paradigm: Does probing accuracy entail task relevance?
\newblock In \emph{Proceedings of the 16th Conference of the European Chapter of the Association for Computational Linguistics: Main Volume}, pages 3363--3377, Association for Computational Linguistics, Online.

\bibitem[{Reif et~al.(2019)Reif, Yuan, Wattenberg, Vi{\'{e}}gas, Coenen, Pearce, and Kim}]{coenen_visualizing_2019}
Reif, Emily, Ann Yuan, Martin Wattenberg, Fernanda~B. Vi{\'{e}}gas, Andy Coenen, Adam Pearce, and Been Kim. 2019.
\newblock Visualizing and measuring the geometry of {BERT}.
\newblock In \emph{Advances in Neural Information Processing Systems 32: Annual Conference on Neural Information Processing Systems 2019, NeurIPS 2019, December 8-14, 2019, Vancouver, BC, Canada}, pages 8592--8600.

\bibitem[{Ribeiro, Singh, and Guestrin(2016)}]{ribeiro_why_2016}
Ribeiro, Marco~T{\'{u}}lio, Sameer Singh, and Carlos Guestrin. 2016.
\newblock "why should {I} trust you?": Explaining the predictions of any classifier.
\newblock In \emph{Proceedings of the 22nd {ACM} {SIGKDD} International Conference on Knowledge Discovery and Data Mining, San Francisco, CA, USA, August 13-17, 2016}, pages 1135--1144, {ACM}.

\bibitem[{Ribeiro, Singh, and Guestrin(2018)}]{ribeiro_anchors_2018}
Ribeiro, Marco~T{\'{u}}lio, Sameer Singh, and Carlos Guestrin. 2018.
\newblock Anchors: High-precision model-agnostic explanations.
\newblock In \emph{Proceedings of the Thirty-Second {AAAI} Conference on Artificial Intelligence, (AAAI-18), the 30th innovative Applications of Artificial Intelligence (IAAI-18), and the 8th {AAAI} Symposium on Educational Advances in Artificial Intelligence (EAAI-18), New Orleans, Louisiana, USA, February 2-7, 2018}, pages 1527--1535, {AAAI} Press.

\bibitem[{Roese and Olson(1995)}]{roese_counterfactual_1995}
Roese, Neal~J. and James~M. Olson. 1995.
\newblock Counterfactual thinking: {A} critical overview.
\newblock In \emph{What might have been: {The} social psychology of counterfactual thinking}. Lawrence Erlbaum Associates, Inc, Hillsdale, NJ, US, pages 1--55.

\bibitem[{Sajjad, Durrani, and Dalvi(2022)}]{sajjad_neuron-level_2022}
Sajjad, Hassan, Nadir Durrani, and Fahim Dalvi. 2022.
\newblock Neuron-level interpretation of deep {NLP} models: A survey.
\newblock \emph{Transactions of the Association for Computational Linguistics}, 10:1285--1303.

\bibitem[{Sakaguchi et~al.(2020)Sakaguchi, Bras, Bhagavatula, and Choi}]{sakaguchi_winogrande_2021}
Sakaguchi, Keisuke, Ronan~Le Bras, Chandra Bhagavatula, and Yejin Choi. 2020.
\newblock Winogrande: An adversarial winograd schema challenge at scale.
\newblock In \emph{The Thirty-Fourth {AAAI} Conference on Artificial Intelligence, {AAAI} 2020, The Thirty-Second Innovative Applications of Artificial Intelligence Conference, {IAAI} 2020, The Tenth {AAAI} Symposium on Educational Advances in Artificial Intelligence, {EAAI} 2020, New York, NY, USA, February 7-12, 2020}, pages 8732--8740, {AAAI} Press.

\bibitem[{Samek et~al.(2015)Samek, Binder, Montavon, Bach, and M{\"u}ller}]{samek_evaluating_2015}
Samek, Wojciech, Alexander Binder, Gr{\'e}goire Montavon, Sebastian Bach, and Klaus-Robert M{\"u}ller. 2015.
\newblock Evaluating the visualization of what a {Deep} {Neural} {Network} has learned.

\bibitem[{Schwartz, Thomson, and Smith(2018)}]{schwartz_bridging_2018}
Schwartz, Roy, Sam Thomson, and Noah~A. Smith. 2018.
\newblock Bridging {CNN}s, {RNN}s, and weighted finite-state machines.
\newblock In \emph{Proceedings of the 56th Annual Meeting of the Association for Computational Linguistics (Volume 1: Long Papers)}, pages 295--305, Association for Computational Linguistics, Melbourne, Australia.

\bibitem[{Serrano and Smith(2019)}]{serrano_is_2019}
Serrano, Sofia and Noah~A. Smith. 2019.
\newblock Is attention interpretable?
\newblock In \emph{Proceedings of the 57th Annual Meeting of the Association for Computational Linguistics}, pages 2931--2951, Association for Computational Linguistics, Florence, Italy.

\bibitem[{Shapley(1953)}]{kuhn_17_1953}
Shapley, L.~S. 1953.
\newblock 17. {A} {Value} for n-{Person} {Games}.
\newblock In Harold~William Kuhn and Albert~William Tucker, editors, \emph{Contributions to the {Theory} of {Games} ({AM}-28), {Volume} {II}}. Princeton University Press, pages 307--318.

\bibitem[{Shrikumar, Greenside, and Kundaje(2017)}]{shrikumar_learning_2017}
Shrikumar, Avanti, Peyton Greenside, and Anshul Kundaje. 2017.
\newblock Learning important features through propagating activation differences.
\newblock In \emph{Proceedings of the 34th International Conference on Machine Learning, {ICML} 2017, Sydney, NSW, Australia, 6-11 August 2017}, volume~70 of \emph{Proceedings of Machine Learning Research}, pages 3145--3153, {PMLR}.

\bibitem[{Shrikumar et~al.(2017)Shrikumar, Greenside, Shcherbina, and Kundaje}]{shrikumar_not_2017}
Shrikumar, Avanti, Peyton Greenside, Anna Shcherbina, and Anshul Kundaje. 2017.
\newblock Not {Just} a {Black} {Box}: {Learning} {Important} {Features} {Through} {Propagating} {Activation} {Differences}.
\newblock \emph{arXiv:1605.01713 [cs]}.
\newblock ArXiv: 1605.01713.

\bibitem[{Sia et~al.(2022)Sia, Belyy, Almahairi, Khabsa, Zettlemoyer, and Mathias}]{sia_logical_2022}
Sia, Suzanna, Anton Belyy, Amjad Almahairi, Madian Khabsa, Luke Zettlemoyer, and Lambert Mathias. 2022.
\newblock Logical {Satisfiability} of {Counterfactuals} for {Faithful} {Explanations} in {NLI}.

\bibitem[{Simonyan, Vedaldi, and Zisserman(2014)}]{simonyan_deep_2014}
Simonyan, Karen, Andrea Vedaldi, and Andrew Zisserman. 2014.
\newblock Deep inside convolutional networks: {Visualising} image classification models and saliency maps.
\newblock In \emph{In {Workshop} at {International} {Conference} on {Learning} {Representations}}.

\bibitem[{Slack et~al.(2020)Slack, Hilgard, Jia, Singh, and Lakkaraju}]{slack_fooling_2020}
Slack, Dylan, Sophie Hilgard, Emily Jia, Sameer Singh, and Himabindu Lakkaraju. 2020.
\newblock Fooling {LIME} and {SHAP}: {Adversarial} {Attacks} on {Post} hoc {Explanation} {Methods}.
\newblock In \emph{Proceedings of the {AAAI}/{ACM} {Conference} on {AI}, {Ethics}, and {Society}}. Association for Computing Machinery, New York, NY, USA, pages 180--186.

\bibitem[{Smilkov et~al.(2017)Smilkov, Thorat, Kim, Vi{\'e}gas, and Wattenberg}]{smilkov_smoothgrad_2017}
Smilkov, Daniel, Nikhil Thorat, Been Kim, Fernanda Vi{\'e}gas, and Martin Wattenberg. 2017.
\newblock {SmoothGrad}: removing noise by adding noise.
\newblock \emph{ArXiv preprint}, abs/1706.03825.

\bibitem[{Springenberg et~al.(2015)Springenberg, Dosovitskiy, Brox, and Riedmiller}]{springenberg_striving_2015}
Springenberg, J., Alexey Dosovitskiy, Thomas Brox, and M.~Riedmiller. 2015.
\newblock Striving for {Simplicity}: {The} {All} {Convolutional} {Net}.

\bibitem[{Strobelt et~al.(2019)Strobelt, Gehrmann, Behrisch, Perer, Pfister, and Rush}]{strobelt_seq2seq-vis_2019}
Strobelt, Hendrik, Sebastian Gehrmann, Michael Behrisch, Adam Perer, Hanspeter Pfister, and Alexander~M. Rush. 2019.
\newblock Seq2seq-{Vis}: {A} {Visual} {Debugging} {Tool} for {Sequence}-to-{Sequence} {Models}.
\newblock \emph{IEEE Transactions on Visualization and Computer Graphics}, 25(1):353--363.
\newblock Conference Name: IEEE Transactions on Visualization and Computer Graphics.

\bibitem[{Strobelt et~al.(2018)Strobelt, Gehrmann, Pfister, and Rush}]{strobelt_lstmvis_2018}
Strobelt, Hendrik, Sebastian Gehrmann, Hanspeter Pfister, and Alexander~M. Rush. 2018.
\newblock {LSTMVis}: {A} {Tool} for {Visual} {Analysis} of {Hidden} {State} {Dynamics} in {Recurrent} {Neural} {Networks}.
\newblock \emph{IEEE Transactions on Visualization and Computer Graphics}, 24(1):667--676.
\newblock Conference Name: IEEE Transactions on Visualization and Computer Graphics.

\bibitem[{Subramanian et~al.(2020)Subramanian, Bogin, Gupta, Wolfson, Singh, Berant, and Gardner}]{subramanian_obtaining_2020}
Subramanian, Sanjay, Ben Bogin, Nitish Gupta, Tomer Wolfson, Sameer Singh, Jonathan Berant, and Matt Gardner. 2020.
\newblock Obtaining faithful interpretations from compositional neural networks.
\newblock In \emph{Proceedings of the 58th Annual Meeting of the Association for Computational Linguistics}, pages 5594--5608, Association for Computational Linguistics, Online.

\bibitem[{Sundararajan, Taly, and Yan(2017)}]{sundararajan_axiomatic_2017}
Sundararajan, Mukund, Ankur Taly, and Qiqi Yan. 2017.
\newblock Axiomatic attribution for deep networks.
\newblock In \emph{Proceedings of the 34th International Conference on Machine Learning, {ICML} 2017, Sydney, NSW, Australia, 6-11 August 2017}, volume~70 of \emph{Proceedings of Machine Learning Research}, pages 3319--3328, {PMLR}.

\bibitem[{Sushil et~al.(2018)Sushil, {\v S}uster, Luyckx, and Daelemans}]{sushil_patient_2018}
Sushil, Madhumita, Simon {\v S}uster, Kim Luyckx, and Walter Daelemans. 2018.
\newblock Patient representation learning and interpretable evaluation using clinical notes.
\newblock \emph{Journal of Biomedical Informatics}, 84:103--113.

\bibitem[{Tafjord, Dalvi, and Clark(2021)}]{tafjord_proofwriter_2021}
Tafjord, Oyvind, Bhavana Dalvi, and Peter Clark. 2021.
\newblock {P}roof{W}riter: Generating implications, proofs, and abductive statements over natural language.
\newblock In \emph{Findings of the Association for Computational Linguistics: ACL-IJCNLP 2021}, pages 3621--3634, Association for Computational Linguistics, Online.

\bibitem[{Tenney et~al.(2020)Tenney, Wexler, Bastings, Bolukbasi, Coenen, Gehrmann, Jiang, Pushkarna, Radebaugh, Reif, and Yuan}]{tenney_language_2020}
Tenney, Ian, James Wexler, Jasmijn Bastings, Tolga Bolukbasi, Andy Coenen, Sebastian Gehrmann, Ellen Jiang, Mahima Pushkarna, Carey Radebaugh, Emily Reif, and Ann Yuan. 2020.
\newblock The language interpretability tool: Extensible, interactive visualizations and analysis for {NLP} models.
\newblock In \emph{Proceedings of the 2020 Conference on Empirical Methods in Natural Language Processing: System Demonstrations}, pages 107--118, Association for Computational Linguistics, Online.

\bibitem[{Tsang, Rambhatla, and Liu(2020)}]{tsang_how_2020}
Tsang, Michael, Sirisha Rambhatla, and Yan Liu. 2020.
\newblock How does this interaction affect me? interpretable attribution for feature interactions.
\newblock In \emph{Advances in Neural Information Processing Systems 33: Annual Conference on Neural Information Processing Systems 2020, NeurIPS 2020, December 6-12, 2020, virtual}.

\bibitem[{Tucker, Qian, and Levy(2021)}]{tucker_what_2021}
Tucker, Mycal, Peng Qian, and Roger Levy. 2021.
\newblock What if this modified that? syntactic interventions with counterfactual embeddings.
\newblock In \emph{Findings of the Association for Computational Linguistics: ACL-IJCNLP 2021}, pages 862--875, Association for Computational Linguistics, Online.

\bibitem[{Tutek and Snajder(2020)}]{tutek_staying_2020}
Tutek, Martin and Jan Snajder. 2020.
\newblock Staying true to your word: (how) can attention become explanation?
\newblock In \emph{Proceedings of the 5th Workshop on Representation Learning for NLP}, pages 131--142, Association for Computational Linguistics, Online.

\bibitem[{Vashishth et~al.(2019)Vashishth, Upadhyay, Tomar, and Faruqui}]{vashishth_attention_2019}
Vashishth, Shikhar, Shyam Upadhyay, Gaurav~Singh Tomar, and Manaal Faruqui. 2019.
\newblock Attention {Interpretability} {Across} {NLP} {Tasks}.
\newblock \emph{ArXiv preprint}, abs/1909.11218.

\bibitem[{Vaswani et~al.(2017)Vaswani, Shazeer, Parmar, Uszkoreit, Jones, Gomez, Kaiser, and Polosukhin}]{vaswani_attention_2017}
Vaswani, Ashish, Noam Shazeer, Niki Parmar, Jakob Uszkoreit, Llion Jones, Aidan~N. Gomez, Lukasz Kaiser, and Illia Polosukhin. 2017.
\newblock Attention is all you need.
\newblock In \emph{Advances in Neural Information Processing Systems 30: Annual Conference on Neural Information Processing Systems 2017, December 4-9, 2017, Long Beach, CA, {USA}}, pages 5998--6008.

\bibitem[{Veldhoen, Hupkes, and Zuidema(2016)}]{veldhoen_diagnostic_2016}
Veldhoen, Sara, Dieuwke Hupkes, and Willem Zuidema. 2016.
\newblock Diagnostic classifiers: revealing how neural networks process hierarchical structure.
\newblock page~9.

\bibitem[{Vig(2019)}]{vig_visualizing_2019}
Vig, Jesse. 2019.
\newblock Visualizing {Attention} in {Transformer}-{Based} {Language} {Representation} {Models}.
\newblock \emph{ArXiv preprint}, abs/1904.02679.

\bibitem[{Vig et~al.(2020)Vig, Gehrmann, Belinkov, Qian, Nevo, Singer, and Shieber}]{vig_investigating_2020}
Vig, Jesse, Sebastian Gehrmann, Yonatan Belinkov, Sharon Qian, Daniel Nevo, Yaron Singer, and Stuart~M. Shieber. 2020.
\newblock Investigating gender bias in language models using causal mediation analysis.
\newblock In \emph{Advances in Neural Information Processing Systems 33: Annual Conference on Neural Information Processing Systems 2020, NeurIPS 2020, December 6-12, 2020, virtual}.

\bibitem[{Voita et~al.(2019)Voita, Talbot, Moiseev, Sennrich, and Titov}]{voita_analyzing_2019}
Voita, Elena, David Talbot, Fedor Moiseev, Rico Sennrich, and Ivan Titov. 2019.
\newblock Analyzing multi-head self-attention: Specialized heads do the heavy lifting, the rest can be pruned.
\newblock In \emph{Proceedings of the 57th Annual Meeting of the Association for Computational Linguistics}, pages 5797--5808, Association for Computational Linguistics, Florence, Italy.

\bibitem[{Voita and Titov(2020)}]{voita_information-theoretic_2020}
Voita, Elena and Ivan Titov. 2020.
\newblock Information-theoretic probing with minimum description length.
\newblock In \emph{Proceedings of the 2020 Conference on Empirical Methods in Natural Language Processing (EMNLP)}, pages 183--196, Association for Computational Linguistics, Online.

\bibitem[{Wallace, Feng, and Boyd-Graber(2018)}]{wallace_interpreting_2018}
Wallace, Eric, Shi Feng, and Jordan Boyd-Graber. 2018.
\newblock Interpreting neural networks with nearest neighbors.
\newblock In \emph{Proceedings of the 2018 {EMNLP} Workshop {B}lackbox{NLP}: Analyzing and Interpreting Neural Networks for {NLP}}, pages 136--144, Association for Computational Linguistics, Brussels, Belgium.

\bibitem[{Wallace et~al.(2019{\natexlab{a}})Wallace, Feng, Kandpal, Gardner, and Singh}]{wallace_universal_2019}
Wallace, Eric, Shi Feng, Nikhil Kandpal, Matt Gardner, and Sameer Singh. 2019{\natexlab{a}}.
\newblock Universal adversarial triggers for attacking and analyzing {NLP}.
\newblock In \emph{Proceedings of the 2019 Conference on Empirical Methods in Natural Language Processing and the 9th International Joint Conference on Natural Language Processing (EMNLP-IJCNLP)}, pages 2153--2162, Association for Computational Linguistics, Hong Kong, China.

\bibitem[{Wallace, Gardner, and Singh(2020)}]{wallace_interpreting_2020}
Wallace, Eric, Matt Gardner, and Sameer Singh. 2020.
\newblock Interpreting predictions of {NLP} models.
\newblock In \emph{Proceedings of the 2020 Conference on Empirical Methods in Natural Language Processing: Tutorial Abstracts}, pages 20--23, Association for Computational Linguistics, Online.

\bibitem[{Wallace et~al.(2019{\natexlab{b}})Wallace, Tuyls, Wang, Subramanian, Gardner, and Singh}]{wallace_allennlp_2019}
Wallace, Eric, Jens Tuyls, Junlin Wang, Sanjay Subramanian, Matt Gardner, and Sameer Singh. 2019{\natexlab{b}}.
\newblock {A}llen{NLP} interpret: A framework for explaining predictions of {NLP} models.
\newblock In \emph{Proceedings of the 2019 Conference on Empirical Methods in Natural Language Processing and the 9th International Joint Conference on Natural Language Processing (EMNLP-IJCNLP): System Demonstrations}, pages 7--12, Association for Computational Linguistics, Hong Kong, China.

\bibitem[{Wang et~al.(2019{\natexlab{a}})Wang, Pruksachatkun, Nangia, Singh, Michael, Hill, Levy, and Bowman}]{wang_superglue_2019}
Wang, Alex, Yada Pruksachatkun, Nikita Nangia, Amanpreet Singh, Julian Michael, Felix Hill, Omer Levy, and Samuel~R. Bowman. 2019{\natexlab{a}}.
\newblock Superglue: {A} stickier benchmark for general-purpose language understanding systems.
\newblock In \emph{Advances in Neural Information Processing Systems 32: Annual Conference on Neural Information Processing Systems 2019, NeurIPS 2019, December 8-14, 2019, Vancouver, BC, Canada}, pages 3261--3275.

\bibitem[{Wang et~al.(2019{\natexlab{b}})Wang, Singh, Michael, Hill, Levy, and Bowman}]{wang_glue_2018}
Wang, Alex, Amanpreet Singh, Julian Michael, Felix Hill, Omer Levy, and Samuel~R. Bowman. 2019{\natexlab{b}}.
\newblock {GLUE:} {A} multi-task benchmark and analysis platform for natural language understanding.
\newblock In \emph{7th International Conference on Learning Representations, {ICLR} 2019, New Orleans, LA, USA, May 6-9, 2019}, OpenReview.net.

\bibitem[{Wang et~al.(2020)Wang, Tuyls, Wallace, and Singh}]{wang_gradient-based_2020}
Wang, Junlin, Jens Tuyls, Eric Wallace, and Sameer Singh. 2020.
\newblock Gradient-based analysis of {NLP} models is manipulable.
\newblock In \emph{Findings of the Association for Computational Linguistics: EMNLP 2020}, pages 247--258, Association for Computational Linguistics, Online.

\bibitem[{Wang et~al.(2022{\natexlab{a}})Wang, Shen, Peng, Zhang, Xiao, Liu, Tang, Chen, Wu, and Wang}]{wang_fine-grained_2022}
Wang, Lijie, Yaozong Shen, Shuyuan Peng, Shuai Zhang, Xinyan Xiao, Hao Liu, Hongxuan Tang, Ying Chen, Hua Wu, and Haifeng Wang. 2022{\natexlab{a}}.
\newblock A fine-grained interpretability evaluation benchmark for neural {NLP}.
\newblock In \emph{Proceedings of the 26th Conference on Computational Natural Language Learning (CoNLL)}, pages 70--84, Association for Computational Linguistics, Abu Dhabi, United Arab Emirates (Hybrid).

\bibitem[{Wang et~al.(2022{\natexlab{b}})Wang, Wei, Schuurmans, Le, Chi, Narang, Chowdhery, and Zhou}]{wang_self-consistency_2022}
Wang, Xuezhi, Jason Wei, Dale Schuurmans, Quoc Le, Ed~Chi, Sharan Narang, Aakanksha Chowdhery, and Denny Zhou. 2022{\natexlab{b}}.
\newblock Self-{Consistency} {Improves} {Chain} of {Thought} {Reasoning} in {Language} {Models}.

\bibitem[{Wei et~al.(2022)Wei, Wang, Schuurmans, Bosma, Ichter, Xia, Chi, Le, and Zhou}]{wei_chain_2022}
Wei, Jason, Xuezhi Wang, Dale Schuurmans, Maarten Bosma, Brian Ichter, Fei Xia, Ed~Chi, Quoc Le, and Denny Zhou. 2022.
\newblock Chain of {Thought} {Prompting} {Elicits} {Reasoning} in {Large} {Language} {Models}.

\bibitem[{Wexler et~al.(2020)Wexler, Pushkarna, Bolukbasi, Wattenberg, Vi{\'e}gas, and Wilson}]{wexler_what-if_2020}
Wexler, James, Mahima Pushkarna, Tolga Bolukbasi, Martin Wattenberg, Fernanda Vi{\'e}gas, and Jimbo Wilson. 2020.
\newblock The {What}-{If} {Tool}: {Interactive} {Probing} of {Machine} {Learning} {Models}.
\newblock \emph{IEEE Transactions on Visualization and Computer Graphics}, 26(1):56--65.
\newblock Conference Name: IEEE Transactions on Visualization and Computer Graphics.

\bibitem[{Wiegreffe et~al.(2022)Wiegreffe, Hessel, Swayamdipta, Riedl, and Choi}]{wiegreffe_reframing_2022}
Wiegreffe, Sarah, Jack Hessel, Swabha Swayamdipta, Mark Riedl, and Yejin Choi. 2022.
\newblock Reframing human-{AI} collaboration for generating free-text explanations.
\newblock In \emph{Proceedings of the 2022 Conference of the North American Chapter of the Association for Computational Linguistics: Human Language Technologies}, pages 632--658, Association for Computational Linguistics, Seattle, United States.

\bibitem[{Wiegreffe, Marasovi{\'c}, and Smith(2021)}]{wiegreffe_measuring_2021}
Wiegreffe, Sarah, Ana Marasovi{\'c}, and Noah~A. Smith. 2021.
\newblock {M}easuring association between labels and free-text rationales.
\newblock In \emph{Proceedings of the 2021 Conference on Empirical Methods in Natural Language Processing}, pages 10266--10284, Association for Computational Linguistics, Online and Punta Cana, Dominican Republic.

\bibitem[{Wiegreffe and Pinter(2019)}]{wiegreffe_attention_2019}
Wiegreffe, Sarah and Yuval Pinter. 2019.
\newblock Attention is not not explanation.
\newblock In \emph{Proceedings of the 2019 Conference on Empirical Methods in Natural Language Processing and the 9th International Joint Conference on Natural Language Processing (EMNLP-IJCNLP)}, pages 11--20, Association for Computational Linguistics, Hong Kong, China.

\bibitem[{Winship and Morgan(1999)}]{winship_estimation_1999}
Winship, Christopher and Stephen~L. Morgan. 1999.
\newblock The {Estimation} of {Causal} {Effects} from {Observational} {Data}.
\newblock \emph{Annual Review of Sociology}, 25(1):659--706.
\newblock \_eprint: https://doi.org/10.1146/annurev.soc.25.1.659.

\bibitem[{Wu et~al.(2021)Wu, Ribeiro, Heer, and Weld}]{wu_polyjuice_2021}
Wu, Tongshuang, Marco~Tulio Ribeiro, Jeffrey Heer, and Daniel Weld. 2021.
\newblock Polyjuice: Generating counterfactuals for explaining, evaluating, and improving models.
\newblock In \emph{Proceedings of the 59th Annual Meeting of the Association for Computational Linguistics and the 11th International Joint Conference on Natural Language Processing (Volume 1: Long Papers)}, pages 6707--6723, Association for Computational Linguistics, Online.

\bibitem[{Xie et~al.(2017)Xie, Ma, Dai, and Hovy}]{xie_interpretable_2017}
Xie, Qizhe, Xuezhe Ma, Zihang Dai, and Eduard Hovy. 2017.
\newblock An interpretable knowledge transfer model for knowledge base completion.
\newblock In \emph{Proceedings of the 55th Annual Meeting of the Association for Computational Linguistics (Volume 1: Long Papers)}, pages 950--962, Association for Computational Linguistics, Vancouver, Canada.

\bibitem[{Yang and Kim(2019)}]{yang_benchmarking_2019}
Yang, Mengjiao and Been Kim. 2019.
\newblock Benchmarking {Attribution} {Methods} with {Relative} {Feature} {Importance}.

\bibitem[{Yang et~al.(2018)Yang, Qi, Zhang, Bengio, Cohen, Salakhutdinov, and Manning}]{yang_hotpotqa_2018}
Yang, Zhilin, Peng Qi, Saizheng Zhang, Yoshua Bengio, William Cohen, Ruslan Salakhutdinov, and Christopher~D. Manning. 2018.
\newblock {H}otpot{QA}: A dataset for diverse, explainable multi-hop question answering.
\newblock In \emph{Proceedings of the 2018 Conference on Empirical Methods in Natural Language Processing}, pages 2369--2380, Association for Computational Linguistics, Brussels, Belgium.

\bibitem[{Ye and Durrett(2022)}]{ye_unreliability_2022}
Ye, Xi and Greg Durrett. 2022.
\newblock The {Unreliability} of {Explanations} in {Few}-shot {Prompting} for {Textual} {Reasoning}.

\bibitem[{Ye and Durrett(2023)}]{ye_explanation_2023}
Ye, Xi and Greg Durrett. 2023.
\newblock Explanation {Selection} {Using} {Unlabeled} {Data} for {In}-{Context} {Learning}.

\bibitem[{Ye et~al.(2022)Ye, Iyer, Celikyilmaz, Stoyanov, Durrett, and Pasunuru}]{ye_complementary_2022}
Ye, Xi, Srinivasan Iyer, Asli Celikyilmaz, Ves Stoyanov, Greg Durrett, and Ramakanth Pasunuru. 2022.
\newblock Complementary {Explanations} for {Effective} {In}-{Context} {Learning}.

\bibitem[{Ye, Nair, and Durrett(2021)}]{ye_connecting_2021}
Ye, Xi, Rohan Nair, and Greg Durrett. 2021.
\newblock Connecting attributions and {QA} model behavior on realistic counterfactuals.
\newblock In \emph{Proceedings of the 2021 Conference on Empirical Methods in Natural Language Processing}, pages 5496--5512, Association for Computational Linguistics, Online and Punta Cana, Dominican Republic.

\bibitem[{Yeh et~al.(2019)Yeh, Hsieh, Suggala, Inouye, and Ravikumar}]{yeh_fidelity_2019}
Yeh, Chih{-}Kuan, Cheng{-}Yu Hsieh, Arun~Sai Suggala, David~I. Inouye, and Pradeep Ravikumar. 2019.
\newblock On the (in)fidelity and sensitivity of explanations.
\newblock In \emph{Advances in Neural Information Processing Systems 32: Annual Conference on Neural Information Processing Systems 2019, NeurIPS 2019, December 8-14, 2019, Vancouver, BC, Canada}, pages 10965--10976.

\bibitem[{Yeh et~al.(2020)Yeh, Kim, Arik, Li, Pfister, and Ravikumar}]{yeh_completeness-aware_2020}
Yeh, Chih{-}Kuan, Been Kim, Sercan~{\"{O}}mer Arik, Chun{-}Liang Li, Tomas Pfister, and Pradeep Ravikumar. 2020.
\newblock On completeness-aware concept-based explanations in deep neural networks.
\newblock In \emph{Advances in Neural Information Processing Systems 33: Annual Conference on Neural Information Processing Systems 2020, NeurIPS 2020, December 6-12, 2020, virtual}.

\bibitem[{Yi et~al.(2018)Yi, Wu, Gan, Torralba, Kohli, and Tenenbaum}]{yi_neural-symbolic_2018}
Yi, Kexin, Jiajun Wu, Chuang Gan, Antonio Torralba, Pushmeet Kohli, and Josh Tenenbaum. 2018.
\newblock Neural-symbolic {VQA:} disentangling reasoning from vision and language understanding.
\newblock In \emph{Advances in Neural Information Processing Systems 31: Annual Conference on Neural Information Processing Systems 2018, NeurIPS 2018, December 3-8, 2018, Montr{\'{e}}al, Canada}, pages 1039--1050.

\bibitem[{Yin et~al.(2022)Yin, Shi, Hsieh, and Chang}]{yin_sensitivity_2022}
Yin, Fan, Zhouxing Shi, Cho-Jui Hsieh, and Kai-Wei Chang. 2022.
\newblock On the sensitivity and stability of model interpretations in {NLP}.
\newblock In \emph{Proceedings of the 60th Annual Meeting of the Association for Computational Linguistics (Volume 1: Long Papers)}, pages 2631--2647, Association for Computational Linguistics, Dublin, Ireland.

\bibitem[{Yin and Neubig(2022)}]{yin_interpreting_2022}
Yin, Kayo and Graham Neubig. 2022.
\newblock Interpreting language models with contrastive explanations.
\newblock In \emph{Proceedings of the 2022 Conference on Empirical Methods in Natural Language Processing}, pages 184--198, Association for Computational Linguistics, Abu Dhabi, United Arab Emirates.

\bibitem[{Zaidan, Eisner, and Piatko(2007)}]{zaidan_using_2007}
Zaidan, Omar, Jason Eisner, and Christine Piatko. 2007.
\newblock Using {\textquotedblleft}{Annotator} {Rationales}{\textquotedblright} to {Improve} {Machine} {Learning} for {Text} {Categorization}.
\newblock In \emph{Human {Language} {Technologies} 2007: {The} {Conference} of the {North} {American} {Chapter} of the {Association} for {Computational} {Linguistics}; {Proceedings} of the {Main} {Conference}}, pages 260--267, Association for Computational Linguistics, Rochester, New York.

\bibitem[{Zeiler and Fergus(2014)}]{fleet_visualizing_2014}
Zeiler, Matthew~D. and Rob Fergus. 2014.
\newblock Visualizing and {Understanding} {Convolutional} {Networks}.
\newblock In David Fleet, Tomas Pajdla, Bernt Schiele, and Tinne Tuytelaars, editors, \emph{Computer {Vision} {\textendash} {ECCV} 2014}, volume 8689. Springer International Publishing, Cham, pages 818--833.
\newblock Series Title: Lecture Notes in Computer Science.

\bibitem[{Zellers et~al.(2019)Zellers, Holtzman, Bisk, Farhadi, and Choi}]{zellers_hellaswag_2019}
Zellers, Rowan, Ari Holtzman, Yonatan Bisk, Ali Farhadi, and Yejin Choi. 2019.
\newblock {H}ella{S}wag: Can a machine really finish your sentence?
\newblock In \emph{Proceedings of the 57th Annual Meeting of the Association for Computational Linguistics}, pages 4791--4800, Association for Computational Linguistics, Florence, Italy.

\bibitem[{Zheng et~al.(2022)Zheng, Booth, Shah, and Zhou}]{zheng_irrationality_2021}
Zheng, Yiming, Serena Booth, Julie Shah, and Yilun Zhou. 2022.
\newblock The irrationality of neural rationale models.
\newblock In \emph{Proceedings of the 2nd Workshop on Trustworthy Natural Language Processing (TrustNLP 2022)}, pages 64--73, Association for Computational Linguistics, Seattle, U.S.A.

\bibitem[{Zhou et~al.(2022{\natexlab{a}})Zhou, Sch{\"a}rli, Hou, Wei, Scales, Wang, Schuurmans, Bousquet, Le, and Chi}]{zhou_least--most_2022}
Zhou, Denny, Nathanael Sch{\"a}rli, Le~Hou, Jason Wei, Nathan Scales, Xuezhi Wang, Dale Schuurmans, Olivier Bousquet, Quoc Le, and Ed~Chi. 2022{\natexlab{a}}.
\newblock Least-to-{Most} {Prompting} {Enables} {Complex} {Reasoning} in {Large} {Language} {Models}.

\bibitem[{Zhou et~al.(2022{\natexlab{b}})Zhou, Booth, Ribeiro, and Shah}]{zhou_feature_2021}
Zhou, Yilun, Serena Booth, Marco~T{\'{u}}lio Ribeiro, and Julie Shah. 2022{\natexlab{b}}.
\newblock Do feature attribution methods correctly attribute features?
\newblock In \emph{Thirty-Sixth {AAAI} Conference on Artificial Intelligence, {AAAI} 2022, Thirty-Fourth Conference on Innovative Applications of Artificial Intelligence, {IAAI} 2022, The Twelveth Symposium on Educational Advances in Artificial Intelligence, {EAAI} 2022 Virtual Event, February 22 - March 1, 2022}, pages 9623--9633, {AAAI} Press.

\bibitem[{Zhou, Ribeiro, and Shah(2022)}]{zhou_exsum_2022}
Zhou, Yilun, Marco~Tulio Ribeiro, and Julie Shah. 2022.
\newblock {ExSum}: {F}rom local explanations to model understanding.
\newblock In \emph{Proceedings of the 2022 Conference of the North American Chapter of the Association for Computational Linguistics: Human Language Technologies}, pages 5359--5378, Association for Computational Linguistics, Seattle, United States.

\bibitem[{Zmigrod et~al.(2019)Zmigrod, Mielke, Wallach, and Cotterell}]{zmigrod_counterfactual_2019}
Zmigrod, Ran, Sabrina~J. Mielke, Hanna Wallach, and Ryan Cotterell. 2019.
\newblock Counterfactual data augmentation for mitigating gender stereotypes in languages with rich morphology.
\newblock In \emph{Proceedings of the 57th Annual Meeting of the Association for Computational Linguistics}, pages 1651--1661, Association for Computational Linguistics, Florence, Italy.

\end{thebibliography}
\end{document}